\documentclass{article} % For LaTeX2e
\usepackage{iclr2021_conference,times}

\usepackage{hyperref}
\usepackage{minitoc}

\usepackage{url}
\usepackage[toc,page,header]{appendix}
\usepackage{amsthm}
\usepackage{tikz}
\usepackage{enumitem}
\usepackage{subcaption}
\usepackage{amssymb}
\usepackage{amsmath,mathtools}

\usepackage{tikz}
\usetikzlibrary{matrix,fit, decorations.pathreplacing}
\usepackage{gensymb}
\usepackage{subcaption}
\newtheorem{definition}{Definition}

\usepackage{amsmath,amsfonts,bm}

\def\eqref#1{equation~\ref{#1}}
\def\1{\bm{1}}

\def\eps{{\epsilon}}

\DeclareMathAlphabet{\mathsfit}{\encodingdefault}{\sfdefault}{m}{sl}
\SetMathAlphabet{\mathsfit}{bold}{\encodingdefault}{\sfdefault}{bx}{n}

\usepackage{hyperref}
\usepackage{url}
\usepackage{tikz}
\usetikzlibrary{matrix,fit, decorations.pathreplacing}

\title{Addressing the Topological Defects of Disentanglement via Distributed Operators}
\iclrfinalcopy

\author{Diane Bouchacourt\thanks{equal contribution},~~Mark Ibrahim\footnotemark[1], ~Stéphane Deny \\
Facebook AI Research\\
\texttt{\{dianeb,marksibrahim,sdeny\}@fb.com} \\
}
\begin{document}

\doparttoc % Tell to minitoc to generate a toc for the parts
\faketableofcontents % Run a fake tableofcontents command for the partocs

\part{} % Start the document part
%\parttoc % Insert the document TOC

\maketitle

\begin{abstract}
% to unobserved contexts remains a key challenge of Machine Learning. Disentanglement is a candidate solution to this problem that consists of identifying and isolating the natural factors of variations present in a dataset, so as to adapt the representation to changes in context.
A core challenge in Machine Learning is to learn to disentangle natural factors of variation in data (e.g. object shape \emph{vs} pose). A popular approach to disentanglement consists in learning to map each of these factors to distinct subspaces of a model's latent representation. However, this approach has shown limited empirical success to date. Here, we show that, for a broad family of transformations acting on images---encompassing simple affine transformations such as rotations and translations---this approach to disentanglement introduces topological defects (i.e. discontinuities in the encoder).  Motivated by classical results from group representation theory, we study an alternative, more flexible approach to disentanglement which relies on distributed latent operators, potentially acting on the entire latent space. We theoretically and empirically demonstrate the effectiveness of this approach to disentangle affine transformations. Our work lays a theoretical foundation for the recent success of a new generation of models using distributed operators for disentanglement.\footnote{All code is available at \url{https://github.com/facebookresearch/Addressing-the-Topological-Defects-of-Disentanglement}}

%\MI{Would "adapt" or "generalize" make more senese than "abstract"?}
%\MI{It's may not be obvious from this first sentence what "new situations" means. CCI-VAE uses "representation learning" as way to set the context for disentanglement. Could we introduce disentanglement through that context too?}
%\MI{Another option rephrasing the first few sentences. I'm not convinced this is necessarily better, so feel free to use or disregard bits at will: 
%"A core challenge in representation learning is uncovering the factors of natural variation in data. Disentanglement, a popular approach, consists of identifying and isolating those factors often by learning a representation with subspaces that capture distinct factors of variation. Despite its intuitive characterization, this approach to disentanglement has had limited empirical success to date..."
%}\SD{It's nice but a bit nerdy :) How could we address a broader audience?}

\end{abstract}

\section{Introduction}

\begin{figure}[t!]
\begin{center}
\includegraphics[width=14cm]{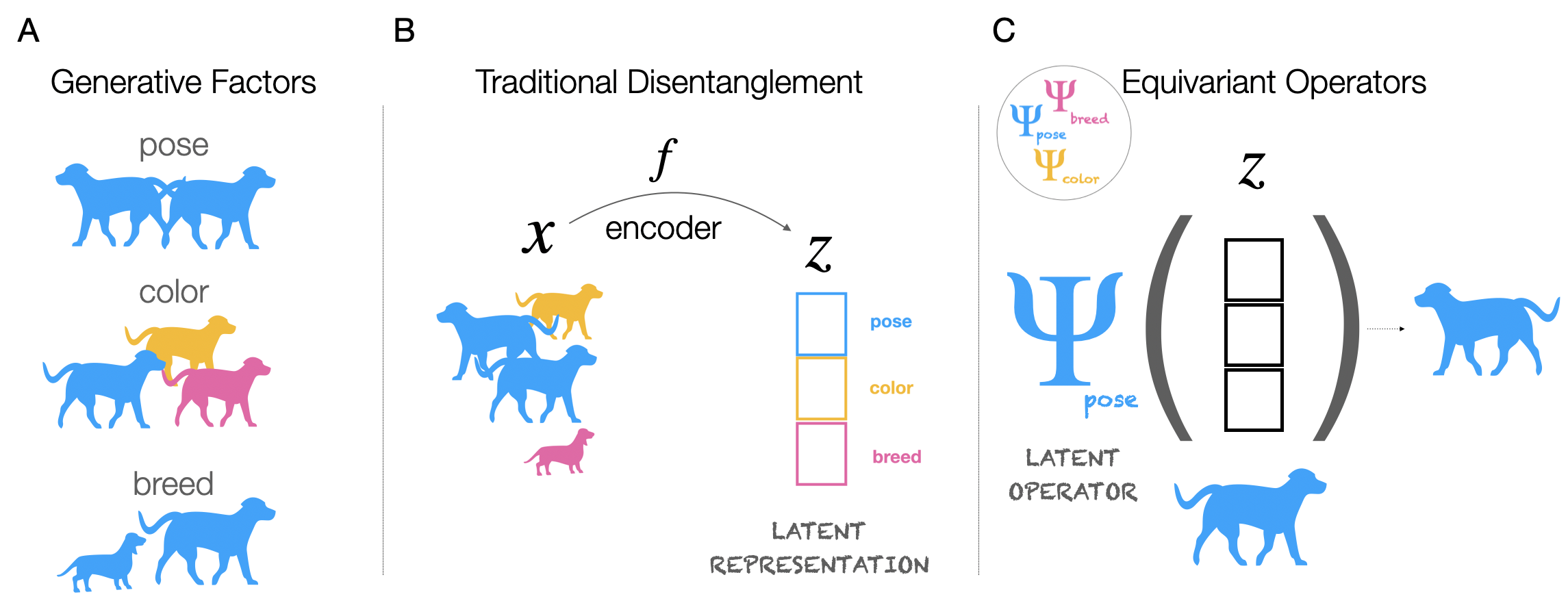}
\end{center}
\caption{\textbf{We propose disentanglement via distributed latent operators as an alternative to traditional disentanglement via subspaces.} \textbf{A.} Consider a dog dataset where the generative factors of variation of the dataset are pose, color, and breed. \textbf{B.} Traditional disentanglement attempts to isolate each factor of variation into a subspace of the latent representation. \textbf{C.} We propose an alternative form of disentanglement via distributed latent operators, acting on the entire latent space. In this example, the pose operator $\psi_{\text{pose}}$, which flips the orientation of the dog, acts on the entire latent space as opposed to a subspace. We show that distributed operators overcome the topological limitations of traditional disentanglement.}
\label{fig:figConceptual}
\end{figure}

%Deep neural networks have achieved remarkable success at learning invariances to complex tranformations \citep{lecun_deep_2015} \MI{define invariances: it may not be obvious to a reader what this means. Perhaps we can say, "invariant to complex transformations"?}. 
Machine Learning systems do not have the human capacity to generalize outside their domain of training. As humans we have a natural intuition of the factors of variation composing the world around us. This allows us to readily generalize outside our immediate experience to new domains. For example, we recognize dogs even when they come in breeds, colors and poses that we have never encountered before (Fig. \ref{fig:figConceptual}A). However, state-of-the-art deep learning models do not have the same intuitions \citep{alcorn2019strike}. While models effectively capture statistical patterns among pixels and labels, they lack a sense of the natural factors of variation of the data. An exciting direction is thus to create models that learn to uncover such natural factors of variation, overcoming the limitations of existing models. Learning the underlying factors of variation would also unlock the potential for better generalisation from fewer data points and stronger interpretability, and would be a step towards building fairer models \citep{pmlr-v97-creager19a,LocatelloFairness}.\\
\\
%: for example, a school bus toppled over is misclassified with high confidence as a snow plow 
% Learning disentangled representations is arguably key to build robust, fair, and interpretable ML systems \citep{bengio_representation_2013,A bus may be toppled over, have faded color (if it's older), or appear smaller depending on the perspective---illustrating some of the natural variations for a bus. lake_building_2017, locatello_fairness_2019}. However, it remains unclear how to achieve disentanglement in practice. Current approaches aim to map different factors of variations in the data to distinct subspaces of a latent representation, but have achieved only limited empirical success \citep{higgins_beta-vae_2016, burgess_understanding_2018}. More work on the theoretical foundations of disentanglement could provide the key to the development of more successful approaches. as generative factors of the objects that compose the world around us, and represent them using deep learning models,
% However, current approaches have achieved only limited empirical success \citep{higgins_beta-vae_2016, burgess_understanding_2018}.
A popular solution to learn underlying factors of variation of the data is disentanglement \citep{bengio_representation_2013, lake_building_2017, locatello_fairness_2019}. In its original formulation, disentanglement consists in isolating statistically independent factors of variation in data into independent latent dimensions. For example, disentanglement might encode a dataset composed of dog images into three separated latent subspaces corresponding respectively to pose, color, and breed (see Fig. \ref{fig:figConceptual}B).
% Thus, learning disentangled representations is arguably key to build robust, fair, and interpretable ML systems \citep{bengio_representation_2013, lake_building_2017, locatello_fairness_2019} \MI{there's some repetition with the last sentence of the first paragraph. Should we consolidate the two?} \DB{Indeed, but the other sentence does not mention disentanglement yet}. 
A range of theoretical studies investigate the conditions under which these factors are identifiable  \citep{locatello_challenging_2019, shu_weakly_2020,locatello_weakly-supervised_2020, hauberg_only_2019,khemakhem_variational_2020}. More recently, \citet{higgins_towards_2018} has proposed an alternative perspective connecting disentanglement to group theory (see Appendix \ref{sec:equiv} for a primer on group theory). In this framework, the factors of variation are different subgroups acting on the dataset, and the goal is to learn representations where separated subspaces are equivariant to distinct subgroups. This is a promising formalism because many transformations found in the physical world are captured by group structures \citep{noether_finiteness_1915}. However, the fundamental principles for how to design models capable of learning such equivariances remain to be discovered (but see \citet{caselles-dupre_symmetry-based_2019}).\\
\\

Existing works have proven successful at discovering the factors of variation in rigid datasets such as a single MNIST digit or a single object class such as a chair \citep{higgins_beta-vae_2016, burgess_understanding_2018}. However, in this work we find these models perform poorly when scaled to multiple classes of objects. Thus, we turn to a theoretical analysis and we attack the problem of disentanglement through the lens of topology \citep{munkres_topology_2014}. We show that for a very broad class of transformations acting on images---encompassing all affine transformations (e.g. translations, rotations), an encoder that would map these transformations into dedicated latent subspaces would necessarily be discontinuous. Thus, we turn to a different view on disentanglement. Rather than restrict each transformation to part of a representation, what if we allow instead a transformation to modify the entire representation? With this intuition, we reframe disentanglement by distinguishing its objective from its traditional implementation, resolving the discontinuities of the encoder. We derive an alternative definition of disentanglement that relies on the learning of an equivariant model equipped with \emph{distributed operators}. Such distributed operators can potentially act on the full latent code, and we enforce that each operator corresponds to a single factor of variation (see Fig. \ref{fig:figConceptual}C). Guided by classical results from group representation theory \citep{Scott:and:Serre:96}, we then propose a proof-of-concept model, equipped with such operators, which is theoretically guaranteed to satisfy our alternative definition of disentanglement. We empirically demonstrate its ability to disentangle a range of affine image transformations including translations, rotations and combinations thereof. 

\section{Empirical Limitations of Traditional Disentanglement}
\label{LimitationsDisentanglement}
%
%We first consider the MNIST dataset of handwritten digits, augmented with either image rotation or translation (see Appendix \ref{sec:datagen} for dataset details). We train different models commonly used for disentanglement, namely a VAE (\citet{kingma_auto-encoding_2014}), a beta-VAE (\citet{higgins_beta-vae_2016}), and a CCI-VAE (\citet{burgess_understanding_2018}) and explore qualitatively and quantitatively whether these models are able to learn a disentangled representation.%, in which the transformation affects only a subspace of the latent representation. \MI{We can remove the Appendix reference here since we reference Appendix B in the paragraph below when we introduce the first experiment}

\begin{figure}[t!]
\begin{center}
\includegraphics[width=14cm]{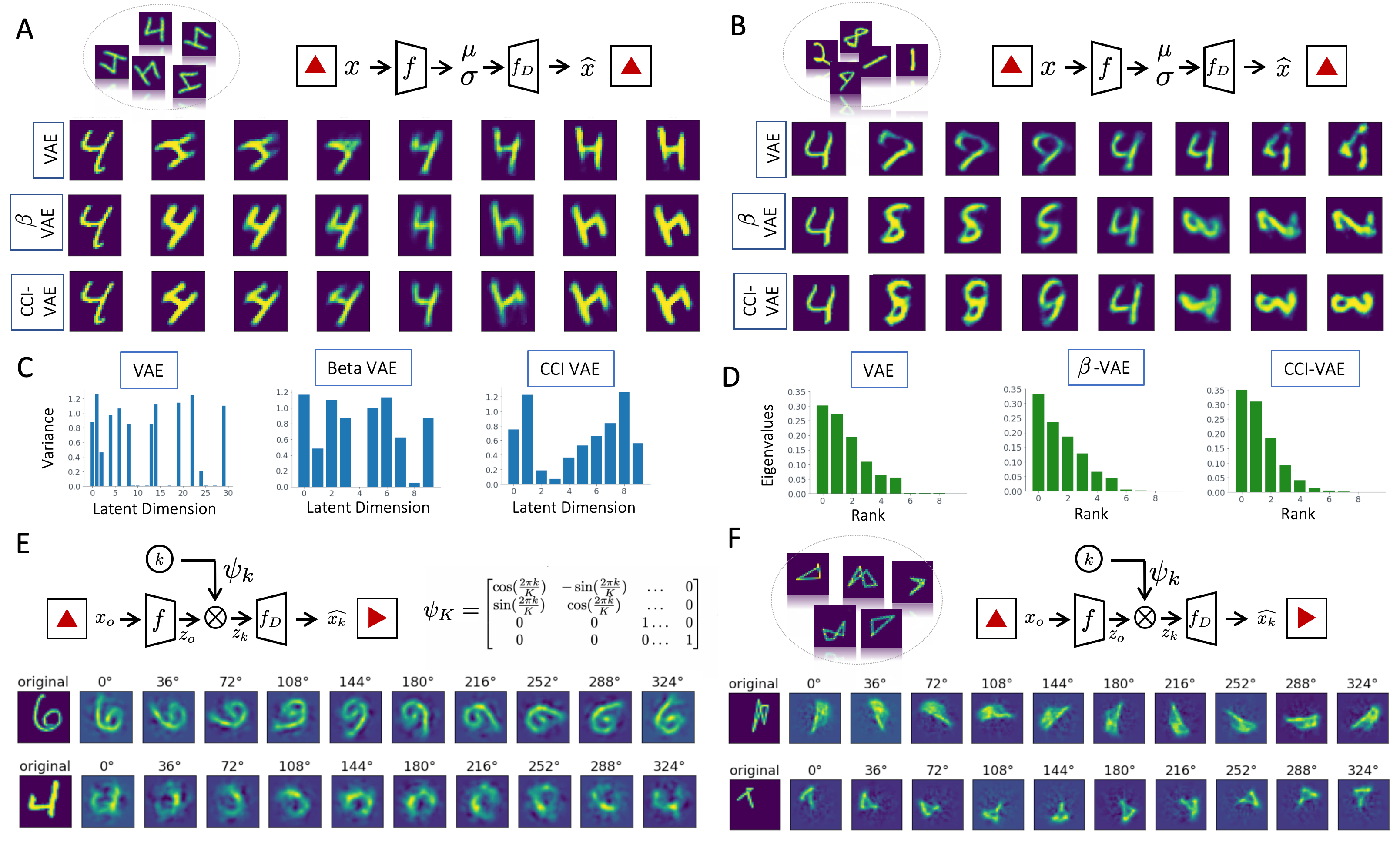}
\end{center}
\caption{\textbf{Failure modes of common disentanglement approaches.} \textbf{A.} Latent traversal best capturing rotation for a VAE, $\beta$-VAE, and CCI-VAE for rotated MNIST restricted to a single digit class ("4"). \textbf{B.} Same as panel A for all 10 MNIST classes. \textbf{C.} Variance of single latents in response to image rotation, averaged over many test images. \textbf{D.} Ranked eigenvalues of the latent covariance matrix in response to image rotation, averaged over many test images. \textbf{E.} A supervised disentangling model successfully reconstructs some digits (top) but fails on other examples (bottom). \textbf{F.} Failure cases of the supervised model trained on a dataset of 2000 rotated shapes (see also Fig. \ref{fig:appendixdisentangled}).}
\label{fig:fig1}
\end{figure}
In this section we empirically explore the limitations of traditional disentanglement approaches, in both unsupervised (variational autoencoder and variants) and supervised settings.
\paragraph{VAE, beta-VAE and CCI-VAE} We show that, consistent with results from prior literature, a variational autoencoder model (VAE) and its variants are successful at disentangling the factors of variation on a simple dataset. We train a VAE, beta-VAE and CCI-VAE \citep{kingma_auto-encoding_2014,higgins_beta-vae_2016,burgess_understanding_2018} on a dataset composed of a single class of MNIST digits (the ``4s"), augmented with $10$ evenly spaced rotations (all details of the models and datasets are in App. \ref{sec:experimentalDetails}). After training, we qualitatively assess the success of the models to disentangle the rotation transformation through traditional latent traversals: we feed an image of the test set to the network and obtain its corresponding latent representation. We then sweep a range of values for each latent dimension while freezing the other dimensions, obtaining a sequence of image reconstructions for each of these sweeps. We present in Fig. \ref{fig:fig1}A examples of latent traversals along a single latent dimension, selected to be visually closest to a rotation (see Fig. \ref{fig:appendixVAEsingle} for latent traversals along all other latent dimensions). We find that all these models are mostly successful at the task of disentangling rotation for this simple dataset, in the sense that a sweep along a single dimension of the latent maps to diverse orientations of the test image.\\ %However, one might notice that the reconstructed digit is also slightly deformed along the traversal, which hints at an imperfect disentanglement of shape versus rotation angle. We also repeat these experiments with a translation augmentation instead of rotation with similar results (@@@APPENDIX FIG). Consistent with results from prior literature (\cite{burgess_understanding_2018}, \cite{higgins_beta-vae_2016}),
\\
We then show that on a slightly richer dataset, MNIST with all digits classes, a VAE model and its variants fail to disentangle shape from pose. We train all three models studied (VAE, beta-VAE, CCI-VAE) on MNIST augmented with rotation, and find that all these models fail to disentangle rotation from other shape-related factors of variation (see Fig. \ref{fig:fig1}B for the most visually compelling sweep and Fig. \ref{fig:appendixVAEmnist} for sweeps along all latent dimensions). We further quantify the failure of disentanglement by measuring the variance along each latent in response to a digit rotation, averaged over many digits (see Fig. \ref{fig:fig1}C and details of analysis in App. \ref{sec:addresults}). We find that the information about the transformation is distributed across the latents, in contradiction with the conventional notion of disentanglement. One possibility would be that the direction of variance is confined to a subspace, but that this subspace is not aligned with any single latent. In order to discard this possibility, we carry a PCA-based analysis on the latent representation (Fig. \ref{fig:fig1}D and App. \ref{sec:addresults}) and we show that the variance in latent representation corresponding to image rotation is not confined to a low-dimensional subspace.%\MI{We likely only the need the second reference to Appendix D.1 since it describes both the variance and PCA analysis.}\\  %We repeat these results with translation (@@@APPENDIX FIG). %\subsection{Supervised Disentanglement} 
\paragraph{Supervised Disentanglement} We further explore the limitations of traditional disentanglement in a supervised framework. We train an autoencoder on pairs of input and target digit images (Fig. \ref{fig:fig1}E), where the target image is a rotated version of the input image with a discrete rotation angle indexed by an integer value $k$. The input image is fed into the encoder to produce a latent representation. This latent representation is then multiplied by a matrix operator $\psi_k$, parameterized by the known transformation parameter $k$. This matrix operator, which we call the \emph{disentangled operator}, is composed of a 2-by-2 diagonal block with a rotation matrix and an identity matrix along the other dimensions (shown in Fig. \ref{fig:fig1}E). The disentangled operator (i) is consistent with the cyclic structure of the group of rotations and (ii) only operates on the first two latent dimensions, ensuring all other dimensions are invariant to the application of the operator. The transformed latent is then decoded and compared to the target image using an L2 loss (in addition, the untransformed latent is decoded and compared to the original image for regularization purposes). The only trainable parameters are the encoder and decoder weights. We use the same architecture for the encoder and decoder of this model that we use for the VAE models in the previous section. This supervised disentanglement model partly succeeds in mapping rotation to a single latent on rotated MNIST (Fig.  \ref{fig:fig1}E top row). However, there remains some digits for which disentanglement fails (Fig.  \ref{fig:fig1}E bottom row). To further expose the limitations of this model, we design a custom dataset composed of 2000 simple shapes in all possible orientations. When trained on this extensive dataset, we find that the model fails to capture rotations on many shapes. Instead, it replaces the shape of the input image with a mismatched stereotypical shape (Fig.  \ref{fig:fig1}F). We reproduce all these results with translation in the appendix (Fig. 9-13).\\ %Using multiple controls, we show in App. \ref{sec:addresults} that the failure of the model can only partly be explained by the complexity of the encoder and decoder architecture.  
% It is difficult to evaluate the capacity of the model to learn to rotate many different images with MNIST, because MNIST is only composed of 10 classes of shapes corresponding to the 10 different digits.
\\
In conclusion, we find that common disentanglement methods are limited in their ability to disentangle pose from shape in a relatively simple dataset, even with strong supervision (see also \citet{locatello_challenging_2019}). We cannot empirically discard the possibility that a larger model, trained for longer on even more examples of transformed shapes, could eventually learn to disentangle pose from shape. However, in the next section we will prove, using arguments from topology, that under the current definition of disentanglement, an autoencoder cannot possibly learn a perfect disentangled representation for all poses and shapes.  In Sec. \ref{sec:shiftopepresentation} and Sec. \ref{sec:shiftopeexpe}, we will show that another type of model---inspired by group representation theory---can properly disentangle pose from shape.
\section{Reframing Disentanglement}
\label{ReframingDisentanglement}
In this section, we formally prove that traditional disentanglement by a continuous encoder is mathematically impossible for a large family of transformations, including all affine transformations. We then provide a more flexible definition of disentanglement that does not suffer from the same theoretical issues.%, while preserving the benefits of traditional disentanglement.
%We prove that a practical consequence of this mathematical fact is that an encoder approximating a disentangled solution would need to be discontinuous in the neighborhood of all images that present a symmetry with respect to the transformation to be disentangled. We then provide an more general definition of disentanglement which does not suffer from the same theoretical issues, while preserving the benefits of traditional disentanglement. We show that this alternative definition of disentanglement enables an autoencoder network to disentangle rotation from shape.
\subsection{Mathematical Impossibility of Traditional Disentanglement}
\begin{figure}[t!]
\begin{center}
\includegraphics[width=14cm]{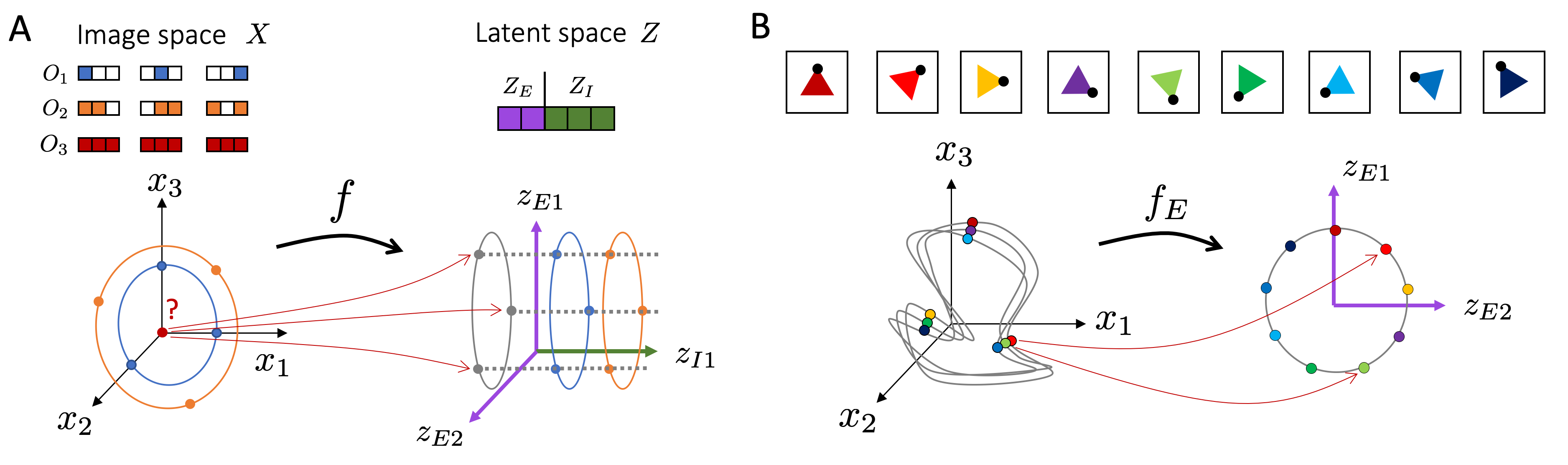}
\end{center}
\caption{\textbf{Visual proof of the topological defects of disentanglement.} \textbf{A}. Top left: $O_1$, $O_2$ and $O_3$ are three examples of orbits of 3-pixel-images transformed by translation. Bottom: (left) orbits visualized in image space (points constitute the orbits, continuous lines are for visualization purposes); (right) orbits in latent space. When projected onto the equivariant subspace $Z_E$ (gray dotted lines), all orbits should collapse onto each other. Yet the orbit of a uniformly black image (red dot) contains a single point and thus cannot be mapped onto the other orbits. \textbf{B}. Discontinuity of $f_E$ around symmetric images. Top: consider an image of an equilateral triangle, with an infinitesimal perturbation on one corner (black dot), undergoing rotation (color changes are for visualisation purposes). Bottom: (left) after a rotation of 120$^\circ$, the orbit in image space (here projected onto 3 dimensions for visualisation) almost loops back on itself; (right) in $Z_E$, each angle of rotation corresponds to a distinct point in space. Therefore, the encoder $f_E$ is discontinuous (as shown by the red arrows).}
\label{fig:fig2}
% \vspace{-6mm}
\end{figure}

We first consider a simple example case where disentanglement is impossible. We consider the space of all images of 3 pixels $X = \mathbb{R}^3$, and the transformation acting on this space to be the group of integer finite translations, assuming periodic boundary conditions of the image in order to satisfy the group axiom of invertibility (Fig. \ref{fig:fig2}A, see App. \ref{sec:equiv} for definitions).  Given an image, the set of images resulting from the application of all possible translations to this image is called the \emph{orbit} of this image. We note that the space of images $\mathbb{R}^3$ is composed of an infinite set of disjoint orbits. Can we find an encoder $f$ which maps every point of image space $X$ to a disentangled space $Z$? \\
\\
To conform to the conventional definition of disentanglement \citep{higgins_towards_2018} (see App. \ref{sec:topo_proof} for a formal definition), $Z$ should be composed of two subspaces, namely (i) an equivariant subspace $Z_E$ containing \emph{all and only} the information about the transformation (i.e. location along the orbit) and (ii) an invariant subspace $Z_I$, invariant to the transformation but containing all other information about the image (i.e. identity of the orbit).
%\end{itemize}
%\begin{itemize}[noitemsep,topsep=0pt]
%    \item an equivariant subspace $Z_E$ containing \emph{all and only} the information about the transformation (i.e. location along the orbit)
%    \item an invariant subspace $Z_I$ invariant to the transformation but containing all other information about the image (i.e. identity of the orbit)
%\end{itemize}
That is, the latent space is disentangled with respect to the transformation.
Each orbit should thus lie in a plane parallel to $Z_E$ (otherwise some information about the transformation would leak into $Z_I$), and all orbits projected onto $Z_E$ should map onto each other (otherwise some information about the identity of the orbit would leak into $Z_E$). We now consider the orbit containing the black image [0,0,0]. Since all translations of the black image are the black image itself, this orbit contains only one point. And yet, the image of this orbit in $Z_E$ should conform to the image of other orbits, which generally consist of 3 distinct points. Since a function cannot map a single point to 3 points, an encoder $f$ ensuring disentanglement for all images cannot exist.\\
\\
Using similar topological arguments, we formally prove the following theorem (App. \ref{sec:topo_proof1}), generalizing the observation above to a large family of transformations including translations, rotations and scalings.\\
\\
\textbf{Theorem 1: }\emph{Disentanglement into subspaces by a continuous encoder is impossible for any finite group acting on Euclidean space $\mathbb{R}^N$.} 

\subsection{Practical Examples of Topological Defects}

The formal theorem of the previous section does not tell us how hard it would be to \emph{approximate} disentanglement in practice. We show next that a disentangling encoder $f$ would need to be \emph{discontinuous} around \emph{all images that present a symmetry with respect to the transformation}, which makes this function very discontinuous in practice. As an example, we consider the image of an equilateral triangle undergoing rotation (Fig. \ref{fig:fig2}B, color changes are for visualisation purposes). Due to the symmetries of the triangle, a rotation of 120$^{\circ}$ of this image returns the image itself. Now we consider the same image with an infinitesimal perturbation on one corner of the triangle, breaking the symmetry of the image.  A rotation of 120$^{\circ}$ of this perturbed image returns an image that is infinitesimally close to the original image. And yet the equivariant part of the encoder $f_E$ (i.e. the projection of $f$ onto the equivariant subspace $Z_E$) should map these two images to disjoint points in the equivariant subspace $Z_E$, in order to properly encode the rotation transformation. Generalizing this argument to all symmetric images, we see that a disentangling encoder would be discontinuous in the neighborhood of all images that present a symmetry with respect to the transformation to disentangle. This is incompatible with most deep learning frameworks, where the encoder is usually a neural network implementing a continuous function. We provide a formal proof of the discontinuity of $f_E$ in App. \ref{sec:topo_proof2}. The invariant encoder $f_I$ (i.e. the projection of $f$ onto the invariant subspace $Z_I$) also presents topological defects around symmetric images. We provide both a visual proof and a formal proof of these defects in App. \ref{sec:topo_proof2}. 
% Older Section
\subsection{A More Flexible Definition of Disentanglement}
%Can we provide a more flexible definition of disentanglement, and in the next section we formally prove that a model can learn to disentangle finite group actions on images using this new definition. \\%Here is the definition of disentanglement proposed by @@@REF HIGGINS: %\emph{"A vector representation is called a disentangled representation with respect to a particular decomposition of a symmetry group into subgroups, if it decomposes into independent subspaces, where each subspace is affected by the action of a single subgroup, and the actions of all other subgroups leave the subspace unaffected."}

An underlying assumption behind the traditional definition of disentanglement is that the data is naturally acted upon by a set of transformations that are orthogonal to each other, and that modify well-separated aspects of the data samples.  However, in many cases this separation between factors of variation of the data is not possible (as also noted by \citet{higgins_towards_2018}). We notice that the current definition of disentanglement unnecessarily conflates the objective of isolating factors of variation with the specific algorithm which consist in mapping these factors into distinct subspaces of the internal representation. In order to build a model that respects the structure of the data and its transformations, the latent space should instead preserve the entanglement between factors of variation that are not independent. 
%We take a step back from this traditional view, and consider that what we want is simply for the model to be equivariant to the group of transformations acting on the input, with a \emph{known} equivariant operator acting on latent space \MI{instead of take "a step back" which is colloquial, we can say something like, "we propose a new view of disentanglement as..."} This leads to a new, more general, definition of disentanglement:\\

% Consequently, we connect the goal of disentanglement to that of learning equivariant latent operators. 
A model equipped with a latent operator is equivariant to a transformation if encoding a sample then applying the latent operator is equivalent to transforming the sample first then encoding it. Formally, we say a that a mapping $f: X \rightarrow Z$ is equivariant to a transformation $g_k$ with parameter $k$ if for any input $x \in X$
\begin{equation}
     f(\phi_k(x)) = \psi_k(f(x)), \forall k \in K,
     \label{eq:maintextequi}
\end{equation}
where $K$ is the space of transformation parameters. The operators $\phi_k$ and $\psi_k$ capture how the transformation $g_k$ acts on the input space and representation space respectively. With this view, we turn to a definition of disentanglement in which the transformations are modelled as \emph{distributed} operators (i.e. not restricted to a subspace) in the latent space. In this new definition, a model's representation is said to be disentangled with respect to a set of transformations, if the model is equipped with a family of operators, potentially acting on the entire representation, where each operator corresponds to the action of a single transformation and the resulting model is equivariant. Formally:

\begin{definition}
    \label{defDisentanglement}

We consider a transformation group G which is exactly factorized into subgroups $G_1, \dots, G_n$ (i.e. G is the Zappa-Szep product of the subgroups). Formally, 

$G = G_1 G_2 \dots G_n = \{g_1 g_2 \dots g_n | g_i \in G_i  \text{ for } i = 1, ..., n\}$

with $G_i \cap G_j = \{e\}$ for all $i \neq j$. 

A model's representation $Z$ built from a map $f: X\rightarrow Z$ is disentangled with respect to $G$ if: 

%$\cdot : G \times Z \rightarrow Z$.
\begin{itemize}
%In the special case of linear disentanglement, each subgroup action admits a linear representation $\rho_i$.
\item The model includes operators $\psi_i$ where each $\psi_i$ corresponds to a subgroup acting on the entire latent representation vector. %In the special case of linear disentanglement, each $\psi_i$ is a matrix and the $\psi_i$ must correspond to one and only of the $\rho_j$.
\item The map $f: X\rightarrow Z$ is equivariant between the actions on $X$ and $Z$.
\end{itemize}
\end{definition}

%In the special case of linear disentanglement, each subgroup action admits a linear representation $\rho_i$. 
In the special case of linear disentanglement, each $\psi_i$ is a matrix operator acting on the latent space by multiplication. %and the $\psi_i$ must correspond to one and only of the $\rho_j$ \MI{combine into one sentence without repeating "In the special case of..."}.

%These operators are known in the sense that they have an explicit form, thus allowing the user to manipulate the latent representation by applying the operator. 

This definition, more flexible than traditional disentanglement in the choice of the latent operators, obeys to the same \emph{desiderata}  of identification and isolation of the factors of variations present in the data. Note that because we do not restrict each operator to act on a subset of the latent dimensions, we handle cases where the different subgroups composing the product do not commute. In contrast, \citet{higgins_towards_2018} propose a definition of disentanglement that requires each subspace of the representation to be acted on by a single subgroup, thus only handling commutative subgroups. 

%If we set ourselves in a case where both the encoder and decoder are linear and invertible, group representation theory provides us with a way to build an equivariant, distributed operator that preserves not only the group structure but also how it interacts with the data at hand.\\
%\\
%Here we suggest an alternative definition of disentanglement, which relaxes the constraints on algorithmic implementation. In the next section we show that this new definition allows us to preserve the topology of affine transformations on $\mathbb{R}^n$.
% We have demonstrated in @@@SECTION that the traditional subspace operator cannot disentangle affine transformations from shape. \\ 
%A finite cyclic group $G$ is such that each $g_k \in G$ writes as $g_k=g_0^k$ ($g_0$ is called the generator of the group), and there exists $K$ such that $g_0^K=e_G.$ See Appendix \ref{sec:equiv} for details.}. 

\subsection{The Shift Operator for Affine Transformations}
\label{sec:shiftopepresentation}
The flexibility of Def.\ref{defDisentanglement} unlocks a powerful toolbox for understanding and building disentangled models. Group transformations are transformations with a specific structure allowing to (1) undo a transformation (invertibility), (2) leave a sample unchanged (identity), and (3) decompose the transformation  applied into elementary bricks (associativity) (see App. \ref{sec:equiv} for a formal definition). Not all transformations present in datasets satisfy the requirements of a group. For example, 3D rotations of objects in 2D images, which introduce occlusion, are not invertible. Nevertheless, many transformations (including common 2D affine transformations) respect the properties of a group, and here we focus on such transformations. \\
%With a group structure, we can decompose transformations into subgroups, describe how to represent transformations as matrices, and build flexible disentangled models.\\
\\
Using classical results from the linear representation of finite groups \citep{Scott:and:Serre:96}, we show that a carefully chosen distributed operator in latent space---the shift operator $\psi_k$---is linearly isomorphic to specific transformations that include integer pixel translations and rotations. This means that with this operator, we can learn a latent space equivariant to any affine transformation using a simple \emph{linear} autoencoder.

Consider a linear encoder model $f=W$. According to our definition of disentanglement, $W$ should be an equivariant invertible linear mapping between $X$ and $Z$:
\begin{equation}
W~\phi_k(x)  = \psi_k(W~x)~~\forall x\in X(=\mathbb{R}^N), \forall k \in K
\label{eq:eqmappMAINTEXT}
\end{equation}
where $\phi_k$ and $\psi_k$ are the representations of $g_k \in G$ on the image and latent space respectively. We assume the following additional properties on the group $G$ and its representation $\phi$: (i) $G$ is cyclic of order $K$ with generator $g_0$ and (ii) $\phi$ is isomorphic to the regular representation of $G$ (see \citet{Scott:and:Serre:96}). These properties are respected by all cyclic linear transformations of finite order $K$ of the images that leave no pixel in place (see App. \ref{sec:equiv} for definitions), such as integer pixel translation with periodic boundary conditions, or discrete image rotations.

For $W$ to be equivariant, eqn. \ref{eq:eqmappMAINTEXT} must be true for every image $x$. Consequently, the two representations $\psi_k$ and $\phi_k$ are isomorphic:
\begin{equation}
\forall k \in K,  \phi_k  =W^{-1} ~\psi_k ~ W
\end{equation}

Two representations are isomorphic if and only if they have the same character \citep[Theorem 4, Corollary 2]{Scott:and:Serre:96} (see App. \ref{sec:equiv} for a definition of characters). We thus need to choose $\psi$ such that it preserves the character of the representation $\phi$ corresponding to the action of $G$ on the dataset of images. \\
%Importantly, we will see that our proposed operator needs to be \emph{distributed} in the sense that it should act on the full latent code. Given that the encoder and decoder are linear and invertible, the two representations $\phi$ and $\psi$ must be isomorphic.\\the  as a representation of the group's action on the latent space. For each $g_k \in G$ its corresponding shift operator is the block diagonal matrix of order $N$ composed of $\frac{N}{K}$ repetition of $M^k$.
\\
Let us consider the matrix $M^k$ of order $K=|G|$ that corresponds to a shift of elements in a $K$-dimensional vector by $k$ positions. We construct the \emph{shift operator} from $M^k$ as:
\begin{small}
\noindent\begin{minipage}{.5\textwidth}
\begin{equation}
M^k:=\begin{bmatrix} 
    0      & 0 & \ldots  & 1 \\
    1      & 0 & \ldots & 0 \\
    0      & 1 &    0   & \vdots\\
    \vdots &   &      & \\
    0      &  \ldots &  1     & 0
\end{bmatrix}^{\scalebox{1.5}{$k$}}
\end{equation}
\end{minipage}%
\begin{minipage}{.5\textwidth}
\begin{equation}
  \psi_k:=\begin{tikzpicture}[decoration={brace,amplitude=5pt},baseline=(current bounding box.west)]
     \matrix (magic) [matrix of math nodes,left delimiter={[},right delimiter={]}] {
        M^k\\
        & M^k\\
        && ...\\
        &&& M^k\\
     };
  \end{tikzpicture}
\end{equation}
\end{minipage}
\end{small}
% Let us compute the character table of this representation. First, for the identity, is it trivial to see that $\chi_\psi(e)=N$. Second, for any $g_k\neq e$, we have 
% \begin{equation}
%     \chi_\psi(g_k) = Tr(\psi_k) = 0
% \end{equation}
% since all diagonal elements of $g_k$ will be $0$. 
% Therefore, the character table of the shift operator is the same as $\chi_\phi$:\\
%\footnote{When we will consider multiple transformations, the resulting group will be non-cyclic but each subgroup is cyclic.}

% The second property might seem counter-intuitive, but it just means that the transformation leaves no pixel unchanged (i.e. permutes all the pixel). In the case of rotations, this is approximately true since only the rotation origin remains in place.
% The character table of $\phi$, given the second property, is 
% {
% \centering
% \begin{tabular}{l|c|c|c|c|c|c}
%   & e &$g_0$ & $g_0^2$ & ... & $g_0^{K-1}$\\
% $\chi_\phi$ & $N$ & $0$ & $0$  & $0$  & $0$\\
% \end{tabular}
% \par}
%$\forall k \in 1, \ldots, K$
% {
% \centering
% \begin{tabular}{l|c|c|c|c|c|c}
%   & e &$g_0$ & $g_0^2$ & ... & $g_0^{K-1}$\\
% $\chi_\psi$ & $N$ & $0$ & $0$  & $0$  & $0$\\
% \end{tabular}
% \par}
For each $g_k \in G$ the corresponding shift operator is the block diagonal matrix of order $N$ composed of $\frac{N}{K}$ repetition of $M^k$. We show in App. \ref{sec:charactheory} that the shift operator has the same character as the representation $\phi$ corresponding to the action of $G$ on the dataset of images. Thus, the two are isomorphic and using this shift operator ensures that an equivariant invertible linear mapping $W$ exists between image space and the latent space equipped with the shift operator. This shift operator can also be replaced by a complex diagonal operator, which is more computationally efficient to multiply with the latent (App. \ref{sec:complexshift}).
% Note that the character of the \emph{disentangled operator} does not match the character table of $\phi$, and so \emph{we verify once again in this linear autoencoder setting that the disentangled operator is unfit to be equivariant to affine transformations of images}. 
\\~
\\
The shift operator can thus be used to represent any finite cyclic group of typical affine transformations, such as rotation, translation in x, translation in y. The role of the encoder is to construct a latent space where transformations can be represented as shifts. Importantly, the shift operator model \textbf{does not require knowledge of the transformation in advance}, only the cycle order of each group. Note that the shift operator only handles cyclic groups with finite order (or a product of such groups). In order to tackle continuous transformations, a discretisation step could be added, but we leave the exploration of this extension for future work. In the next section, we show how these theoretical results can lead to practical and effective disentangling models for simple affine transformations such as rotations and translations.
\section{Illustration of Distributed Disentanglement in Practice}
\label{sec:shiftopeexpe}
Our empirical (Sec. 2) and theoretical (Sec. 3) findings converge to show the difficulties of disentangling even simple affine transformations into distinct subspaces. Distributed latent operators present a promising alternative by overcoming some of the limitations of traditional disentanglement. Here, on a series of toy problems, we show how distributed latent operators can successfully learn to disentangle cyclic affine transformations, even in cases where traditional disentanglement struggles.

\subsection{The Supervised Shift Operator Model}

\begin{figure}[t!]
\begin{center}
\includegraphics[width=14cm]{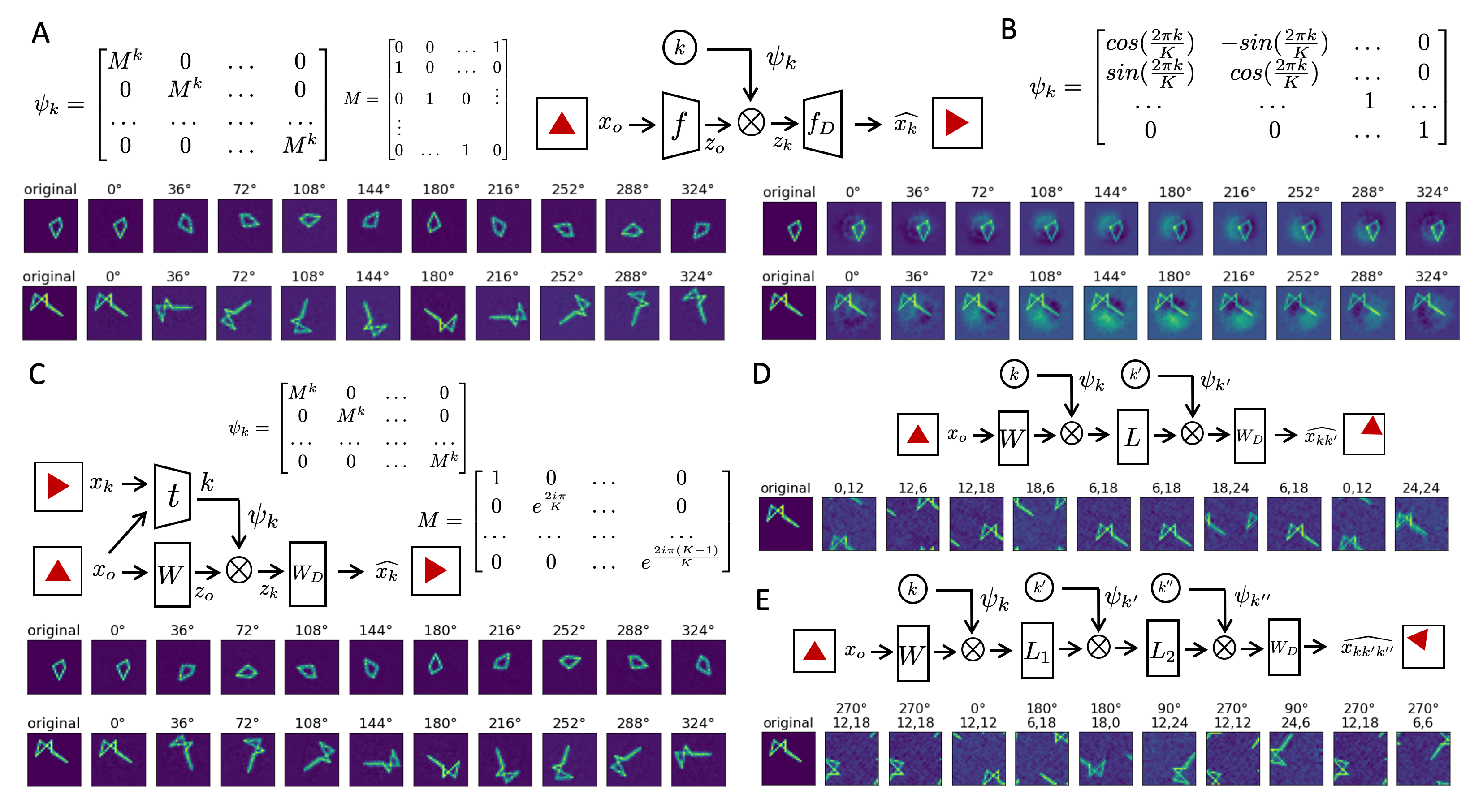}
%\framebox[4.0in]{$\;$}
\end{center}
\caption{\textbf{Success and flexibility of proposed distributed shift operator models}: \textbf{A.} Proposed shift operator model successfully learns rotation on simple shapes. \textbf{B.} Disentangled operator fails to learn rotation.  \textbf{C.} Weakly supervised shift operator model, using the complex version of the shift operator, successfully rotates simple shapes. Note that the model maps ground-truth counter clockwise rotations to clockwise rotations, while respecting the cyclic structure of the group. \textbf{D.} Stacked shift operator model succeeds on conjunction of translations. \textbf{E.} Stacked shift operator model succeeds on conjunction of translations and rotations. Numbers above plots indicate rotation angle and/or translation in $x$ and $y$ respectively.}
% \vspace{-4mm}
\label{fig:fig3}
\end{figure}

Guided by our theoretical results, we train a supervised non-variational autoencoder using pairs of samples and their transformed version (with a known transformation indexed by $k$) using the distributed shift operator from Sec. \ref{sec:shiftopepresentation} (shown in Fig. \ref{fig:fig3}A) instead of the disentangled operator from Sec. 2. We feed the original sample $x$ to a linear invertible encoder that produces a latent representation. The latent representation is then multiplied by the shift operator matrix parametrized by $k$. The transformed latent is then decoded and L2 loss between the two reconstructions ($x$ and its transformed version) and their respective ground-truth images is back-propagated.\\
%(or quasi-invertible, see App. \ref{autoencoderDetails})
\\
As predicted by character theory, our proposed model is able to correctly structure the latent space, such that applying the shift operator to the latent code at test time emulates the learned transformation (see Fig. \ref{fig:fig3}A, test MSE reported in Table \ref{tab:mse}). Consistent with the theory, the same linear autoencoder equipped with a disentangled operator fails at learning the transformation (Fig. \ref{fig:fig3}B). We reproduce these results for translation in App. \ref{sec:addresults}. We thus show that, unlike prior approaches, our theoretically motivated model is able to disentangle affine transformations from examples. 
%and the LSBD disentanglement measure from \citet{anonymous2021quantifying} and reported in App. \ref{sec:quantresults}.1
%Having assessed the relevance of the proposed real shift operator model,
%Interestingly, test MSE for rotations are lower than for translations. We believe this is due to the fact that in the case of translations, changes in the image induced by each transformation are less visually striking than with rotations.
    
\subsection{The Weakly Supervised Shift Operator Model}
\label{sec:weakshift}

Here we show that our method can also learn to disentangle transformations in a weakly supervised setting where the model is not given the transformation parameter between pairs of transformed images (e.g. rotation angle) during training. We consider the case of a single transformation for simplicity. We encode samples by pairs $x_1,x_2$ (with $x_2$ a transformed version of $x_1$) into $z_1$ and $z_2$ respectively, and use a phase correlation technique to identify the shift between $z_1$ and $z_2$, as in \citet{Reddy1996}. An L2 loss is applied on reconstructed samples, with the original sample transformed according to all possible $k$ and weighted according to soft-max scores given by the cross-correlation method (see App. \ref{sec:appendixReddy} for details). Here and in the remainder of the experiments, we use the complex version of the shift operator for computational efficiency (shown in Fig. \ref{fig:fig3}C). This weakly supervised version of the model has an extra free parameter which is the number of latent transformations, not known a priori. Let us denote $K_L$ this number, which can be different than the ground-truth order of the group $K$. We explore the effect of different $K_L$ in App. \ref{sec:appendixReddy}. The results of this model (with $10$ latent transformations $K_L$) are shown in Fig. \ref{fig:fig3}C. The weakly supervised shift operator model works almost as well as its supervised counterpart, and this is confirmed by test MSE (see Table \ref{tab:mse}). The same model can successfully be trained on MNIST digits (Fig. \ref{fig:appendixShiftMNIST}). Finally, we highlight that distributed operators successfully disentangle rotations, translations, and their combinations using only a single-layer linear encoder. 
%Note that the model did not learn to match each rotation to its ground-truth value of the angle in degrees, but the cyclic structure of the group is respected.

\subsection{Multiple transformations: Stacking Shift Operators}

%Namely, we showed our method covers the case of either (i) translations in x-axis (ii) translations in y-axis (iii) rotations.
So far, we have only considered the case of a single type of transformation at a time.  When working with real images, there are more than one type of transformation jointly acting on the data, for example a rotation followed by a translation. Here we show how we can adapt our proposed shift operator model to the case of multiple transformations.
\paragraph{Stacked shift operator model} In the case of multiple transformations, a group element is a composition of consecutive single transformations. For example, elements of the Special Euclidean group (i.e. translations and rotations) are a composition of the form $a_y~a_x~h$, where $a_x$ is an element of the $x$-translation subgroup, $a_y$ of the $y$-translation subgroup, and $h$ of the image rotation subgroup. Our theory in Sec. \ref{sec:charactheory} ensures that each of these subgroup's action in image space is linearly isomorphic to the repeated regular representation $\psi$. We can thus match the structure of the learning problem by simply stacking linear layers and operators. We build a stacked version of our shift operator model, in which we use one complex shift operator for each type of transformation, and apply these operators to the latent code in a sequential manner, akin to \cite{tai2019equivariant}. Specifically, consider an image $x \in X$ that encounters a consecutive set of transformations $g_n, g_{n-1}, \ldots g_1$, with $g_i \in G_i~\forall i$.  We encode $x$ into $z$ with a linear invertible encoder $z = W ~ x$. We then apply the operator on the latent space, $\psi_1(g_1)$, corresponding to the representation of $g_1$ on the latent space $Z$. We then apply a linear layer $L_1$ before using the operator corresponding to $G_2$. The resulting latent code after all operators have been applied is $z' = \psi_n(g_n) L_{n-1} \ldots \psi_2(g_2) L_1  \psi_1 z.$ The transformed latent code $z'$ is fed to the linear decoder, in order to produce a reconstruction that will be compared with the ground-truth transformed image $x'$, as in the single transformation case.
% For example, this could be a rotation (and $G_1$ the group of rotations) followed by $1D$ translation on the $x$-axis (the group $G_2$) followed by $1D$ translation on the $y$-axis (the group $G_3$). Generally, we have that $x' = g_n \circ g_{n-1} \ldots \circ g_1 x$.\\
% , or using the linear representation of each group: $x' = \phi_n(g_n) \phi_{n-1}(g_{n-1}) \ldots \phi_1(g_1) x$ where $\phi_i: G_i \rightarrow GL(X)$ is a representation of the group $G_i$ when acting on image space $X$.\\
%, i.e. $\psi_1: G_1 \rightarrow GL(Z)$
% We know each element in the case of multiple transformations is a composition of consecutive single transformations. For example, elements of the Special Euclidean group (i.e. translations and rotations) are a composition of the form $a_y~a_x~r$, where $a_x$ is an element of the $x$-translation subgroup, $a_y$ of the $y$-translation subgroup, and $h$ of the image rotation subgroup. Our theory in \ref{sec:charactheory} ensures that each of these subgroup's action in image space is linearly isomorphic to the repeated regular representation $\phi$. We can thus match the structure of the learning problem by stacking multiple linear layers,  with a transformation $\phi$ acting between each layer. 
% In the example of the Special Euclidean Group, we decompose the group in three subgroups, we need to stack three operator layers $\phi$, and thus two hidden layers $L_1$ and $L_2$. Each linear layer can be seen as re-expressing the image representation in a basis where the transformation corresponding to the target subgroup is the regular representation.
%
\paragraph{Translations in X and Y} Consider the conjunction of translations in $x$ and $y$ axes of the image. This is a finite group of $2$D translations. This group is a direct product of two cyclic groups, and it is abelian (i.e. commutative). We refer the interested reader to App. \ref{sec:appendixdp} for details on direct products. To tackle this case with the stacked shift operator model, we first use the shift operator $ \psi_{x,k,N}$ corresponding to the translation in $x$, then apply a linear layer denoted $L_1$, before using the operator $\psi_{y,k',N}$ corresponding to the translation in $y$:
$z' = \psi_{y,k',N} L_1 \psi_{x,k,N}~z$. We train this stacked model on translated shapes with $5$ integer translations in both $x$ and $y$ axes (i.e. the group order is $25$). Results reported in Fig. \ref{fig:fig3}D show that the stacked shift operator model is able to correctly handle the group of $2$D translations.
\paragraph{Translations and rotations} We consider a discrete and finite version of the Special Euclidean group, where $A$ is a finite group of 2D translations presented in the previous section and $H$ a finite cyclic group of rotations. This group has a semi-direct product structure (see App. \ref{sec:appendixsm} for details) and is non-commutative, contrary to the $2$D translations case. With the stacked shift operators model, we first use the operator $\psi_{h,j,N}$ corresponding to the rotation $h$, then the one for translation in $x$, then the one for $y$-translation. The resulting transformed latent code is $z'=\psi_{y,k',N} L_2 \psi_{x,k,N} L_1 \psi_{h,j,N}~z$. We train this stacked model on discrete rotations followed by integer translations and discrete rotations, using $5$ integer translations in both $x$ and $y$ axes and $4$ rotations. Results reported in Fig. \ref{fig:fig3}E and MSE Table \ref{tab:mse} show that the model is perfectly able to structure the latent space such that the group structure is respected. Additionally, Figures \ref{fig:reconstructionstxty} and \ref{fig:reconstructionsRtxty} show pairs of samples and the reconstructions by the stacked shift operator model in the cases of (i) translation in both x and y axes and (ii) rotations and translations in both axes. In appendix Figure \ref{fig:appendixShapesSM2} we also explore the case where the order of the group of rotations is $5$, breaking the semi-direct product structure (see note in Appendix \ref{sec:se2note}) and show that the stacked shift operator nonetheless performs with great performance. 
\paragraph{Insight from representation theory on the structure of hidden layers} When dealing with multiple transformations (e.g. rotations and translations), we know the form of the operator for every subgroup (shift operator), but we do not know \emph{a priori} the form of the resulting operator for the entire group. In App. \ref{sec:appendixdp} and \ref{sec:appendixsm2} we derive from representation theory the operator for the entire group in the $2$D translation and Special Euclidean group cases and show that they can be built from each subgroup's operator in a non-trivial way. Importantly, we show that the resulting operator for the discrete finite Special Euclidean case has a block matrix form representation based on representations of both translations and rotations. This is expected: this group is non-commutative, so the correct operator cannot be diagonal otherwise two operators corresponding to two elements would commute. Equipped with this theory we can derive insights about the form that intermediate layers should take after training. In particular, we show (App. \ref{sec:insights1}) that the layer $L_1$ should be a block diagonal matrix consisting of repetitions of a permutation matrix that reorders elements in $\psi_{x,k,N}~z$. Similarly, in the case of translations and rotations together, $L_2$ must reorder $\psi_{y,k’,N}$, and $L_1$ be the product of two matrices $L_1=PQ$, where $Q$ is a $N$ by $N$ block diagonal matrix and $P$ is reordering the rows of the vector $Q \psi_{h,j,N} z$ (see App. \ref{sec:insights2}). In future work, we plan to explore the use of these insights to regularize internal layers of stacked shift operator models.
%We show (App. \ref{sec:insights1}) that in order for the product $\psi_{y,k',N} L_1 \psi_{x,k,N}$ applied to $z$ to match the result obtained from representation theory, the layer $L_1$ needs to be a block diagonal matrix consisting of repetitions of a permutation matrix that reorders elements in $\psi_{x,k,N}~z$.
%in order for the product $\psi_{y,k',N} L_2 \psi_{x,k,N} L_1 \psi_{h,j,N}$ applied to $z$ to match the result obtained from representation theory
% (similar to the $2$D translation case)
\section{Discussion}

Finding representations that are equivariant to transformations present in data is a daunting problem with no single solution. A large body of work \citep{cohen_general_2020,cohen_spherical_2018,esteves_polar_2018,greydanus_hamiltonian_2019, romero_attentive_2020,finzi_generalizing_2020,tai2019equivariant} proposes to hard-code equivariances in the neural architecture, which requires \emph{a priori} knowledge of the transformations present in the data. In another line of work, \citet{falorsi_explorations_2018,davidson_hyperspherical_2018, falorsi_reparameterizing_2019} show that the topology of the data manifold should be preserved by the latent representation, but these studies do not address the problem of disentanglement. \citet{higgins_towards_2018} have proposed the framework of disentanglement as equivariance that we build upon here. Our work extends their original contribution in multiple ways. First, we show that traditional disentanglement introduces topological defects (i.e. discontinuities in the encoder), even in the case of simple affine transformations. Second, we conceptually reframe disentanglement, allowing latent operators to act on the entire latent space, so as to resolve these topological defects. Finally, we show that models equipped with such operators successfully learn to disentangle simple affine transformations.

From our perspective, each component of the model---the encoder, the latent operators, and the decoder---plays a role in disentanglement. The encoder restructures information from the input into a latent space amenable to transformation by the latent operators. For example if the latent operators are linear, the encoder restructures information so that linear operators can act on the representations. We note that in cases where not all input information is useful to a downstream task, the encoder can also play the role of removing useless information. The role of the latent operators is to isolate each factor of variation in the latent space. Finally, the decoder combines information from the latent representations and their relationships (described by the latent operators) to solve a downstream task such as reconstruction or classification. A simplified framework could consist in learning operators directly in image space. For example, both \citet{benton2020learning} and \citet{Hashimoto2017} learn the invariances or equivariances present in the data directly in image space. However, as we noted the encoder can play a useful role by either restructuring the information in a convenient way or removing useless information.
%,  recover disentangled transformations linking data samples with each others. 
%

An important direction for future work will be to expand the reach of the theory to a broader family of transformations. In particular, it is unclear how the proposed approach should be adapted to learn transformations which are not affine or linear in image space, such as local deformations, compositional transformations (acting on different objects present in an image), and out-of plane rotations of objects in images (but see \citet{dupont_equivariant_2020} for an empirical success using a variant of the shift operator). Another important direction would be to extend the theory and proposed models to continous Lie groups. Moreover, our current implementation of disentanglement relies on some supervision, by including pairs of transformed images. It would be important to understand how disentangled representations can be learned without such pairs (see \citet{anselmi_symmetry-adapted_2019,zhou_meta-learning_2020} for relevant work).\\
\\
Finally, our work lays a theoretical foundation for the recent success of a new family of methods that---instead of enforcing disentangled representations to be restricted to distinct subspaces---use operators (hard-coded or learned) acting on the entire latent space \citep{connor_representing_2020,connor_variational_2020,dupont_equivariant_2020,giannone_no_2020,quessard_learning_2020} (see also \citep{memisevic_learning_2010, cohen_learning_2014, culpepper_learning_2009,sohl-dickstein_unsupervised_2017} for precursor methods). These methods work well where traditional disentanglement methods fail: for instance by learning to generate full 360$\degree$~in-plane rotations of MNIST digits \citep{connor_representing_2020}, and even out-of-plane rotations of 3D objects \citep{dupont_equivariant_2020}. These methods use distributed operators in combination with non-linear autoencoder architectures, an interesting direction for future theoretical investigations. Moreover, in the case where the latent operators cannot be determined in advance like in the affine case, these operators could be learned like in \cite{connor_representing_2020}. A benefit of this approach is that multiple operators can be learned in the same subspace, instead of the stacking strategy that we needed to use in the case of hard-coded shift operators.

\subsubsection*{Acknowledgments}
We would like to thank Armand Joulin, Léon Bottou, Yann LeCun, Pascal Vincent, Gus Lonergan, Fabio Anselmi, Li Jing, Marissa Connor, Chris Rozell, Emilien Dupont, Nathalie Baptiste, Taco Cohen, Kamesh Krishnamurthy for useful discussions and comments on the manuscript.

\bibliography{iclr2021_conference}
\bibliographystyle{iclr2021_conference}

\newpage
\appendix
\addcontentsline{toc}{section}{Appendix} % Add the appendix text to the document TOC
\part{Appendix} % Start the appendix part
\parttoc % Insert the appendix TOC

\section{Prerequisites in Group Theory and Representation Theory}
\label{sec:equiv}

\paragraph{Definition of a group}
A group is a set $G$ together with an operation $\circ: G \times G \rightarrow G$ such that it respects the following axioms:
\begin{itemize}
    \item Associativity: $(g_k \circ g_{k'}) \circ g_{k''} = g_k \circ (g_{k'} \circ g_{k''})$ with $g_k,~g_{k'},~g_{k''} \in G$.
    \item Identity element: there exists an element $e_G \in G$ such that, for every $g_k \in G$, $g_k \circ e_G = e_G \circ g_k = g_k$, and $e_G$ is unique.
    \item Inverse element: for each $g_k \in G$, there exists an element $g_{k'} \in G$, denoted $g_k^{-1}$ such that $g_k \circ g_k^{-1} = g_k^{-1} \circ g_k = e_G$.
\end{itemize}
In the paper, for clarity we do not write explicitly the operation $\circ$ unless needed.
\paragraph{Finite cyclic groups}
We will be interested in \emph{finite} groups, composed of a finite number of elements (i.e. the order of $G$).\\
\\
A \emph{cyclic} group is a special type of group that is generated by a single element, called the generator $g_0$, such that each element can be obtained by repeatedly applying the group operation $\circ$ to $g_0$ or its inverse. Every element of a cyclic group can thus be written as $g^k=g_0^k$. Note that every cyclic group is abelian (i.e. its elements commute).\\
\\
A group $G$ that is both finite and cyclic has a finite order $K$ such that $g_0^K=e_G.$

\paragraph{Representation and equivariance}
Informally, for a model to be equivariant to a group of transformations means that if we encode a sample, and transform the code, we get the same result as encoding the transformed sample. \citep{higgins_towards_2018} show that disentanglement can be viewed as a special case of equivariance, where the transformation of the code is restricted to a subspace. We provide a formal definition of equivariance below after introducing notations.\\
\\
In the framework of group theory, we consider $\phi$ a \textbf{linear representation} of the group $G$ acting on $X$ \citep{Scott:and:Serre:96}:~ $\phi: G \rightarrow GL(X)$. Each element of the group $g_k \in G$ is represented by a matrix $\phi(g_k)=\phi_k$, and $\phi_k$ is a matrix with specific properties:
\begin{enumerate}
    \item $\phi$ is a homomorphism: $\phi(g_kg_{k'})=\phi(g_k)\phi_(g_{k'}),~g_k, g_{k'} \in G$.
    \item $\phi_k$ is invertible and $\phi_k^{-1} = \phi(g_k^{-1})$ as $\phi(g_k^{-1})\phi(g_k) = \phi(g_k g_k^{-1}) = \phi_(e_G) = I$. where $I$ is the identity matrix in $GL(X)$.
\end{enumerate}
The set of matrices $\phi_k$ form a \emph{linear representation} of $G$ and they multiply with the vectors in $X$ as follows:
\begin{flalign}
~&\phi_k: X(=\mathbb{R}^{N}) \rightarrow X~s.t.~\forall x\in X, \phi_k(x)  \in X.
\end{flalign}
The \textbf{character} of $\phi$ is the function $\chi_{\phi}$ such that for each $g_k \in G$, it returns the trace of $\phi(g_k)=\phi_k$, i.e. $\chi_{\phi}(g_k) = Tr(\phi_k)$. Importantly, the character of a representation completely characterizes the representation up to a linear isomorphism (i.e. change of basis) \citep{Scott:and:Serre:96}. The \textbf{character table} of a representation is composed of the values of the character evaluated at each element of the group. \\
\\
Similarly, we denote a linear representation of the action of $G$ onto the latent space $Z$ by $\psi: G \rightarrow GL(Z)$ such that $\forall k,~\psi(g_k)=\psi_k$. While corresponding to the action of the same group element $g_k \in G$, $\psi_k$ does not have to be the same as $\phi_k$, as it represents the action of $G$ on $Z$ and not on $X$. 
\begin{flalign}
\psi_k:~Z(=~\mathbb{C}^N) \rightarrow Z~s.t.~\forall z\in Z, \psi_k(z) \in Z.
\end{flalign}
Note that we consider a complex latent space. With these notions, formally the model $f: X \rightarrow Z $ is \textbf{equivariant} to the group of transformations $G$ that acts on the data if, for all elements of the group $g_k$ and its action $\phi_k$ and $\psi_k$ on the spaces $X$ and $Z$ respectively, we have:
\begin{equation}
\forall x\in X, \forall k \in K, f(\phi_k(x))  = \psi_k(f(x))
\label{eq:eqmapp1}
\end{equation}

Finally, a group action on image space $\mathbb{R}^N$ is said to be \textbf{affine} if it affects the image through an affine change of cooordinates.

\section{Experimental Details}
\label{sec:experimentalDetails}

\subsection{Dataset generation}
\label{sec:datagen}

\paragraph{Simple shapes}
We construct a dataset of randomly generated simple shapes, consisting of 5 randomly chosen points connected by straight lines with 28x28 pixels. We normalize pixel values to ensure they lie within $[0, 1]$. For all experiments, we use a dataset with $2000$ shapes. For each shape, we apply 10 counterclockwise rotations by $\{0^\circ, 36^\circ, 72^\circ, \dots, 324^\circ\}$ using scikit-image's rotation functionality (see \citet{scikit-image}). For translations along the x-axis or y-axis we apply 10 translations using numpy.roll (see \citet{van2011numpy}). This ensures periodic boundary conditions such that once a pixels is shifted beyond one edge of the frame, it reappears on the other edge. For experiments with supervision involving pairs, we construct every possible combination of pairs, $x_1, x_2$ and apply every transformation to both $x_1$ and $x_2$. For datasets containing multiple transformations, we first rotate, then translate along the x-axis, then translate along the y-axis. We use 50\% train-test split then further split the training set into 20\%-80\% for validation and training. 

\paragraph{MNIST \citep{lecun2010mnist}}
Similar to simple shapes, we construct rotated and translated versions of MNIST by applying the same transformations to the original MNIST digits. We normalize pixel values to lie between $[0, 1]$. For supervised experiments, we similarly construct every combination of pairs $x_1, x_2$ and apply transformations to both $x_1$ and $x_2$. Since constructing every combination of transformed pairs would lead to many multiples the size of the original MNIST, we randomly sample from the training set to match the original number of samples. We use the full test set augmented with transformations for reporting losses.

\subsection{Model architectures and training}
\label{sec:architectures}

We implement all models using PyTorch with the Adam optimizer \citep{NEURIPS2019_9015,kingma2014adam}. 

%For VAE, $\beta$-VAE, and CCI-VAE, we train on a transformed sample $x_1$ assuming a Gaussian likelihood.
\paragraph{Variational Autoencoder and Variants}
We implement existing state-of-the-art disentanglement methods $\beta$-VAE and CCI-VAE, which aim to learn factorized representations corresponding to factors of variation in the data. In our case the factors of variation are the image content and the transformation used (rotation or translations). We use the loss from CCI-VAE made up of a reconstruction mean squared error and a Kullback–Leibler divergence scaled by $\beta$: 
\begin{equation}
    L = \sum_i^m(x_i - f_D(f(x_1))^2 / m + \beta |KL(q(z | x), p(z)) - C|
\end{equation}

where $m$ is the number of samples and Kullback-Leibler divergence is estimated as $KL(q(z | x), p(z) = -0.5 * \sum_i^d(1 + \ln(\sigma_i^2) - \mu_i^2 - \sigma_i^2)$, where $d$ is the latent dimension (see \citet{kingma_auto-encoding_2014}). For a standard VAE, we use $C = 0$ and $\beta=1.0$. For $\beta$-VAE, we sweep over choices of $\beta$ with $C=0$. For CCI-VAE, we sweep over choices of $\beta$ and linearly increase $C$ throughout training from $C=0$ to $C=36.0$.

We use the encoder/decoder architectures from \citet{burgess_understanding_2018} comprised of 4 convolutional layers, each with 28 channels, 4x4 kernels, and a stride of 2. This is followed by 2 fully 256-unit connected layers. We apply ReLU activation after each layer. The latent distribution is generated from 30 units: 15 units for the mean and 15 units for the log-variance of a Gaussian distribution. The decoder is comprised of the same transposed architecture as the encoder with a final sigmoid activation. 

\paragraph{Autoencoder with latent operators}\label{autoencoderDetails}
For the standard autoencoder and autoencoders with the shift/disentangled latent operators, we use supervised training with pairs $(x_1, x_2)$ and a transformation parameter $k$ corresponding to the transformation between $x_1$ and $x_2$. The loss is the sum of reconstruction losses for $x_1$ and $x_2$: 
\begin{equation}
    L = \sum_{i=1}^m(x_{1, i} - f_D(f(x_{1, i}))^2 / m + \sum_{i=1}^m (x_{2, i } - f_D(\psi_k(f(x_{1, i}))))^2 / m
    \label{eq:l2loss}
\end{equation}

where $m$ is the number of samples and $\psi$ is the disentangled or shift latent operator. For a standard autoencoder, only the first reconstruction term is present in the loss function. 

For the non-linear autoencoders, we use the same architecture above based on CCI-VAE. In the linear case, we use a single fully 28x28 connected layer and an 800 dimensional latent space. We use 800 dimensions to approximate the number of pixels (28x28) and ensure that $K$=10 divides the latent dimension. This is an approximation of the correct theoretical operator (which should be invertible) that works well in practice. 

\paragraph{Weakly supervised shift operator model}
We use linear encoders and decoders with a $784$ dimensional latent space to match the number of pixels. Indeed the weakly supervised shift operator uses the complex version of the shift operator, so we can perfectly match the size of the image. Training is done with L2 loss on all possible reconstructions of $x_1$ (of each training pair $i$), weighted by scores $\alpha_{i,k}$. Appendix \ref{sec:appendixReddy} below gives a detailed explanation of the computation of the scores $\alpha_{i,k}$:
\begin{equation}
    L = \sum_{i=1}^m(x_{1, i} - f_D(f(x_{1, i}))^2 / m + \sum_{i=1}^m \sum_{k=1}^{K_L} \alpha_{i,k} (x_{2, i} - f_D(\psi_k(f(x_{1, i, k}))))^2 / m
\end{equation}
where $m$ is the number of samples. At test-time, we use the transformation with maximum score $\alpha_{i,k}$. 

\paragraph{Stacked shift operator model}
We use linear encoders and decoders with a $784$ dimensional latent space to match pixel size, as we use in the stacked model the complex version of the shift operator. Intermediate layers $L_i$ are invertible linear layers of size $784$ as well.  Training is done with L2 loss on reconstructed samples as in the autoencoder with shift latent operator (see Equation \ref{eq:l2loss}.)

\subsection{Weakly supervised shift operator training procedure}
\label{sec:appendixReddy}

\paragraph{Method} We experimentally show in Section \ref{sec:weakshift} that the proposed operator $\psi_k$ works well in practice. Additionally, we developed a method for inferring the parameter of the transformation $k$ that needs to act on the sample. We encode samples by pairs $x_1,x_2$ (with $x_2$ a transformed version of $x_1$) into $z_1$ and $z_2$ respectively, and use a classical phase correlation technique \citep{Reddy1996} to identify the shift between $z_1$ and $z_2$, described below. Then, we use the complex diagonal shift operator parametrized by the inferred transformation parameter $k$.\\
\\
Importantly, in the weakly supervised version of the shift operator, the model has an extra free parameter which is the number of latent transformations, not known a priori. Let us denote $K_L$ this number, which can be different than the ground-truth order of the group $K$.\\
\\
To infer the latent transformation that appear between $z_1$ and $z_2$, we compute the cross-power spectrum between the two codes $z_1$ and $z_2$, that are both complex vectors of size $N$ and obtain a complex vector of size $N$. We repeat $K_L$ times this vector, obtaining a $K_L$ x $N$ matrix, of which we compute the inverse Fourier transform. The resulting matrix should have rows that are approximately 0, except at the row $k$ corresponding to the shift between the two images, see \citet{Reddy1996}. Thus, we compute the mean of the frequencies of the real part of the inverse Fourier result (i.e. the mean over the $N$ values in the second dimension). This gives us a $K_L$-dimensional vector, which we use as a vector of scores of each $k$ to be the correct shift between $z_1$ and $z_2$. During training, we compute the soft-max of these scores with a temperature parameter $k$, this gives us $K_L$ weights $\alpha_k$. We transform $z_1$ with all $K$ possible shift operators, decode into $K$ reconstructions $x_k$, and weight the mean square error between $x_2$ and each $x_k$ by $\alpha_k$ before back-propagating. This results, for each samples pair $(x_{1, i},x_{2, i})$, in the loss:
\begin{equation}
    L = \sum_{i=1}^m(x_{1, i} - f_D(f(x_{1, i}))^2 / m + \sum_{i=1}^m \sum_{k=1}^{K_L} \alpha_{i,k} (x_{2, i} - f_D(\psi_k(f(x_{1, i, k}))))^2 / m
    \label{eq:l2lossweak}
\end{equation}
where $m$ is the number of samples and $\alpha_{i,k}$ the scores for the pair $(x_{1, i},x_{2, i})$. At test-time, we use the transformation with maximum score $\alpha_{i,k}$.

\begin{table}[h!]
\begin{center}
\begin{tabular}{l|c|c}
\bf Dataset  &\multicolumn{2}{c}{\bf $K_L$}\\
~ & $10$ & $21$ \\
Shapes (10,0,0) & $0.001 \pm 0.0013$ & $0.0005 \pm 0.0001$\\
Shapes (0,10,0)& $0.0097 \pm 0.0052$ & $0.0038 \pm 0.0013$\\
Shapes (0,0,10)& $0.0115 \pm 0.0049$ & $0.005 \pm 0.0033$ \\
MNIST (10,0,0)& $0.0035 \pm 0.0039$ & $0.0074 \pm 0.0083$\\
\end{tabular}
\end{center}
\caption{Comparing test mean square error (MSE) $\pm$ standard deviation of the mean over random seeds for different $K_L$. Numbers in $()$ refer to the number of rotations, the number of translations on the $x$-axis, and the number of translations on the $y$-axis respectively.}
\label{tab:mse2}
\end{table}

\paragraph{Effect of the number of latent transformations} In the weakly supervised shift operator model, the number of latent transformations $K_L$ is a free parameter. Interestingly, when using rotations the best cross-validated number of transformations is $10$, which matches the ground-truth order of the group. For translations (either on the $x$ or the $y$ axis), best results are obtained using $K_L=21$ which is larger than the ground-truth order of the group $K$. Table \ref{tab:mse2} compares test MSE for both values of $K_L$. We think that in the case of translations, changes in the image induced by each shift (each transformation) are less visually striking than with rotations, and a larger $K_L$ gives extra flexibility to the model to identify the group elements and respects the group structure (namely its cyclic aspect). 

\subsection{Hyper-parameters}
\label{sec:hparams}

\paragraph{General hyper-parameters} We sweep across several sets of hyper-parameters for our experiments. We report results for the model with the lowest validation test loss. To avoid over-fitting, results for any given model are also stored during training for the parameters yielding the lowest validation loss. 

For experiments with simple shapes we sweep across combinations of 
\begin{itemize}
    \item 5 seeds: 0, 10, 20, 30, 40 
    \item 4 batch sizes: 4,8,16,32
    \item 2 learning rates: 0.0005, 0.001
\end{itemize}

For MNIST, we sweep across combinations of 
\begin{itemize}
    \item 5 seeds: 0, 10, 20, 30, 40 
    \item 4 batch sizes: 8,16,32, 64
    \item 2 learning rates: 0.0005, 0.001
\end{itemize}

In addition to these general parameters, we also sweep across choices of $\beta$, $\{4, 10, 100, 1000\}$ and latent dimension $\{10, 30\}$ for the variational autoencoder models and variants. We repeat all experiments across four seeds used to initialize random number generation and weight initialization.
%for computation libraries used in data generation and training. Note we approximate all loss functions by averaging over batch losses.

\paragraph{Weakly supervised shift operator hyper-parameters}

%as described in \ref{sec:hparams}
We perform a sweep over hyper-parameters as described above. Additionally for the weakly supervised model, we sweep over temperature $\tau$ of the soft-max that shapes the scores of each transformation over values $\tau=\{0.01,0.1,1.0\}$, and the number of transformations composing the operator family (i.e. the order of the group) over values $10$ and $21$, where $10$ is the ground-truth order of the group.

\paragraph{Stacked shift operator model hyper-parameters}

We perform a sweep over hyper-parameters as described above. The only exception is that for the case of the Special Euclidean group, we train only for $5$ epochs, and try batch sizes $\{32,64\}$ for MNIST and $\{16,32\}$ for simple shapes, as the number of generated samples is high. Similarly, for the case of $2$D translations, we use batch sizes $\{16,32\}$ for simple shapes.

\section{Reframing Disentanglement: Formal Proofs}
\label{sec:topo_proof}
We consider the action of a finite group on image space $\mathbb{R}^N$. Using tools from topology, we show that it is impossible to learn a representation which disentangles the action of this group with a continuous encoder $f$.

\subsection{Topological proof against disentanglement}
\label{sec:topo_proof1}

We consider a finite group $G$ of cardinal $|G|$ that acts on $\mathbb{R}^N$. Given an image $x \in \mathbb{R}^N$, an orbit containing that image is given by $\{g_1 x,g_2 x,..., g_K x\}$.

We consider an encoder $f:\mathbb{R}^N \rightarrow M$ that disentangles the group action. The image of $f$ is composed of an equivariant subspace and an invariant subspace. We define $f_E:\mathbb{R}^N \rightarrow M_E$ the projection of $f$ on its equivariant subspace and $f_I:\mathbb{R}^N \rightarrow M_I$ the projection of $f$ on its invariant subspace. 

$f_E$ is equivariant to the group action. In equations:
\begin{equation}
\forall x, \forall g, f_E(gx) = g f_E(x)
\end{equation}

For the disentanglement to be complete, $f_E$ should not contain any information about the identity of the orbit the image belongs to. If $O_1$ and $O_2$ are two distinct orbits, we thus have: 
\begin{equation}
\forall x_1 \in O_1,\forall x_2 \in O_2, \exists g\in G, f_E(x_1) = g f_E(x_2)
\label{eq:equi2}
\end{equation}

$f_I$ is invariant to the group action:
\begin{equation}
\forall x, \forall g, f_I(gx) = f_I(x)
\end{equation}

We also assume that the representation contains all the information needed to reconstruct the input image:
\begin{equation}
\forall x_1 \in O_1,\forall x_2 \in O_2, f_I(x_1) \neq f_I(x_2)
\label{eq:inv2}
\end{equation}

This last assumption corresponds to assuming that every image can be perfectly identified from its latent representation by a decoder (i.e. perfect autoencoder). We also assume that both the encoder and decoder are continuous functions. This is a reasonable assumption as most deep network architectures are differentiable and thus continuous. In the language of topology, the encoder $f$ is a \emph{homeomorphism}, a continuous invertible function whose inverse is also a continuous function. Also called a topological isomorphism, an homeomorphism is a function that preserves all topological properties of its input space (see \citet{munkres_topology_2014} p.105). Here we prove that $f$ cannot preserve the topology of $\mathbb{R}^N$ while disentangling the group action of $G$.

Consider $f_E|_O$ the restriction of $f_E$ to a single orbit of an image $x$ without any particular symmetry. $f_E|_O$ inherits continuity from $f$, and it can easily be shown that $f_E|_O$ is invertible (otherwise, information about the transformation on this orbit is irremediably lost, and $f$ can thus not be invertible).

$f_E|_O$ is thus also a homeomorphism, and so it preserves the topology of the orbit $O$ in image space, which is a set of $|G|$ disconnected points.

By equation \ref{eq:equi2}, we know that restrictions of $f_E$ to all other orbits have an image contained in the same topological space. As a consequence, the image of $f_E$ itself is a set of $|G|$ disconnected points.  

Since $f_E$ is a projection of $f$, the image of $f$ should at least be composed of $|G|$ disconnected parts (this follows from the fact that the projection of a connected space cannot be disconnected). However, this is impossible because the domain of $f$ is $R^N$, which is connected, and $f$ is an homeomorphism, thus preserving the connectedness of $R^N$.

%Note that an analog argument can be made in all cases where the topology of the image space cannot be projected onto the topology of a single orbit. For example, in the case where the transformation is isomorphic to the continuous group of rotations $SO(2)$, the topology of an orbit is not simply connected, which is incompatible with any projection of $R^N$. Thus, even in the case of continuous rotations, disentanglement is impossible.

In summary, we have shown that, for topological reasons, a continuous invertible encoder cannot possibly disentangle the action of a finite group acting on image space $R^N$. In the next section, we show that topological defects arise in the neighborhood of all images presenting a symmetry with respect to the transformation.

\subsection{Topological defects arise in the neighborhood of symmetric images}
\label{sec:topo_proof2}

We proved in the previous section that it is impossible to map a finite group acting on $R^N$ to a disentangled representation with a continuous invertible encoder. In this section, in order to gain intuition of why disentanglement is impossible, we show that topological defects appear in the neighborhood of images that present a symmetry with respect to the group action. 

\subsubsection{$f_E$ is not continuous about symmetric images: Formal Proof}

Let's consider an image $x_s$ that presents a symmetry with respect to the group action:
\begin{equation}
\exists g\in G, g\neq e_G, g x_s = x_s
%\label{eq:}
\end{equation}

Let's further assume that an infinitesimal perturbation to this image along a direction $u$ breaks the symmetry of the image:

\begin{equation}
\forall 0<\eps<E, x':=x_s+\eps u, g x' \neq x'
%\label{eq:}
\end{equation}

Since $f_E$ preserves the information about the transformation,
\begin{equation}
|f_E(g x') - f_E(x')|>C\neq0
\label{eq:ineq}
\end{equation}

where $C$ is the smallest distance between two disconnected points of $M_E$. We assume that the group action is continuous: 
\begin{equation}
g x' = g x_s + O(\eps) = x_s + O(\eps)
%\label{eq:}
\end{equation}

We can rewrite equation \ref{eq:ineq} as:
\begin{equation}
|f_E(x_s + O(\eps)) - f_E(x_s +O(\eps))|>C\neq0
\label{eq:ineq2}
\end{equation}

which is in contradiction with the continuity hypothesis on the encoder. We have thus shown that the equivariant part of the encoder $f_E$ presents some discontinuities around all images that present a symmetry. Note that for both rotations and translations, the uniform image is an example of symmetric image with respect to these transformations.

\subsubsection{$f_I$ is not Differentiable about Symmetric Images: Visual Proof}

\begin{figure}[h]
\begin{center}
\includegraphics[width=9cm]{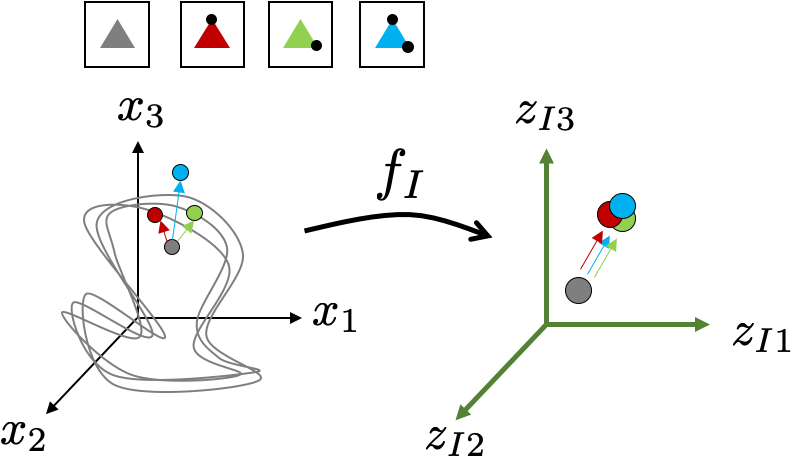}
%\framebox[4.0in]{$\;$}
\end{center}
\caption{Visual proof that the invariant part of the encoder $f_I$ cannot be differentiable about symmetric figures. We assume $f_I$ is differentiable and show a contradiction. We consider an equilateral triangle which is perturbed at its top corner, left corner, or both corners. When perturbed either one of its corner, the perturbation brings the image to the same orbit, because of the symmetry. In latent space, the perturbation should thus move the latent representation in the same direction. The perturbation along the two corners simultaneously brings to image to a different orbit, and yet, since the perturbation is a simple linear combination of the single-corner perturbations, it can only be colinear to these perturbations. This collinearity leads to the encoder not being injective, and thus loosing information about the identity of the image. }
\label{fig:fig_suppl}
\end{figure}

As an example, we consider an equilateral triangle which is either perturbed at its top corner, left corner, or both corners (Fig. \ref{fig:fig_suppl}). When perturbed on either one of its corner, the perturbation moves the image to the same orbit, because the triangle perturbed on its right corner is a rotated version of the triangle perturbed on its top corner. The gradient of $f_I$ along these two directions at the equilateral triangle image should thus be the same (so as not to leak information about the transformation in the invariant subspace $Z_I$). The simultaneous perturbation along the two corners moves the image to a different orbit, so the gradient of $f_I$ along this direction should not be aligned with the previous gradients (so as to preserve all information about identity of the orbit). And yet, if the function $f_I$ was differentiable everywhere, this gradient should be a linear combination of the former gradients, and thus all three gradients should be collinear. The function $f_I$ can thus not be differentiable everywhere. This imperative is incompatible with many deep learning frameworks, where the encoder is implemented by a neural network that is differentiable everywhere (with a notable exception for networks equipped with relu nonlinearities which are differentiable almost everywhere).

\subsubsection{$f_I$ is not differentiable about Symmetric Images: Formal proof}

Next, we show that, if we add an extra assumption on the encoder, namely that it is differentiable everywhere, it is also impossible to achieve the invariant part of the encoder $f_I$. Note that this extra assumption if true for networks equipped with differentiable non-linearities, such as tanh or sigmoid, but not for networks equipped with relus. 

Let's consider an image $x_s$ presenting a symmetry w.r.t the group action. We consider perturbations of that image along two distinct directions $u$ and $g u$.

By symmetry, it is easy to see that:

\begin{equation}
\frac{\partial f_I}{\partial u}\bigg|_{x_s} = \frac{\partial f_I}{\partial g u}\bigg|_{x_s}
%\label{eq:ineq2}
\end{equation}

As a consequence, a perturbation by $u' = {(u + gu) \over  2}$, is equal to the perturbation along one or the other direction:

\begin{equation}
\frac{\partial f_I}{\partial u'}\bigg|_{x_s} = {1 \over 2} * \left(\frac{\partial f_I}{\partial u}\bigg|_{x_s} + \frac{\partial f_I}{\partial g u}\bigg|_{x_s}\right) = \frac{\partial f_I}{\partial u}\bigg|_{x_s}
%\label{eq:ineq2}
\end{equation}

In the general case, $x' = x_s + \eps u'$ does not belong to the same orbit as $x = x_s + \eps u$. $f_I$ is thus losing information about which orbit the perturbed image $x'$ belongs to, which is in contradiction with the assumption shown in Equation \ref{eq:inv2}.

\subsection{Character Theory of the disentanglement of finite discrete linear transformations}
\label{sec:charactheory}
\subsubsection{The Shift Operator}
 Datasets are structured by group transformations that act on the data samples. Our goal is for the model to reflect that structure. Specifically, we want our model to be equivariant to the transformations that act on the data. We defined group equivariance in Section \ref{sec:equiv}. \\
\\
We show here that a carefully chosen distributed operator in latent space, —the shift operator— is linearly isomorphic to all cyclic linear transformations of finite order $K$ of images (i.e. the group $G$ of transformations is cyclic and finite and it acts linearly on image space). The practical consequences of this fact is that it is possible to learn an equivariant mapping to every affine transformation using this operator, using linear invertible encoder and decoder architectures.\\ 
\\
Consider a linear encoder model $f=W$. If we want $W$ to be an equivariant invertible linear mapping $W$ between $X$ and $Z$, Equation \ref{eq:eqmapp1} rewrites as follows:
\begin{equation}
W~\phi_k(x)  = \psi_k(W~x)~~\forall x\in X(=\mathbb{R}^N), \forall k \in K
\label{eq:eqmapp}
\end{equation}
where $\phi_k$ and $\psi_k$ are the representations of $g_k \in G$ on the image and latent space respectively, as defined in \ref{sec:equiv}.\\
\\
For $W$ to be equivariant, Equation \ref{eq:eqmapp} must be true for every image $x$. As $W$ is invertible, Equation \ref{eq:eqmapp} is true if and only if, $\forall k \in K$, the two representations $\psi_k$ and $\phi_k$ are isomorphic:
\begin{equation}
\forall k \in K,  \phi_k  =W^{-1} ~\psi_k ~ W
\end{equation}

%\footnote{When we will consider multiple transformations, the resulting group will be non-cyclic but each subgroup is cyclic.}

We consider additional properties on $\phi$ corresponding to the assumptions (i) that $G$ is cyclic of order $K$ with generator $g_0$ and (ii) $\phi$ is isomorphic to the regular representation of $G$ (see \citet{Scott:and:Serre:96}):
\begin{enumerate}
    \item $\phi$ is cyclic, i.e. such that $\phi_k=\phi_0^k$ where  $\phi_0=\phi(g_0)$ and thus $\phi_0^K=I$.
    \item The character of $\phi$ is s.t. $\chi_{\phi}(e)=N$ and $\chi_{\phi}(g_k)=0$ for $g_k\neq e$.
\end{enumerate}
The second property might seem counter-intuitive, but it just means that the transformation leaves no pixel unchanged (i.e. permutes all the pixel). In the case of rotations, this is approximately true since only the rotation origin remains in place. The character table of $\phi$, given the second property, is 

{
\centering
\begin{tabular}{l|c|c|c|c|c|c}
  & e &$g_0$ & $g_0^2$ & ... & $g_0^{K-1}$\\
$\chi_\phi$ & $N$ & $0$ & $0$  & $0$  & $0$\\
\end{tabular}
\par}

%$\forall k \in 1, \ldots, K$
 We have just seen that if the encoder and decoder are linear and  invertible, the two representations $\phi$ and $\psi$ must be isomorphic. Two representations are isomorphic if and only if they have the same character \citep[Theorem 4, Corollary 2]{Scott:and:Serre:96}. We thus want to choose $\psi$ such that it preserves the character of the representation $\phi$ corresponding to the action of $G$ on the dataset of images. Importantly, we will see that our proposed operator needs to be \emph{distributed} in the sense that it should act on the full latent code.\\
\\
Let us consider the matrix of order $K=|G|$ that corresponds to a shift of elements in a $K$-dimensional vector by $k$ positions. It is the permutation corresponding to a shift of $1$, exponentiated by $k$. We construct from $M_k$ the \emph{shift operator} as a representation of the group's action on the latent space. For each $g_k \in G$ such that $g_k = g_0^k$, its corresponding shift operator is the block diagonal matrix of order $N$ composed of $\frac{N}{K}$ repetition of $M^k$.

\noindent\begin{minipage}{.5\textwidth}
\begin{equation}
M_k:=\begin{bmatrix} 
    0      & 0 & \ldots  & 1 \\
    1      & 0 & \ldots & 0 \\
    0      & 1 &    0   & \vdots\\
    \vdots &   &      & \\
    0      &  \ldots &  1     & 0
\end{bmatrix}^{\scalebox{1.5}{$k$}}
\end{equation}
\end{minipage}%
\begin{minipage}{.5\textwidth}
\begin{equation}
  \psi_k:=\begin{tikzpicture}[decoration={brace,amplitude=5pt},baseline=(current bounding box.west)]
     \matrix (magic) [matrix of math nodes,left delimiter={[},right delimiter={]}] {
        M^k\\
        & M^k\\
        && ...\\
        &&& M^k\\
     };
  \end{tikzpicture}
\end{equation}
\end{minipage}

Let us compute the character table of this representation. First, for the identity, is it trivial to see that $\chi_\psi(e)=N$. Second, for any $g_k\neq e$, we have 
\begin{equation}
    \chi_\psi(g_k) = Tr(\psi_k) = 0
\end{equation}
since all diagonal elements of $g_k$ will be $0$. Therefore, the character table of the shift operator is the same as $\chi_\phi$:\\

{
\centering
\begin{tabular}{l|c|c|c|c|c|c}
  & e &$g_0$ & $g_0^2$ & ... & $g_0^{K-1}$\\
$\chi_\psi$ & $N$ & $0$ & $0$  & $0$  & $0$\\
\end{tabular}
\par}
Using this shift operator ensures that an equivariant invertible linear mapping $W$ exists between image space and the latent space equipped with the shift operator. Note that the character of the \emph{disentangled operator} does not match the character table of $\phi$, and so we verify once again in this linear autoencoder setting that the disentangled operator is unfit to be equivariant to affine transformations of images.  \\
\\
When $K$ does not divide $N$, we use a latent code dimension that is slightly different than $N$ but divisible by $K$. This is an approximation of the correct theoretical operator and we verify that it works well in practice.

In the next Section \ref{sec:complexshift}, we show that we can also replace this shift operator by a complex diagonal operator, which is more computationally efficient to multiply with the latent.

\subsubsection{Complex Diagonal Shift Operator}
\label{sec:complexshift}
In order to optimise computational time, we can also consider for $M_k$ the following diagonal complex matrix:
\begin{equation}
M_k := \begin{bmatrix} 
    1 & 0 & \ddots  & 0 \\
    0 & \omega &  & \\
    \vdots &  & \ddots & \\
    0 &     &     & \omega^{K-1} 
    \end{bmatrix} ^{\scalebox{1.5}{$k$}}
\end{equation}
with $\omega = e^{\frac{2 i \pi}{K}}$. The shift operator in this case is a diagonal matrix, as follows:
\begin{equation}
   \psi_{k,N}:=\begin{tikzpicture}[decoration={brace,amplitude=5pt},baseline=(current bounding box.west)]
     \matrix (magic) [matrix of math nodes,left delimiter={[},right delimiter={]}] {
        1\\
        & \ldots\\
        && \omega^{K-1} \\
        &&& 1 \\
        &&&& \ldots \\
        &&&&& \omega^{K-1} \\
        &&&&&& \ldots \\
        &&&&&&& \omega^{K-1} \\
     };
      \draw[decorate] (magic-4-4.north) -- (magic-6-6.north east) node[above=5pt,midway,sloped] {$M_k$};
     \draw[decorate] (magic-1-1.north) -- (magic-3-3.north east) node[above=5pt,midway,sloped] {$M_k$};
   \end{tikzpicture}
\end{equation}
Let us compute the character table of this representation. First, for the identity, is it trivial to see that $\chi_\psi(e)=N$. Second, for any $g_k\neq e$, we have 
\begin{equation}
    \chi_\psi(g_k) = Tr(\psi_k) = \sum_{n=0}^{N-1} (\omega^k)^n = \frac{1 - (\omega^k)^N}{1 - \omega^k} = 0
\end{equation}
as ${\omega^k}^N = e^{\frac{2 i \pi k N}{K}} = 1$ since we're assuming $N$ can be divided by $K$. Again, the character table of the shift operator is the same as $\chi_\phi$. Using this operator fastens computation since it requires only multiplying the diagonal values (a vector of size $N$) with the latent code (a vector of size $N$ as well) instead of doing a matrix multiplication. Note that when using this complex version of the shift operator, the encoding and decoding layers of the autoencoder should be complex as well. \\
\\
When $K$ does not divide $N$, we still use a latent code of size $N$ and the operator $\psi_k$ is a diagonal matrix of order $N$, but the last cycle $1, \omega, \ldots $ is unfinished (does not go until $\omega^{K-1}$). The character of this representation is no longer equal to $0$ for non-identity elements $g_k \neq e$, but equals a small value $<<N$ and this approximation works well in practice.

\section{The Case of Multiple Transformations: Formal Derivations}

\subsection{Translations in both axes as a direct product}
\label{sec:appendixdp}

To cover the case of $2$D translations (acting on both $x$ and $y$ axes of the image), we consider an abelian group $G$ that is the direct product of two subgroups $G=A_x \times A_y$. Both $A_x$ and $A_y$ are normal in $G$ because every subgroup of an abelian group is normal. Moreover, we consider that $A_x$ and $A_y$ are both cyclic of order $K$ and $K'$ respectively, which is the case for integer translation of an image using periodic boundary condition. We denote $a_{x,0}$ and $a_{y,0}$ the generators of $A_x$ and $A_y$, and write each translation as $a=(a_{x,k},a_{y,k'})$ with $a_{x,k}=a_{x,0}^k,~a_{y,k'}=a_{y,0}^{k'}$.\\
\\
We show that the shift operator can handle this case, with differences. We assume for simplicity that both $K$ and $K'$ divide $N$, and $KK'$ divides $N$. Following \citet{Scott:and:Serre:96}, we write group elements $g \in G$ as $g_{(k,k')}=(a_{x,k},a_{y,k'})$. The order of the group is $K=K K'$. If we consider the regular representation of $2$D translations over the image space, as in Section \ref{sec:charactheory}, its character table is

{
\centering
\begin{tabular}{l|c|c|c|c|c|c}
  & e &$g_{(0,1)}$ & $g_{(0,2)}$ & ... & $g_{(K-1,K'-1)}$\\
$\chi_\phi$ & $N$ & $0$ & $0$  & $0$  & $0$
\end{tabular}
\par}

We consider two representations over the latent space: $\psi_{x}: A_x \rightarrow GL(\mathbb{C}^K)$ is a linear representation of $A_x$ and $\psi_{y}: A_y \rightarrow GL(\mathbb{C}^{K'})$ a linear representation of $A_y$, each of the matrix form described in Section \ref{sec:complexshift}. Group theory \citep{Scott:and:Serre:96} tells us that $\psi$ is the tensor product of $\psi_{x}$ and $\psi_{y}$, i.e. for $g=(k,k') \in G$:
\begin{equation}
\psi_k = \phi(g) = \phi((a_{x,k},a_{y,k'})) = (\psi_{x} \otimes \psi_{y})(a_{x,k},a_{y,k'}) = \psi_{x}(a_{x,k}) \otimes \psi_{y}(a_{y,k'}) 
\end{equation}

Let us consider the two shift operators: 
\begin{align}
    &\psi_{x,k} := Diag(1,\omega_1,\omega_1^2 ..., \omega_1^{K-1})^{k}\\
    &\psi_{y,k'} := Diag(1, \omega_2, \omega_2^2, \ldots\omega_2^{K'-1})^{k'}
\end{align}
where $\omega_1 = e^{\frac{2 i \pi}{K}}$ and $\omega_2 = e^{\frac{2 i \pi}{K'}}$. Then the tensor product $\psi_{x,k}\otimes\psi_{y,k'}$ writes as a diagonal matrix of order $KK'$:

\begin{equation}
\psi_{x,k}\otimes\psi_{y,k'} := \begin{bmatrix} 
    1 \\
    & \omega_2^{k'} \\
    && \omega_2^{2k'} \\
    &&& \ldots \\
    &&&& \omega_2^{k'(K'-1)} \\
    &&&&& \omega_1^{k} \\
    &&&&&& \omega_1^k\omega_2^{k'} \\
    &&&&&&& \ldots \\     
    &&&&&&&& \omega_1^{k(K-1)}\omega_2^{k'(K'-1)}  
    \end{bmatrix}
    \label{eq:tensorprod}
\end{equation}
The character of $\psi_{x,k}\otimes\psi_{y,k'}$ is
\begin{equation}
    \chi_{\psi_{x,k}\otimes\psi_{y,k'}} = \sum_{n=0}^{K'-1}\omega_2^{nk'} \times \sum_{n=0}^{K-1}\omega_1^{nk}
\end{equation}
If $k \neq 0$, $\sum_{n=0}^{K-1}{\omega_1^k}^n=\frac{1-{\omega_1^k}^K}{1 - \omega_1^k} = 0$ since ${\omega_1^k}^K = e^{2 i \pi k} = 1$ (and similarly for $\omega_2$ if $k' \neq 0$). Hence, for $(k,k') \neq (0,0)$,  $\sum_{n=0}^{K'-1}\omega_2^{nk'} \times \sum_{n=0}^{K-1}\omega_1^{nk}=0$. For $(k,k') = (0,0), \chi_{\psi_{x,k}\otimes\psi_{y,k'}} = KK'$. 
Thus, the character table of $\psi_x\otimes\psi_y$ is\\

{
\centering
\begin{tabular}{l|c|c|c|c|c|c}
  & e &$g_{(0,1)}$ & $g_{(0,2)}$ & ... & $g_{(K-1,K'-1)}$\\
$\chi_\psi$ & $KK'$ & $0$ & $0$  & $0$  & $0$
\end{tabular}
\par}
We will use for $2$D translations a diagonal operator that is the repetition of $\frac{N}{KK'}$ times $\psi_{x,k}\otimes\psi_{y,k'}$ (assuming $KK'$ divides $N$), denoted $\psi$, with is for $g_{k,k'}=(a_{x,k},a_{y,k'})$
\begin{equation}
   \psi_{k,k',N}:=\begin{tikzpicture}[decoration={brace,amplitude=5pt},baseline=(current bounding box.west)]
        \matrix (magic) [matrix of math nodes,left delimiter={[},right delimiter={]}] {
        1\\
        & \omega_2^{k'}\\
        && \omega_2^{2k'}\\
        &&&\ldots\\
        &&&&\omega_1^{k(K-1)}\omega_2^{k'(K'-1)}  \\
        &&&&& 1 \\
        &&&&&& \omega_2^{k'}\\
        &&&&&&& \omega_2^{2k'}\\
        &&&&&&&& \ldots \\
        &&&&&&&&&\omega_1^{k(K-1)}\omega_2^{k'(K'-1)}\\
        &&&&&&&&&& \ldots \\
     };
     \draw[decorate] (magic-1-1.north) -- (magic-5-5.north east) node[above=5pt,midway,sloped] {$\psi_{x,k}\otimes\psi_{y,k'}$};
     \draw[decorate] (magic-6-6.north) -- (magic-10-10.north east) node[above=5pt,midway,sloped] {$\psi_{x,k}\otimes\psi_{y,k'}$};
   \end{tikzpicture}
\end{equation}
Thus, the character table of $\psi$ is

{
\centering
\begin{tabular}{l|c|c|c|c|c|c}
  & e &$g_{(0,1)}$ & $g_{(0,2)}$ & ... & $g_{(K-1,K'-1)}$\\
$\chi_\psi$ & $N$ & $0$ & $0$  & $0$  & $0$
\end{tabular}
\par}

We see that $\psi$ has the same character table as $\psi$ the representation in image space, hence is a suited operator to use for this case. 

\subsection{Insights on the stacked shift operators model for the $2$D translation group}
\label{sec:insights1}
For the case of $2$D translations where $G=A_x \times A_y$, we first use the operator corresponding to $a_{x,k}$, the translation in $x$, then intersect a linear layer denoted $L_1$ before using the operator corresponding to $a_{y,k'}$, the translation in $y$. When we operate on the latent code, we perform
\begin{equation}
    z' = \psi_{y,k',N} L_1 \psi_{x,k,N} z
\end{equation}
where $\psi_{x,k,N}$ is the operator representing the translation in $x$. It is a matrix of order $N$, where $\psi_{x,k}$ is repeated $\frac{N}{K}$ times.
\begin{equation}
\psi_{x,k, N}:= \begin{tikzpicture}[decoration={brace,amplitude=5pt},baseline=(current bounding box.west)]
        \matrix (magic) [matrix of math nodes,left delimiter={[},right delimiter={]}] {
    1 \\
    & \omega_1^{k} \\
    && \omega_1^{2k} \\
    &&& \ldots \\
    &&&& \omega_1^{k(K-1)} \\
    &&&&& 1 \\
    &&&&&& \omega_1^{k} \\
    &&&&&&& \omega_1^{2k} \\
    &&&&&&&& \ldots \\     
    &&&&&&&&& \omega_1^{k(K-1)} \\
    &&&&&&&&&& \ldots \\
  };
    \draw[decorate] (magic-1-1.north) -- (magic-5-5.north east) node[above=5pt,midway,sloped] {$\psi_{x,k}$};
    \draw[decorate] (magic-6-6.north) -- (magic-10-10.north east) node[above=5pt,midway,sloped] {$\psi_{x,k}$};
   \end{tikzpicture}
 \label{eq:psiak}
\end{equation}
Similarly, the operator corresponding to $a_{y,k'}$ is a matrix of order $N$, where $\psi_{y,k'}$ is repeated $\frac{N}{K'}$ times
\begin{equation}
\psi_{y,k',N}:= \begin{tikzpicture}[decoration={brace,amplitude=5pt},baseline=(current bounding box.west)]
        \matrix (magic) [matrix of math nodes,left delimiter={[},right delimiter={]}] {
    1 \\
    & \omega_2^{k'} \\
    && \omega_2^{2k'} \\
    &&& \ldots \\
    &&&& \omega_2^{k'(K'-1)} \\
    &&&&& 1 \\
    &&&&&& \omega_2^{k'}\\     
    &&&&&&& \omega_2^{2k'} \\
    &&&&&&&& \ldots \\
    &&&&&&&&& \omega_2^{k'(K'-1)} \\
    &&&&&&&&&& \ldots \\
  };
    \draw[decorate] (magic-1-1.north) -- (magic-5-5.north east) node[above=5pt,midway,sloped] {$\psi_{y,k'}$};
    \draw[decorate] (magic-6-6.north) -- (magic-10-10.north east) node[above=5pt,midway,sloped] {$\psi_{y,k'}$};
   \end{tikzpicture}
    \label{eq:psiak2}
\end{equation}
We see that in order for the product $\psi_{y,k',N} L_1 \psi_{x,k,N}$ when applied to $z$, to match the result of $\psi_{k,k',N}$ applied to $z$, we need a permutation matrix $P$ that operates on a matrix of order $KK'$ made of blocks of $K'$ times $\psi_{x,k}$ and returns a matrix of order $KK'$ with:
\begin{itemize}
    \item $1$ at the first $K'$  rows and columns $c = 1,  (K+1), \ldots KK'-K + 1$
    \item $\omega_1^k$ at rows from $K'+1$ to $2K$ and columns $c = 2,  K+2, \ldots  KK'-K + 2$ etc.
    \item until $\omega_1^{k(K-1)}$ at rows from $KK'-K'$ to $KK'$ and columns $c = K, 2K, KK'$
\end{itemize}
And the layer $L_1$ would be a matrix of order $N$ made of this permutation matrix $P$, repeated in block diagonal form $\frac{N}{KK'}$ times, such that 
$(\psi_{y,k',N} L_1 \psi_{x,k,N})~z = \psi_{k,k',N}~z$.

\subsection{Semi-direct product of two groups}
\label{sec:appendixsm}
\subsubsection{Definition of a semi-direct product}

A semi-direct product $G=A \rtimes H$ of two groups $A$ and $H$ is a group such that:
\begin{itemize}
    \item $A$ is normal in $G$. 
    \item There is an homomorphism $f: H \rightarrow Aut(A)$ where $Aut(A)$ is the group of automorphism of $A$. For $a \in A$, denote $f(h)a$ by $h(a)$. In other words, $f(h)$ represents how $H$ acts on $A$.
    \item The semi-direct product $G=A \rtimes H$ is defined to be the product $A \times H$ with multiplication law
    \begin{equation}
        (a_1, h_1)(a_2, h_2) = (a_1h_1(a_2), h_1h_2)
    \end{equation}
\end{itemize}
Note that this enforces that $h_1(a_1)= h_1 a_1 h_1^{-1}$.

\subsubsection{Irreducible representation of a semi-direct product}
\label{sec:Serremethod}
In this section, we also assume $A$ is abelian. One can derive the irreducible representations of $G$ in this case, as explained in \citet{Scott:and:Serre:96} Section 8.2.\\
\\
First, consider the irreducible characters of $A$, they are of degree $1$ since $A$ is abelian and form the character group $X = Hom(A, \mathbb{C}^*)$. We use $X$ to match \citet{Scott:and:Serre:96} notation, but note that $X$ does not denote image space here, but the group of characters of $A$.  $H$ acts on this group by:
\begin{equation}
    h_{\chi}(a)=\chi(h^{-1}ah)
\end{equation}
Second, consider a system of representative of the orbits of $H$ in $X$. Each element of this system is a $\chi_i$, $i \in G/H$. For a given $\chi_i$, denote $H_i$ the stabilizer of $\chi_i$, i.e. $h_{\chi_i} = \chi_i$. This means
\begin{equation}
    h_{\chi_i}(a)=\chi_i(h^{-1}ah) = \chi_i(a), \forall a \in A
\end{equation}
Third, extend the representations of $A$ to a representation of $G_i = A \rtimes H_i$ by setting 
\begin{equation}
    \chi_i(ah) =\chi_i(a), h \in H_i, a \in A
\end{equation}
The $\chi_i$ are also characters of degree $1$ of $G_i$.\\
\\
Fourth, consider the irreducible representations of $H_i$. \citet{Scott:and:Serre:96} propose to use the irreducible representation $\rho$ of $H_i$. Combining with the canonical projection $G_i \rightarrow H_i$ we get irreducible representations $\tilde{\rho}$ of $G_i$. Irreducible representations of $G_i$ are obtained by taking the tensor product $\chi_i \otimes \tilde{\rho}$.\\
\\
Finally, the irreducible representations of $G$ are computed by taking the representation induced by $\chi_i \otimes \tilde{\rho}$.

\citet{Etingof:et:al:09} show that the character of the induced representation is
  \begin{flalign}
    \chi_{Ind^G_{G_i} (\chi_r \otimes \rho)}(a,h) = \frac{1}{|H_i|} \sum_{h' \in H s.t. h'^{-1} h h' \in H_i} \chi_r(h(a)) \chi_{\rho}(h'^{-1}h h')
 \end{flalign}
 
\subsection{Representations of the (discrete finite) Special Euclidean group}
\label{sec:appendixsm2}
In this section, we focus on the specific case of the semi-direct product $G=A \rtimes H$ where $A=A_x \times A_y$ is the group of 2D translations and $H$ the group of rotations. Hence, $A$ is abelian and $H$ is a cyclic group. We will derive the irreducible representations of this group using the method presented in the previous section \ref{sec:Serremethod}.

\subsubsection{A note on the discrete finite special Euclidean group}
\label{sec:se2note}
While the Special Euclidean group respects the structure of a semi-direct product in the continuous transformations case, we will consider its discrete and finite version. That is, we consider a finite number of translations and rotations. With integer valued translation, which is of interest when working with images (considering translations of a finite number of pixels), we cannot consider all rotations. Indeed for example rotations of $2\frac{\pi}{8}$ break the normal aspect of the subgroup of translations. 

\begin{proof}
    Take the translation element $a = (1,1)$ (one pixel translation in $x$ and $y$ respectively), and the rotation $h$ of angle $2\frac{\pi}{8}$. Consider the composition $h a h^{-1}$, applied to a point of coordinates $i,j$ this gives $h a h^{-1} (i,j) = h (h^{-1}(i,j) + (1,1)) = (i,j) + h(1,1) = (i,j) + (0, \sqrt{2})$ and the translation $(0, \sqrt{2})$ is not an integer translation. Thus, $A$ is not normal in $G$ in this case.
\end{proof}

In what follows, we consider rotations that preserve the normality of the group of integer $2$D translations of the image. Namely, these are rotations of $\frac{2\pi}{4}$, $\pi$, $\frac{3\pi}{4}$ and identity and their multiples. Nonetheless, we think the approach is insightful and approximate solutions could be found with this method. Furthermore, to ease derivations we consider that both $K \ge 2$ and $K'\ge 2$ are odd (and thus the product $KK'$ is odd), such that stabilizers of the character group of $A$ are either the entire $H$ or only the identity. We leave the exploration of even $K$ and $K'$ for future work.

\subsubsection{Finding the orbits}
\label{sec:orbits}
As we consider integer translations with periodic boundary conditions, characters of $A$ are $2$D complex numbers and are function of $\mathbb{Z}^2$ elements of the $2$D discrete and finite translation group of the form $a=(x,y)$ with $x=x_0^k$ and $y=y_0^{k'}$. The characters of this group are evaluated as $\chi_{(x_1,y_1)}(a)=e^{i2 \pi (\dfrac{x_1}{K}k+\dfrac{y_1}{K}k')}$, where $x_1, y_1$ are of the form with $x_1 \in 1 \ldots K$ and $y_1 \in 1 \ldots K'$, respectively. 

We consider two cases:
\begin{enumerate}
    \item $\chi_{0,0}$.
    \item $\chi_{(x_1,y_1)}$ with $x_1 \neq 0$ or $y_1 \neq 0$.
\end{enumerate}

Note that the total number of orbits of $H$ in $X$ is $\frac{1}{H} \sum_{\chi_x \in X} |H_x|$ where $H_x$ is the stabilizer of $\chi_x$. Either $\chi_x = e_X$ is the identity element of $X$ and $|H_x|=|H|$ or $\chi_x \neq e_X$ and the only stabilizer is $e_H$ (as shown in Section \ref{sec:proofstab}) thus $|H_x|=1$. Thus, we have that: 
\begin{equation}
    \frac{1}{H} \sum_{\chi_x \ in X} |H_x| = \frac{1}{H} (1 \times |H| + (|A| - 1) \times 1) = 1 + \frac{|A|-1}{|H|}
\end{equation} 
Hence, there are $\frac{|A|-1}{|H|}$ orbits in case 2 (the total number of orbits minus the orbit considered in the first case, $\chi_{(0,0)}$).

\subsubsection{Action of $H$ on the characters of $A$}
Elements of $H$ acts on characters of $A$ as:
\begin{flalign}
    h_{\chi_{(x_1,y_1)}}(a)&= \chi_{(x_1,y_1)}(h^{-1}ah)\\
    &=\chi_{(x_1,y_1)}(h^{-1}(a))\\
    &=e^{i 2\pi (\frac{x_1}{K} (\cos (-\theta) x - \sin (-\theta) y) + \frac{y_1}{K'} (\sin (-\theta) x + \cos (-\theta) y))}\\
    &=e^{i 2\pi ((\frac{x_1}{K} \cos (\theta) + \frac{y_1}{K'} \sin (-\theta)) x + (\frac{x_1}{K}(- \sin (-\theta)) + \frac{y_1}{K'}\cos (-\theta))y)}\\
    &=e^{i 2\pi ((\frac{x_1}{K} \cos (\theta) - \frac{y_1}{K'} \sin (\theta)) x + (\frac{x_1}{K}\sin (\theta) + \frac{y_1}{K'}\cos (\theta))y)}\\
    &=\chi_{(x_1\cos(\theta)-y_1\sin(\theta), x_1\sin(\theta)+y_1\cos(\theta))}(a)\\
    &= \chi_{h((x_1,y_1))}(a)
\end{flalign}
where $\theta$ is the angle of rotation of $h$.
\subsubsection{Case 1 $\chi_{0,0}$ (orbit of the origin)}
A representative is $\chi_{0,0}=1$. The stabilizer group $H_i$ is the entire $H$, and the irreducible representations of $H$ are of the form $e^{i 2 \pi \frac{n}{|H|}}$ for $n \in 1, \ldots, |H|$ where $|H|$ is the total number of rotations. Thus we use the tensor product $\theta_n(0,\phi)=1 \otimes e^{i 2 \pi \frac{n}{|H|}}$ as a representation of $G$. There are $|H|$ of such irreducible representations of $G$, all of degree $1$, since the group of rotation is abelian.\\
\\
The resulting representations, corresponding to the irreducible we get from combining this orbit and each one of the irreducible of $H$ can be represented in matrix form: 
\begin{equation}
\rho(a,h) := \begin{bmatrix} 
    1 & 0 & \ddots  & 0 \\
    0 & e^{i 2 \pi \frac{1}{|H|}} &  & \\
    \vdots &  & \ddots & \\
    0 &     &     & e^{i 2 \pi \frac{|H|-1}{|H|}}
    \end{bmatrix}^{j}
\end{equation}
where $j$ is such that $h=h_0^j$ and $h_0$ is the generator of the group of rotations. 

\subsubsection{Case 2 $\chi_{(x_1,y_1)}$ ($\frac{|A|-1}{|H|}$ of them)}
 A representative can be taken to be $\chi_r = \chi_{(x_1, y_1)}(x,y)$, and $x_1, y_1$ are now fixed. Its stabilizer group $H_i$ is only $\{e_H\}$ (see \citep{Berndt:2007} and proof in Section \ref{sec:proofstab}).\\
 \\
We now select $\rho$ an irreducible representation of $\{e_H\}$, and we select the trivial representation $1$ and combining with the canonical projection step $1$ will also be a representation of $G_i$. We then take the tensor product $\chi_r \otimes 1$ as a representation of $G_i = A \rtimes  \{e_H\}$.  Let us now derive the representation of the entire group $A \rtimes {G} = Ind^G_{G_i} (\chi_r \otimes 1$).

\paragraph{Induced representation $Ind^G_{G_i} (\chi_r \otimes 1)$}

First, we need a set of representatives of the left coset of $G$ in $G/G_k=A\rtimes{e_H}$. The left cosets are defined as $(a,h)G_k=\{(a,h)(a_k,e_H)~\forall a_k \in A\}$. There are $|H|$ cosets (one per $h$), which we can denote $(e_A, h) G_k$. We take a representative $g_i$ for each coset $G_k$. We take as representative $(e_A, h_i)  = (e_A, h_i) (e_A, e_H) \in (e_A,h)G_k$ \footnote{An element $(a,h)$ is in the same coset as $(e_A, h)$ as $(a,h)^{-1}(e_A,h)=(h^{-1}(a^{-1}), h^{-1})(e_A,h) = (h^{-1}(a^{-1}), e_H) \in G_k$}, so the $h_i$ are now fixed.\\
\\
The representation $(\rho, W)$ is induced by ($\chi_r \otimes 1$, $V$) if
\begin{equation}
    W = \bigoplus_{i=1}^{|H|} \rho(g_i) V
\end{equation}
with $g_i=(e_A,h_i)$ and $\rho(g_i)$ is described below.\\
\\
For each $g$ and each $g_i$, there is $j(i) \in \{1, \ldots |H| \}$ and $f_j \in G_k$ such that we have $g g_i = g_{j(i)}f_j$. Indeed:
\begin{flalign}
    g g_i &= (a, h) (e_A, h_i) \\
    &= (a, h h_i)\\
    & = (e_A, h h_i)((hh_i)^{-1}(a), e_H)
\end{flalign}
so $g_{j(i)} = (e_A, h h_i)$. In other words, the action of $g=(a,h)$ on an element $w \in W$ permutes the representatives:
\begin{flalign}
    \rho(g) w = \rho((a,h)) \sum_{i=1}^{|H|} \rho(g_i) v_i = \sum_{i=1}^{|H|} \rho(g_{j(i)}) \chi((hh_i)^{-1}(a)) v_i
\end{flalign}
Now, we need to find for each element of $G$, the resulting permutation of the coset representatives. $G$ is generated by $a \in A, h \in H$: any element $(a,h) \in G$ can be written as $(a, e_H)(e_A, h)$. For $(a, e_H)$, we get $g_{j(i)} = (e_A, h_i) = g_i$. The induced representations are thus: 
\begin{equation}
\rho(a, e_H) = \begin{bmatrix} 
    \chi_r(h_1^{-1}(a)) & ... & 0 & 0 \\
    0 & \chi_r(h_2^{-1}(a)) & ... & 0 \\
    0 & ... & 0 & \chi_r(h_{|H|}^{-1}(a))
    \end{bmatrix}
\end{equation}
with no permutation. For $(e_A, h)$ we get $g_{j(i)} = (e_A, h h_i)$. The resulting induced representations are:
\begin{equation}
\rho(e_A, h) = \begin{bmatrix} 
    0 & 0 & ... & \chi_r((h h_{j^{-1}(1)})^{-1}(e_A))\\
    0 & \chi_r((hh_{j^{-1}(2)})^{-1}(e_A)) & ...\\
    0 & ... & \chi_r((hh_{j^{-1}(|H|)})^{-1}(e_A)) & 0
    \end{bmatrix}
\end{equation}
i.e.
\begin{equation}
\rho(e_A, h) = P_h = \begin{bmatrix} 
    0 & 0 & ... & 1\\
    0 & 1 & ...\\
    0 & ... & 1 & 0
    \end{bmatrix}
\end{equation}

Where the permutation matrix $P_h$ above represents how $h$ acts on its own group. If $hh_i=h_j$ then $P_h$ has a $1$ at the column $j$ of its $i$-th row. For example, if $hh_{|H|}=h_1$ then $j(|H|) = 1$, $j^{-1}(1) = |H|$ and so for the row $1$, there is $\chi_r((h h_{H})^{-1}(e_A)) = 1$ at the $|H|$-th column as above.\\
\\
Let us denote for clarity $\chi_{r,i} = \chi_r(h_i^{-1}(a))$. The representations of $(a,h)$ will be 
\begin{equation}
\rho(a, h) = \rho(a, e_H)\rho(e_A, h) = 
    \begin{bmatrix} 
    \chi_{r,1} & ... & 0 & 0 \\
    0 & \chi_{r,2}& ... & 0 \\
    0 & ... & 0 & \chi_{r,|H|}
    \end{bmatrix}
    P_h
    =
    \begin{bmatrix} 
    0 & 0 & ... & \chi_{r,1}\\
    0 & \chi_{r,2} & ...\\
    0 & ... & \chi_{r,|H|} & 0
    \end{bmatrix}
\end{equation}
For each orbit, we get one of these (of degree $H$).

\subsubsection{Resulting representation}

The resulting representation will be a block diagonal matrix of size $|A||H|$ of the form 
\begin{small}
\begin{equation}
   \rho(a,h):=\begin{tikzpicture}[decoration={brace,amplitude=5pt},baseline=(current bounding box.west)]
     \matrix (magic) [matrix of math nodes,left delimiter={[},right delimiter={]}] {
        1\\
        & e^{i 2 \pi \frac{1}{|H|}j}\\
        && ...\\
        &&& e^{i 2 \pi \frac{H-1}{|H|}j}\\
        &&&&M_{r_1}(a)P_h\\
        &&&&&\ldots\\
        &&&&&&M_{r_1}(a)P_h\\
        &&&&&&& \ldots \\
        &&&&&&&& M_{R}(a) P_h\\
        &&&&&&&&& \ldots\\
        &&&&&&&&&& M_{R}(a) P_h\\
     };
     \draw[decorate] (magic-5-5.north) -- (magic-7-7.north east) node[above=5pt,midway,sloped] {$|H|$ copies};
     \draw[decorate] (magic-9-9.north) -- (magic-11-11.north east) node[above=5pt,midway,sloped] {$|H|$ copies};
   \end{tikzpicture}
\end{equation}
\end{small}
The first $|H|$ elements on the diagonal correspond to the irreducible representation of $h=h_0^j$, this is the shift operator we have been considering in the single transformation case. Second, each matrix product $M_r(a,h)P_h$, representative $r$, is repeated $|H|$ times as it is of degree $|H|$ (see calculation of degree in the Section \ref{sec:degreescalculation}) and corresponds to:
\begin{equation}
    M_r(a) = \begin{bmatrix} 
    \chi_r(h_1^{-1}(a))& ... & 0 & 0 \\
    0 & \chi_r(h_2^{-1}(a)) & ... & 0 \\
    0 & ... & 0 & \chi_r(h_{|H|}^{-1}(a))
    \end{bmatrix}
\end{equation}

If we assume $|A||H|$ divides $N$, we can use an operator $\psi_{a,h,N}$ that is the repetition of  $\frac{N}{|A||H|}$ times $\rho(a,h)$ into a matrix of order $N$:
\begin{equation}
   \psi_{a,h,N}:=
   \begin{tikzpicture}[decoration={brace,amplitude=5pt},baseline=(current bounding box.west)]
     \matrix (magic) [matrix of math nodes,left delimiter={[},right delimiter={]}] {
        1\\
        & e^{i 2 \pi \frac{1}{|H|}j}\\
        && \ldots \\
        &&& M_{R}(a) P_h\\
        &&&& 1\\
        &&&&& e^{i 2 \pi \frac{1}{|H|}j}\\
        &&&&&& \ldots\\
        &&&&&&& M_{R}(a) P_h\\
        &&&&&&&& \ldots\\
     };
     \draw[decorate] (magic-1-1.north) -- (magic-4-4.north east) node[above=5pt,midway,sloped] {$\rho(a,h)$};
     \draw[decorate] (magic-5-5.north) -- (magic-8-8.north east) node[above=5pt,midway,sloped] {$\rho(a,h)$};
   \end{tikzpicture}
 \label{eq:smoperator}
\end{equation}

Its character table is:

{
\centering
\begin{tabular}{l|c|c|c|c|c|c}
  & $(e_A,e_H)$ & $(a_1,e_H)$ & $(a_2, e_H)$ & ... & $(a_{|A|}, h_{|H|})$\\
$\chi_\psi$ & $N$ & $0$ & $0$  & $0$  & $0$
\end{tabular}
\par}
and is the same character as the operator acting on image space, as shown with the computation of the character of $\rho(a,h)$ in Section \ref{sec:chism}. 

\subsubsection{Character table} 
\label{sec:chism}
For the identity element $(e_A,e_H)$, we get for $\rho(a,h)$ the identity matrix, and its trace is $|A||H|$. When $h \neq e_H$, $P_h$ has no diagonal element and $\sum_{k=0}^{|H|-1} e^{i 2 \pi \frac{k}{|H|} k'} = 0$, hence the trace of $\rho(a,h)=0$. When $a \neq e_A$, if $h \neq e_H$ we are in the previous case. If $h = e_H$, $P_H$ is the identity matrix and the representation will be 
\begin{equation}
 \rho(a,e_H) = 
    \begin{bmatrix}
        1\\
        & 1\\
        && ...\\
        &&& 1\\
        &&&&M_{r_1}(a)I\\
        &&&&&M_{r_1}(a)I\\
        &&&&&&\\
        &&&&&&...\\
        &&&&&&&&M_{R}(a)I
    \end{bmatrix}
\end{equation}
The trace of $\rho(a,e_H)$ is 
\begin{flalign}
    Tr(\rho(a,e_H)) &= |H| + |H| \sum_{h_i \in H} \sum_{r=r_1}^{R} \chi_r(h_i^{-1}(a))
\end{flalign}
as each block-diagonal matrix is repeated $|H|$ times. If we interchange the order of summation we get:
\begin{flalign}
    Tr(\rho(a,e_H)) &=  |H| + |H| \sum_{r=r_1}^{R} \sum_{h_i \in H} \chi_r(h_i^{-1}(a)))\\
    &=  |H| + |H|\sum_{r=r_1}^{R} \sum_{h_i \in H} h_{i,\chi_r}(a))
\end{flalign}
where $h_{i,\chi_r}$ represents the action of $h_i$ on $\chi_r$. The action of every $h_i$ on $\chi_r$ gives all the elements in the orbit of $\chi_r$, so the double sum results in the sum over all characters $\chi \in X$ of $A$, apart from the orbit of $\chi_{0,0}$ which contains only $\chi_{0,0}$. 
\begin{flalign}
    Tr(\rho(a,e_H)) &= |H| + |H|\sum_{\chi \in X, \chi \neq X_{0,0}} \chi(a)\\
\end{flalign}
The sum of the all the irreducible characters, for $a \neq e_A$ is $0$ \citep[Corollary 2]{Scott:and:Serre:96}: $\sum_{\chi \in X}\chi(a) = 0$. Hence,
\begin{flalign}
    \sum_{\chi \in X}\chi(a) = 0 &= X_{0,0}(a) + \sum_{\chi \in X, \chi \neq X_{0,0}} \chi(a)\\
    &= 1 + \sum_{\chi \in X, \chi \neq X_{0,0}} \chi(a)\\
    & \rightarrow  \sum_{\chi \in X, \chi \neq X_{0,0}} \chi(a) = -1.
\end{flalign}
Hence,
\begin{flalign}
    Tr(\rho(a,e_H)) &= |H| + |H|(-1) = 0.
\end{flalign}

Consequently, the character table of $\rho$ is the following:

{
\centering
\begin{tabular}{l|c|c|c|c|c|c}
  & $(e_A,e_H)$ & $(a_1,e_H)$ & $(a_2, e_H)$ & ... & $(a_{|A|}, h_{|H|})$\\
$\chi_\rho$ & $|A||H|$ & $0$ & $0$  & $0$  & $0$
\end{tabular}
\par}

The character of $\psi$ is $\frac{N}{|A||H|}$ times the character of $\chi_\rho$, i.e. 

{
\centering
\begin{tabular}{l|c|c|c|c|c|c}
  & $(e_A,e_H)$ & $(a_1,e_H)$ & $(a_2, e_H)$ & ... & $(a_{|A|}, h_{|H|})$\\
$\chi_\psi$ & $N$ & $0$ & $0$  & $0$  & $0$
\end{tabular}
\par}

\subsection{Insights on the stacked shift operators model for the discrete finite Special Euclidean group}
\label{sec:insights2}

Similarly to $2$D translation case, we can use the theoretical form of the representation to use to gain insight on what the intersected layers $L_i$ should be for the case of $2$D translations in conjunction with rotations. Elements of this group are $(a, h)$ where $a$ is the $2$D translation, composed of $a_{x,k}$ the translation on the $x$-axis, $a_{y,k'}$ translation on the $y$-axis, and $h$ the rotation. Using the stacked shift operators model, we first use the operator corresponding to the rotation, then the one for translation in $x$, then the one for $y$-translation. The resulting transformed latent code is
\begin{equation}
    z'=\psi_{y,k',N} L_2 \psi_{x,k,N} L_1 \psi_{h,j,N}z
\end{equation}
with $a_{y,k'}=a_{y,0}^{k'},~a_{x,k}=a_{x,0}^{k},~h=h_0^j$. The operators for translations $\psi_{x,k,N}, ~\psi_{y,k',N}$ are described in Equations \ref{eq:psiak} and  \ref{eq:psiak2}. We repeat them here for clarity.
\begin{small}
\begin{equation}
\psi_{a_{x,k}, k, N}:= \begin{tikzpicture}[decoration={brace,amplitude=5pt},baseline=(current bounding box.west)]
        \matrix (magic) [matrix of math nodes,left delimiter={[},right delimiter={]}] {
    1 \\
    & \omega_1^{k} \\
    && \omega_1^{2k} \\
    &&& \ldots \\
    &&&& \omega_1^{k(K-1)} \\
    &&&&& 1 \\
    &&&&&& \omega_1^{k} \\
    &&&&&&& \omega_1^{2k} \\
    &&&&&&&& \ldots \\     
    &&&&&&&&& \omega_1^{k(K-1)} \\
    &&&&&&&&&& \ldots \\
  };
    \draw[decorate] (magic-1-1.north) -- (magic-5-5.north east) node[above=5pt,midway,sloped] {$\psi_{x,k}$};
    \draw[decorate] (magic-6-6.north) -- (magic-10-10.north east) node[above=5pt,midway,sloped] {$\psi_{x,k}$};
   \end{tikzpicture}
\end{equation}
\end{small}
\begin{small}
\begin{equation}
\psi_{y,k',N}:= \begin{tikzpicture}[decoration={brace,amplitude=5pt},baseline=(current bounding box.west)]
        \matrix (magic) [matrix of math nodes,left delimiter={[},right delimiter={]}] {
    1 \\
    & \omega_2^{k'} \\
    && \omega_2^{2k'} \\
    &&& \ldots \\
    &&&& \omega_2^{k'(K'-1)} \\
    &&&&& 1 \\
    &&&&&& \omega_2^{k'}\\     
    &&&&&&& \omega_2^{2k'} \\
    &&&&&&&& \ldots \\
    &&&&&&&&& \omega_2^{k'(K'-1)} \\
    &&&&&&&&&& \ldots \\
  };
    \draw[decorate] (magic-1-1.north) -- (magic-5-5.north east) node[above=5pt,midway,sloped] {$\psi_{y,k'}$};
    \draw[decorate] (magic-6-6.north) -- (magic-10-10.north east) node[above=5pt,midway,sloped] {$\psi_{y,k'}$};
   \end{tikzpicture}
\end{equation}
\end{small}
where $\omega_1 = e^{\frac{2 i \pi}{K}}$ and $\omega_2 = e^{\frac{2 i \pi}{K'}}$. And $\psi_{h,j,N}$ is the repetition of the shift operator for the rotation $\frac{N}{|H|}$ times, as follows,
\begin{small}
\begin{equation}
\psi_{h,j,N}:=\begin{tikzpicture}[decoration={brace,amplitude=5pt},baseline=(current bounding box.west)]
        \matrix (magic) [matrix of math nodes,left delimiter={[},right delimiter={]}] {
    1 \\
    & e^{i 2 \pi \frac{j}{|H|}} \\
    && e^{i 2 \pi \frac{2j}{|H|}}\\
    &&& \ldots \\
    &&&&e^{i 2 \pi \frac{j(|H|-1)}{|H|}}\\
    &&&&& 1 \\
    &&&&&& e^{i 2 \pi \frac{j}{|H|}}\\     
    &&&&&&& e^{i 2 \pi \frac{2j}{|H|}} \\
    &&&&&&&& \ldots \\
    &&&&&&&&& e^{i 2 \pi \frac{j(|H|-1)}{|H|}} \\
    &&&&&&&&&& \ldots \\
  };
    \draw[decorate] (magic-1-1.north) -- (magic-5-5.north east) node[above=5pt,midway,sloped] {$\psi_{h,j}$};
    \draw[decorate] (magic-6-6.north) -- (magic-10-10.north east) node[above=5pt,midway,sloped] {$\psi_{h,j}$};
   \end{tikzpicture}
\end{equation}
\end{small}
The representations for each transformation $\psi_{x,k,N}, ~\psi_{y,k',N}, ~\psi_{h,j,N}$ are linked with the representation that the theory gives us in Equation \ref{eq:smoperator}. \\
\\
First, recall that elements of $H$ acts on a character $\chi_r=\chi_{(x_1,y_1)}$:
\begin{flalign}
    \chi_{(x_1,y_1)}(h_i^{-1}(a))= \chi_{h_i((x_1,y_1))}(a)
\end{flalign}
Thus when $h_i$ spans $H$ in a given matrix $M_r$, we obtain $|H|$ distinct characters built from $\chi_r$, i.e.
\begin{equation}
    M_r(a) = \begin{bmatrix} 
    \chi_{h_1(r)}(a))& ... & 0 & 0 \\
    0 & \chi_{h_2(r)}(a) & ... & 0 \\
    0 & ... & 0 & \chi_{h_{|H|}(r)}(a)
    \end{bmatrix}
\end{equation}
For another matrix $M_{r'}$, we obtain again $|H|$ distinct orbits, distinct from the one present in $M_r$ (otherwise they would be in the same orbit). We have $\frac{A-1}{H}$ representatives (excluding $\chi_{(0,0)}$), thus $|A|-1$ characters of $A$ evaluated at $a$ are obtained by considering all the matrices $M_r$. The remaining character is $\chi_{(0,0)}=1$. Hence, the diagonal matrix $\rho(a,e_H)$ of order $|A||H|$ contains all the characters of $A$ evaluated at the element $a$, each one repeated $|H|$ times since each matrix $M_r$ is repeated $|H|$ times, and the first diagonal elements of $\rho(a,e_H)$ are $1$. \\

So we see that what we need are $|H|$ times each character of $A$. As explained in Section \ref{sec:orbits}, these characters are of the form:
\begin{equation}
\chi_{(x_1,y_1)}(a)=e^{i(2 \pi \dfrac{x_1}{K}k+ 2 \pi \dfrac{y_1}{K'}k')}=e^{i(2 \pi \dfrac{x_1}{K}k)}e^{i (2 \pi \dfrac{y_1}{K'}k')}=\omega_1^{x_1k}\omega_2^{y_1k'}.
\end{equation}
The characters of each translation group are $\chi_{x_1}=\omega_1^{x_1}$ for the $x$-translation and $\chi_{y_1}=\omega_2^{y_1}$ for the $y$-translation, thus:
\begin{equation}
\chi_{(x_1,y_1)}(a)=\chi_{(x_1,y_1)}(a_{x,k},a_{y,k'})=\omega_1^{x_1k}\omega_2^{y_1k'}=\chi_{x_1}(a_{x,k})\chi_{y_1}(a_{y,k'})
\end{equation}
Since $\psi_{x,k}$ is represented $\frac{N}{K}$ times in $\psi_{x,k,N}$ and $\psi_{y,k'}$ is represented $\frac{N}{K'}K'$ times in $\psi_{y,k',N}$, we can take $L_2$ to reorder $\psi_{y,k',N}$ into blocks of $K'$ repeated diagonal elements such that when multiplied with $\psi_{x,k,N}$ by doing $\psi_{y,k',N} L_2 \psi_{x,k,N}$, we obtain $\frac{N}{KK'} = \frac{N}{|A|}$ blocks of size $|A|$, each containing every character of $A$ once. Furthermore, we can see that the resulting matrix $\psi_{y,k',N} L_2 \psi_{x,k,N}$ is composed of $\frac{N}{|A||H|}$ blocks of size $|A||H|$ containing every character of $A$, but repeated $|H|$ times.\\
\\
This resulting matrix is then multiplied with the matrix $L_1 \psi_{h,j,N}$. So let us now turn to the representation of the rotation $h$, that is $\psi_{h,j,N}$. It is composed of the elements that appear in the upper-left diagonal of $\rho(a,h)$. But we also need to make the permutation matrices $P_h$ appear. Recall that diagonal matrix of order $|H|$ that corresponds to the shift representation of $h=h_0^j$ (not repeated $\frac{N}{|H|}$ times) is
\begin{equation}
    \psi_{h,j}=\begin{bmatrix}
    1 \\
    & e^{i 2 \pi \frac{j}{|H|}} \\
    && e^{i 2 \pi \frac{2j}{|H|}}\\
    &&& \ldots \\
    &&&&e^{i 2 \pi \frac{j(|H|-1)}{|H|}}\\
    \end{bmatrix}
\end{equation}
If we right-multiply $\psi_{h,j}$ by 
\begin{equation}
    C=\begin{bmatrix}
    1 & 1 & \ldots & 1\\
    1 & e^{i 2 \pi \frac{1}{|H|}} & \ldots & e^{i 2 \pi \frac{|H|-1}{|H|}}\\
    &&\ldots&\\
    1 & e^{i 2 \pi \frac{(|H|-1)}{|H|}} & \ldots & e^{i 2 \pi \frac{(|H|-1)(|H|-1)}{|H|}}\\
    \end{bmatrix}
\end{equation}
and left-multiply it by 
\begin{equation}
    B=\begin{bmatrix}
    1 & 1 & \ldots & 1\\
    1 & e^{ -i 2 \pi \frac{1}{|H|}} & \ldots & e^{-i 2 \pi \frac{|H|-1}{|H|}}\\
    &&\ldots&\\
    1 & e^{-i 2 \pi \frac{(|H|-1)}{|H|}} & \ldots & e^{-i 2 \pi \frac{(|H|-1)(|H|-1)}{|H|}}\\
    \end{bmatrix}
\end{equation}
we obtain a matrix that is filled with $0$ except for each row $r$, one time the value $|H|$ at the column $c$ where $c$ such that $h_0^{-r} h h_0^{c}=e_H$, i.e. $h h_0^c=h_0^r$, this is exactly what $P_h$ represents, as we defined it such that if $h h_i = h_j$ then $P_h$ has a $1$ at the column $j$ of its $i$-th row. Thus,
\begin{equation}
   P_h = \dfrac{1}{|H|}B\psi_{h,j}C 
\end{equation}
Consider a block diagonal matrix $M$ where the first $|H|$ elements are all ones (on the diagonal), and then the matrix $B$ is repeated on the block diagonal $|A|-1$ times. 
\begin{equation}
   M=\begin{tikzpicture}[decoration={brace,amplitude=5pt},baseline=(current bounding box.west)]
     \matrix (magic) [matrix of math nodes,left delimiter={[},right delimiter={]}] {
        1\\
        & 1\\
        && \ldots \\
        &&& 1\\
        &&&& B\\
        &&&&& B\\
        &&&&&& \ldots\\
        &&&&&&& B \\
     };
     \draw[decorate] (magic-1-1.north) -- (magic-4-4.north east) node[above=5pt,midway,sloped] {$|H|$ times };
     \draw[decorate] (magic-5-5.north) -- (magic-8-8.north east) node[above=5pt,midway,sloped] {$|A| - 1$ times };
   \end{tikzpicture}
\end{equation}

Consider $M'$ that is the repetition of $|A|$ times $\psi_{h,j}$
\begin{equation}
   M'=\begin{tikzpicture}[decoration={brace,amplitude=5pt},baseline=(current bounding box.west)]
     \matrix (magic) [matrix of math nodes,left delimiter={[},right delimiter={]}] {
        1\\
        & e^{i 2 \pi \frac{1}{|H|}j}\\
        && \ldots \\
        &&& e^{i 2 \pi \frac{|H|-1}{|H|}j}\\
        &&&& 1\\
        &&&&& e^{i 2 \pi \frac{1}{|H|}j}\\
        &&&&&& \ldots\\
        &&&&&&& e^{i 2 \pi \frac{|H|-1}{|H|}j}\\
        &&&&&&&& \ldots\\
     };
     \draw[decorate] (magic-1-1.north) -- (magic-4-4.north east) node[above=5pt,midway,sloped] {$\psi_{h,j}$ (of order $|H|$)};
     \draw[decorate] (magic-5-5.north) -- (magic-8-8.north east) node[above=5pt,midway,sloped] {$\psi_{h,j}$};
   \end{tikzpicture}
\end{equation}
$M'$ is of order $|A||H|$. Thus, 
\begin{equation}
   MM'C=\frac{1}{|H|}\begin{tikzpicture}[decoration={brace,amplitude=5pt},baseline=(current bounding box.west)]
     \matrix (magic) [matrix of math nodes,left delimiter={[},right delimiter={]}] {
        1\\
        & e^{i 2 \pi \frac{1}{|H|}j}\\
        && \ldots \\
        &&& e^{i 2 \pi \frac{|H|-1}{|H|}j}\\
        &&&& P_h\\
        &&&&& P_h\\
        &&&&&& \ldots\\
        &&&&&&& P_h\\
     };
     \draw[decorate] (magic-1-1.north) -- (magic-4-4.north east) node[above=5pt,midway,sloped] {$|H|$ elements};
     \draw[decorate] (magic-5-5.north) -- (magic-8-8.north east) node[above=5pt,midway,sloped] {$(A-1)$ times};
   \end{tikzpicture}
   \label{eq:result1}
\end{equation}
which is of order $|A||H|$. Assuming the encoder takes care of the right-multiplication by $C$ and the scaling by $|H|$ (which it can learn to do), we do not write the scaling and right-multiplication by $C$.\\
\\
The operator $\psi_{h,j,N}$ is $\frac{N}{|H|}$ repetitions of $\psi_{h,j}$, hence it can be seen as $\frac{N}{|A||H|}$ repetitions of $M'$. Thus, let us denote $Q$ a $N$ by $N$ block diagonal matrix form of $\frac{N}{|A||H|}$ repetitions of $M$. When multiplying $Q$ with $\psi_{h,j,N}$ we can get $MM'$ repeated $|A||H|$ times:
\begin{equation}
  Q \psi_{h,j,N}=\begin{tikzpicture}[decoration={brace,amplitude=5pt},baseline=(current bounding box.west)]
     \matrix (magic) [matrix of math nodes,left delimiter={[},right delimiter={]}] {
        1\\
        & e^{i 2 \pi \frac{1}{|H|}j}\\
        && \ldots \\
        &&& P_h\\
        &&&& \ldots\\
        &&&&& P_h\\
        &&&&&& 1\\
        &&&&&&& e^{i 2 \pi \frac{1}{|H|}j}\\
        &&&&&&&& \ldots\\
        &&&&&&&&& P_h\\
        &&&&&&&&&& \ldots\\
     };
     \draw[decorate] (magic-1-1.north) -- (magic-6-6.north east) node[above=5pt,midway,sloped] {MM'};
     \draw[decorate] (magic-7-7.north) -- (magic-10-10.north east) node[above=5pt,midway,sloped] {MM'};
   \end{tikzpicture}
   \label{eq:result1}
\end{equation}
This is the matrix that is multiplied with $\psi_{y,k',N} L_2 \psi_{x,k,N}$. However, the order of the rows in the resulting does not correspond to the ordering of the characters in $\psi_{y,k',N} L_2 \psi_{x,k,N}$. If we want to match the operator in Equation  \ref{eq:smoperator} such that 
\begin{equation}
     \psi_{a,h,N} z = (\psi_{y,k',N} L_2 \psi_{x,k,N} L_1 \psi_{h,j,N}) z
\end{equation}
We need $L_1$ be the product of two matrices $P$ and $Q$, such that $P$ reorders the rows of the vector $Q \psi_{h,j,N} z$ to match the ordering of the characters in $\psi_{y,k',N} L_2 \psi_{x,k,N}$. The result of 
\begin{equation}
     (\psi_{y,k',N} L_2 \psi_{x,k,N}) P Q \psi_{h,j,N} z
\end{equation}
will be a vector that is a permuted version of the vector we would get with $\psi_{a,h,N} z$, and a linear decoder can learn to reorder it if we want to exactly match $\psi_{a,h,N} z$.

\subsection{Details of Section \ref{sec:appendixsm2}}
\subsubsection{Stabilizers of orbits Case 2}
\label{sec:proofstab}
We show here that for $\chi_r = \chi_{(x_1, y_1)}(x,y)$ with $x_1 \neq 0$ or $y_1 \neq 0$ the stabilizer group $H_i$ is only $\{e_H\}$.
\begin{proof}
Let us consider the stabilizer group of $\chi_{(x_1,y_1)}$. It is composed of the elements $h \in H$ such that for $a \in A$, $h_{\chi_{(x_1,y_1)}}(a)= \chi_{h((x_1,y_1))}(a) = \chi_{(x_1,y_1)}(a)$. In other words, we must have that the character of the rotated vector $h((x_1,y_1))$ is the same as the character corresponding to $(x_1,y_1)$, for any $a$. Seeing characters as $2$D complex numbers, we must have $(x_1,y_1)=(\cos \theta x_1 - \sin \theta y_1, \sin \theta x_1 + \cos \theta y_1)$, i.e. the rotated vector corresponding to the character is equal to itself. As we employ boundary conditions, this is for example the case for $\theta=\pi$ or $\theta=-\pi$ for $x_1=\frac{K}{2}, y_1=\frac{K'}{2}$, since the inverse of $(\frac{K}{2},\frac{K'}{2})$ is itself. But we restrict ourselves and we consider only odd $K$ and $K'$, such that there is no angle $\pi$ such that the result of the rotation of $(x_1,y_1)$ gives the same vector in complex space.
\end{proof}

\subsubsection{Degrees of the representations}
\label{sec:degreescalculation}
\paragraph{Case 1}
In Case $1$, we have $|H|$ representations of degree $1$. 
\paragraph{Case 2}
Recall that in Case $2$, $H_i = \{e_H\}$ and we use $\rho=1$, thus the character of the induced representation  is 
  \begin{flalign}
    \chi_{Ind^G_{G_i} (\chi_r \otimes 1)}(a,h) = \sum_{h' \in H s.t. h'^{-1} h h' = e_H} \chi_r(h(a)) \chi_1(e_H)
 \end{flalign}
$h'^{-1} h h' = e_H$ means $h'^{-1} h = h'^{-1}$. If $h=e_H$, $h'$ will span the entire $H$ and we have 
  \begin{flalign}
    \chi_{Ind^G_{G_i} (\chi_r \otimes 1)}(a,e_H) = \sum_{h' \in H} \chi_r(a) \chi_1(e_H) = |H| \chi_r(a)
 \end{flalign}
 If $h \neq e_H$ then there is no element $h'$ for which this is true. Indeed, otherwise it means $h=h' h'^{-1}=e_H$ leading to a contradiction. Thus:
  \[
    \chi_{Ind^G_{G_i} (\chi_r \otimes 1)}(a,h)=\left\{
                \begin{array}{ll}
                  |H| \chi_r(a)~\text{if}~h = e_H\\
                  0~\text{if}~h \neq e_H\\
                \end{array}
              \right.
  \]
  
To obtain the degree of the representation, we calculate the character of the identity element.
\begin{flalign}
    \chi_{Ind^G_{G_i} (\chi_r \otimes 1)}(e_A,e_H) &=  \sum_{h' \in H s.t. h'^{-1} e_H h' = e_H} \chi_r(e_A) \chi_1(e_H)\\
    & = \sum_{h' \in H} \chi_r(e_A) \chi_1(e_H) = |H| \chi_r(e_A) \chi_1(e_H) = |H| \chi_1(e_H) = |H|
 \end{flalign}
 where we use the fact $\chi_r(e_A) = 1$ ($A$ is abelian hence its irreducible characters are of degree $1$) and that $h'^{-1} e_H h' = e_H$ is true for all $h'\ \in H$. Hence, the degree of the induced representation is $|H| \chi_1(e_H) = |H|$ since $\chi_1(e_H) = |\{e_H\}|=1$.\\
 \\
Thus in case 2 each orbit induces a unique induced representation of degree $|H|$, and we have $\frac{|A|-1}{|H|}$ orbits in the second case. This shows we derived all the irreducible representations of $G$, as if we do the sum of the degree squared of irreducible representations we have
 \begin{equation}
     |H| + \frac{|A|-1}{|H|} |H|^2 = |H|+|A||H|-|H|=|A||H|
 \end{equation}
 which correctly equals the order of $G$, see \citet[Corollary 2]{Scott:and:Serre:96}.
 
 \section{Additional Results}
\label{sec:addresults}
\subsection{Quantifications}
\label{sec:quantresults}

\paragraph{Test Mean Squared Error} Table \ref{tab:mse} reports test MSE for disentangled, supervised shift and weakly supervised shift operators.

\begin{table}[h!]
\caption{Test mean square error (MSE) $\pm$ standard deviation of the mean over random seeds. Numbers in $()$ refer to the number of rotations, the number of translations on the $x$-axis, and the number of translations on the $y$-axis in this order. The case of multiple transformation does not apply to the weakly supervised shift operator, and we did not experiment on translated MNIST with the weakly shift operator as its performance on translated simple shapes and rotated MNIST were showing its relevance already.}
\begin{center}
\begin{tabular}{l|c|c|c}
\bf Dataset  &\multicolumn{3}{c}{\bf Model}\\
~ & Disentangled & Shift & Weak. sup. shift ($K_L=10$)\\
Shapes (10,0,0) & $0.0208 \pm 5.2\mathrm{e}{-6}$ & $\mathbf{0.0002 \pm 1.8\mathrm{e}{-6}}$  & $0.001 \pm 1.3\mathrm{e}{-3}$ \\
Shapes (0,10,0) & $0.0352 \pm 9.0\mathrm{e}{-6}$ & $\mathbf{0.0052 \pm 9.6\mathrm{e}{-6}}$ & $0.0097 \pm 5.2\mathrm{e}{-3}$ \\
Shapes (0,0,10) & $0.0353 \pm 7.6\mathrm{e}{-6}$ & $\mathbf{0.0052 \pm 1.5\mathrm{e}{-5}}$ & $0.0115 \pm 4.9\mathrm{e}{-3}$ \\
Shapes (0,5,5) & n/a & $0.0047 \pm 1.9\mathrm{e}{-5}$ & n/a \\
Shapes (4,5,5) & n/a & $0.0049 \pm 7.7\mathrm{e}{-6}$ & n/a \\
Shapes (5,5,5) & n/a & $ 0.0021 \pm 5.9\mathrm{e}{-6}$ & n/a \\
MNIST (10,0,0) & $0.0660 \pm 8.2\mathrm{e}{-5}$ &  $\mathbf{0.0004 \pm 5.6\mathrm{e}{-6}}$ & $0.0035 \pm 3.9\mathrm{e}{-3}$\\
MNIST (0,10,0) & $0.0838 \pm 3.2\mathrm{e}{-5}$ & $\mathbf{0.0079 \pm 5.1\mathrm{e}{-5}}$ & n/a \\
MNIST (0,0,10) & $0.0857 \pm 5.1\mathrm{e}{-5}$ & $\mathbf{0.0062 \pm 3.2\mathrm{e}{-5}}$ & n/a \\
MNIST (4,5,5) & n/a & $0.004 \pm 4.0\mathrm{e}{-5}$ & n/a
\end{tabular}
\end{center}
\label{tab:mse}
\end{table}

\paragraph{LSBD Disentanglement Measure} We compute the LSBD disentanglement metric to quantify how well the shift operator captures the factors of variation compared to the disentangled operator (see \citet{anonymous2021quantifying}). Note traditional disentanglement metrics are not appropriate as they describe how well factors of variation are restricted to subspaces in contrast to our proposed framework using distributed latent operators. LSBD, on the other hand, measures how well latent operators capture each factor variation to quantify disentanglement even for distributed operators. Using LSBD, we  quantify the advantage of the shift operator with LSBD of 0.0020 versus the disentangled operator with LSBD of 0.0106 for the models in Fig. \ref{fig:fig3}A and \ref{fig:fig3}B.

\subsection{Additional analyses}

\paragraph{Latent Variance and PCA Analysis} The analysis in Fig. \ref{fig:fig1}C and \ref{fig:fig1}D quantitatively measures disentanglement in the latent representation for VAE and variants on rotated MNIST. We first apply every transformation to a given shape and compute the variance of the latent representation as we vary the transformation. The final figure shows the average variance across all test samples. For the PCA analysis, we seek to determine whether the transformation acts on a subspace of the latent representations. We first compute the ranked eigenvalues of the latent representations of each shape with all transformations applied to the input. We then normalize the ranked eigenvalues by the sum of all eigenvalues to obtain the proportion of variance explained each latent dimension. Finally, we plot the average of the normalized ranked eigenvalues across all test samples. 

\begin{figure}[ht]
    \includegraphics[width=\textwidth]{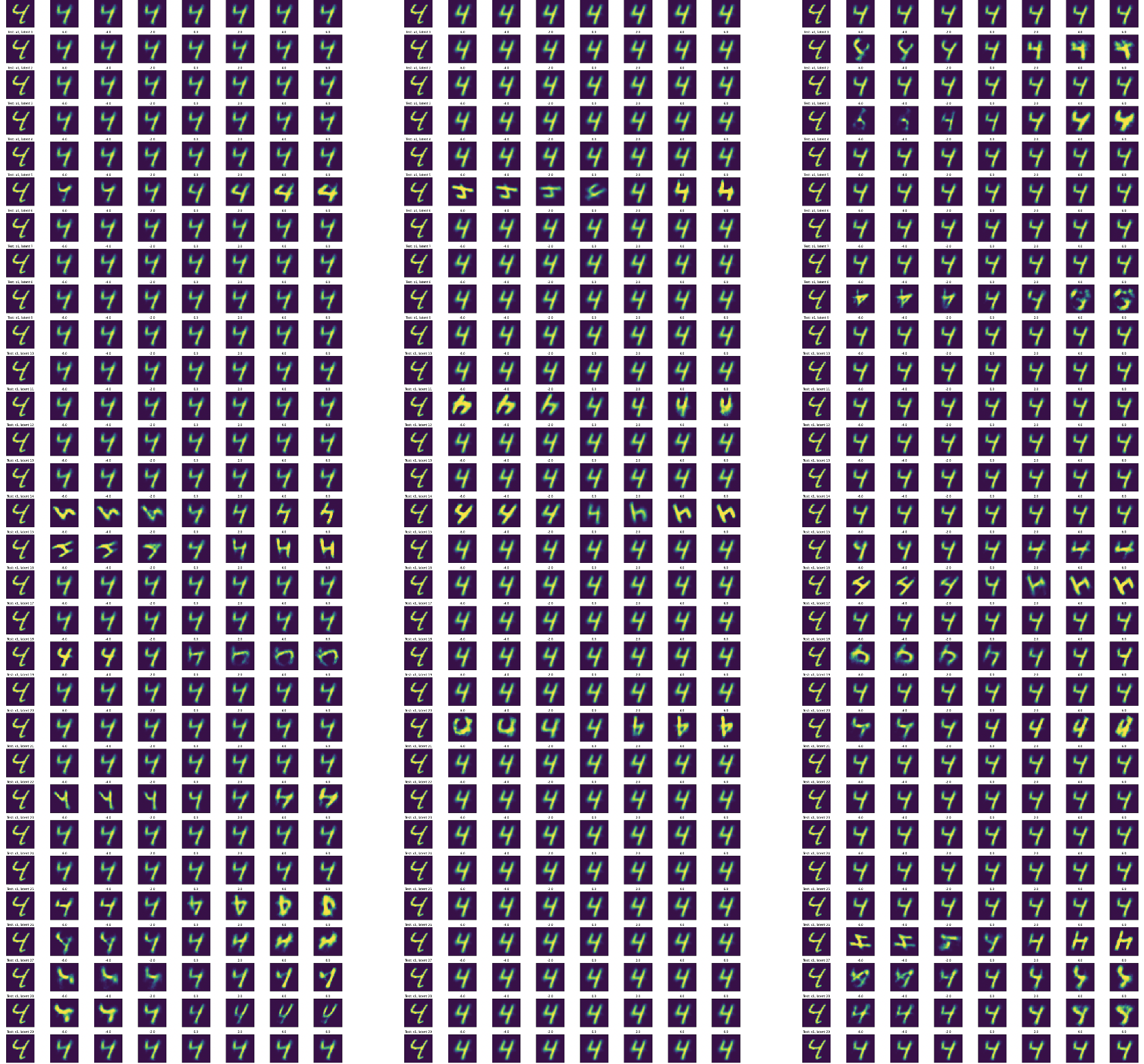}
\caption{\textbf{Single Rotated MNIST Digit Label}: Latent traversals for VAE (left), $\beta$-VAE (middle), CCI-VAE (right) trained on a single rotated MNIST digit ($10$ rotations). Latent traversal spans the range $[-6, 6]$ for each latent dimension.}
\label{fig:appendixVAEsingle}
\end{figure}

\begin{figure}[ht]
    \includegraphics[width=\textwidth]{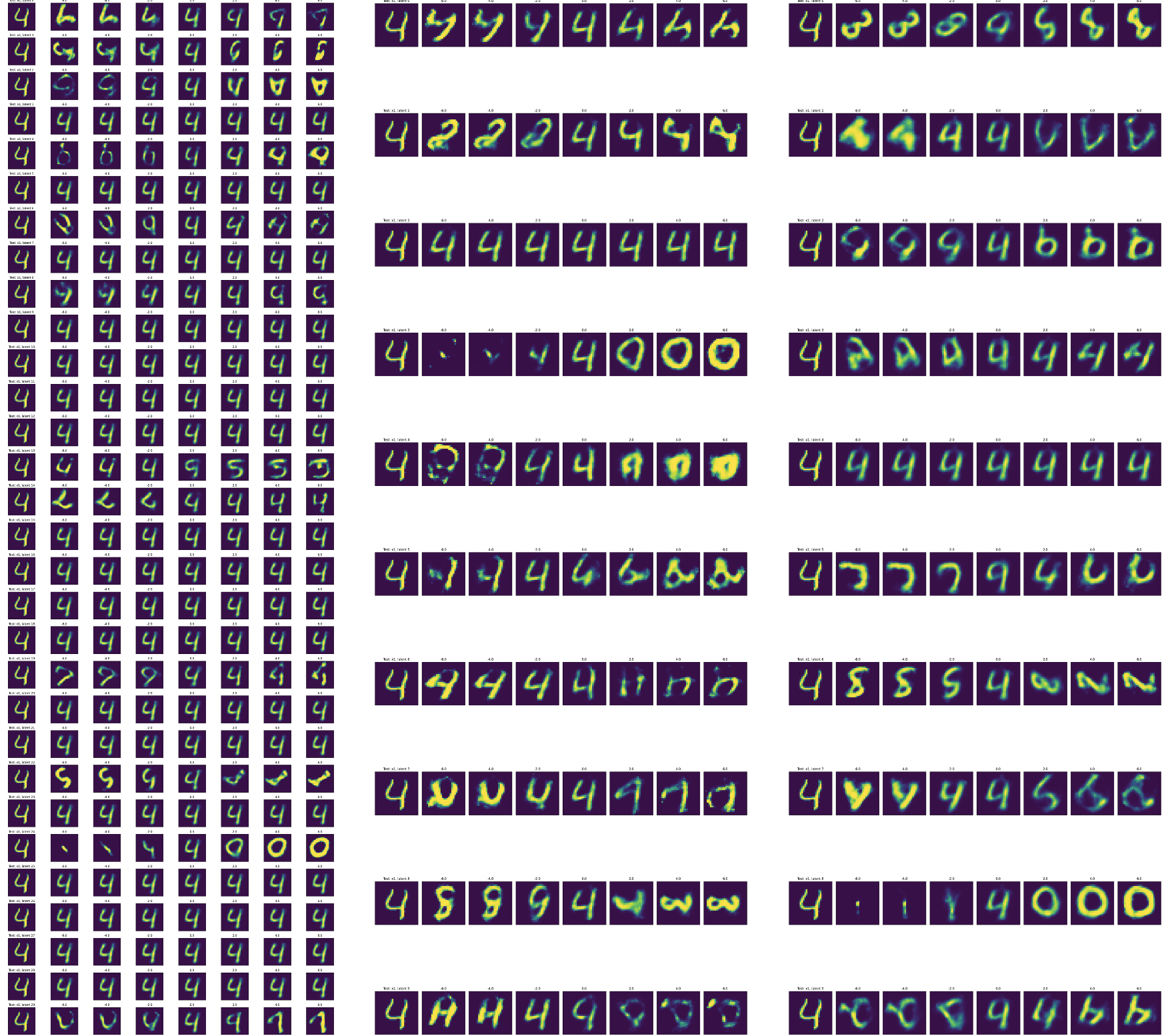}
\caption{\textbf{Rotated MNIST}: Latent traversals for VAE (left), $\beta$-VAE (middle), CCI-VAE (right) trained on all rotated MNIST digits ($10$ rotations). Latent traversal spans the range $[-6, 6]$ for each latent dimension. Note in this case, the best validation model for VAE contained 30 latent dimensions whereas $\beta$-VAE and CCI-VAE contain 10. }
\label{fig:appendixVAEmnist}
\end{figure}

\paragraph{Additional latent traversals} Figures \ref{fig:appendixVAEsingle} and \ref{fig:appendixVAEmnist} show all latent traversals for the model with the best validation loss. We see the success of disentanglement in the case of a single digit and the failure of a latent to capture rotation in the case of multiple digits. In Figures \ref{fig:appendixVAEsingleShiftX}, \ref{fig:appendixVAEsingleShiftY} we show comparable results for the case of translations along the x-axis and y-axis. 

%We show a more granular traversal with 50 plots per latent for each baseline in \ref{fig:appendixVAEDenseTraversal}

%\paragraph{Controls for the disentangled operator}
%\label{appendixControls}
%The lack of expressivity of the disentangled operator model can only partly be explained by the complexity of the encoder and the decoder. Indeed, a simple autoencoder trained on the same dataset of simple shapes without any transformation between the input image and target image reconstructs shapes better than its disentangled counterpart (see fig. \ref{fig:appendixDisentangledControl2}). This discrepancy in performance points at an additional difficulty of finding a disentangled representation for these shapes.  In addition, increasing the expressivity of the latent representation by increasing the latent dimension from 30 to 800 (equating the number of pixels in the image) does not yield good disentangled representations for all tested shapes (see fig. \ref{fig:appendixDisentangledControl}). 

\paragraph{Additional results for the distributed operator models} Fig. \ref{fig:appendixShiftMNIST} shows additional results on MNIST. Fig. \ref{fig:MNISTweak} shows that the weakly supervised shift operator performs well on Rotated MNIST, and in Fig. \ref{fig:MNISTSM} we see the stacked shift operator model is able to correctly encode the multiple transformations case on MNIST. Fig. \ref{fig:appendixShiftShapes} shows performance of the weakly supervised shift operator on translated simple shapes (either $x$ or $y$ translations). The effect of the number of latent transformations is explored in App. \ref{sec:appendixReddy}. It shows the best model is in the case of translation obtained with $21$ transformations. Nonetheless, to be able to feed to the model the correct $10$ transformations needed in these plots, the reported plots are for the model with $10$ latent transformations.
Fig. \ref{fig:appendixShapesSM2} shows example results for the case of the Special Euclidean group (i.e. rotations in conjunction with translations) in a setting where the semi-direct product structure is broken (see Appendix \ref{sec:se2note}). Here, we used a rotation group of order $5$. We see the stacked shift operator model performs well nonetheless.

%\begin{figure}[ht]
%\begin{subfigure}{\textwidth}
%  \centering
%      \includegraphics[width=\textwidth]{vae-dense-traversal.png}
%\caption{Granular latent traversals for VAE trained on all rotated MNIST digits ($10$ rotations).}
%\end{subfigure}
%\hfill
%\begin{subfigure}{\textwidth}
%  \centering
%    \includegraphics[width=\textwidth]{beta-vae-dense-traversal.png}
%\caption{Granular latent traversals for %$\beta$-VAE trained on all rotated MNIST %digits ($10$ rotations).}
%\end{subfigure}
%\hfill
%\begin{subfigure}{\textwidth}
%  \centering
%    \includegraphics[width=\textwidth]{cci-vae-dense-traversal.png}
%\caption{Granular latent traversals for CCI-VAE trained on all rotated MNIST digits ($10$ rotations).}
%\end{subfigure}
%\caption{Granular latent traversals for VAE $\beta$-VAE, CCI-VAE trained on all rotated MNIST digits ($10$ rotations). Latent traversal spans the range $[-6, 6]$ with step size $0.25$ yielding 50 plots per latent dimension. Note in this case, the best validation model for VAE contained 30 latent dimensions whereas $\beta$-VAE and CCI-VAE contain 10}
%\label{fig:appendixVAEDenseTraversal}
%\end{figure}

\begin{figure}[ht]
\begin{subfigure}{.45\textwidth}
  \centering
      \includegraphics[width=\textwidth]{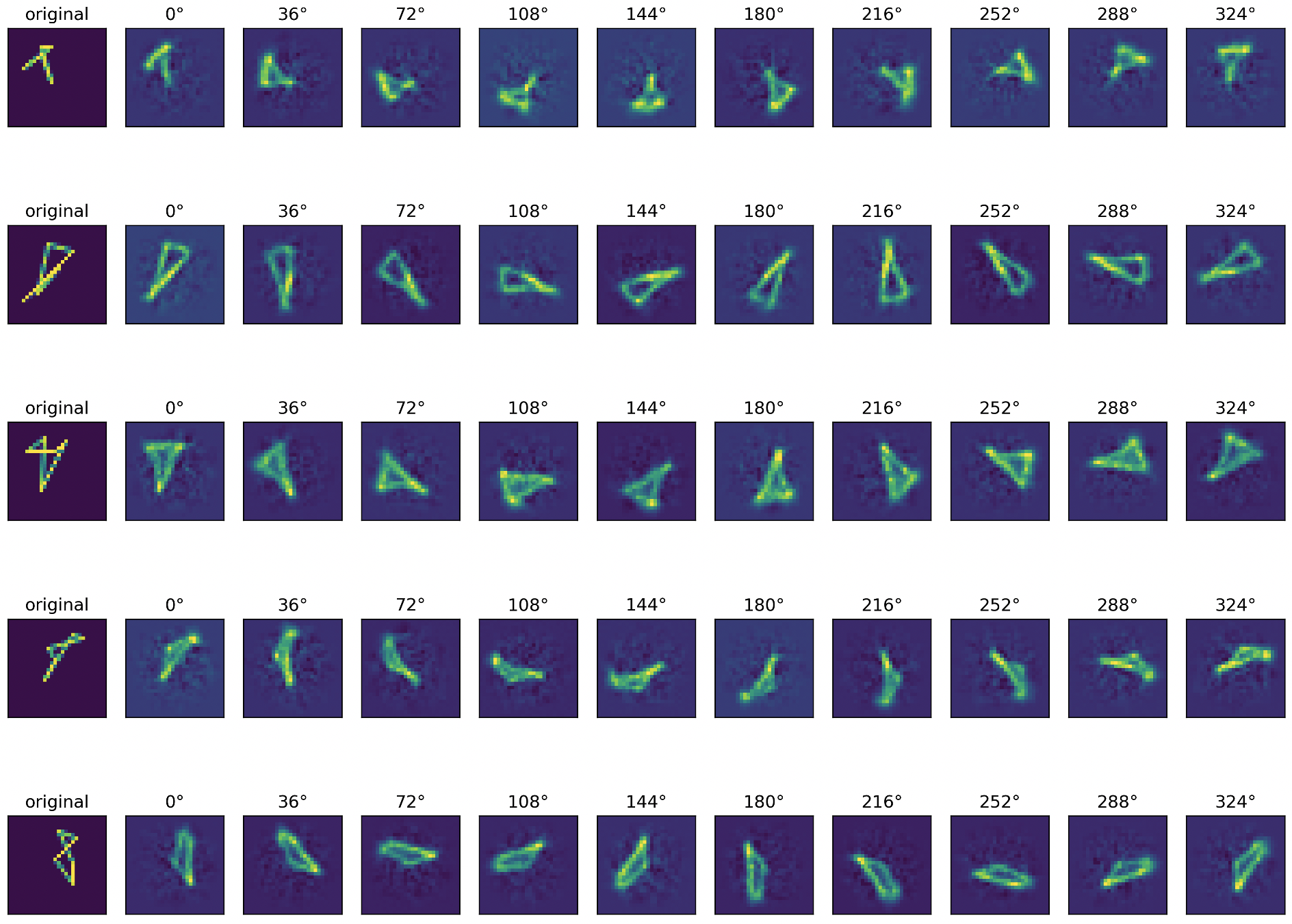}
\caption{Supervised disentangled operator on Rotated Shapes ($10$ rotations)}
\label{fig:appendixdisentangledShape}
\end{subfigure}
\hfill
\begin{subfigure}{.45\textwidth}
  \centering
    \includegraphics[width=\textwidth]{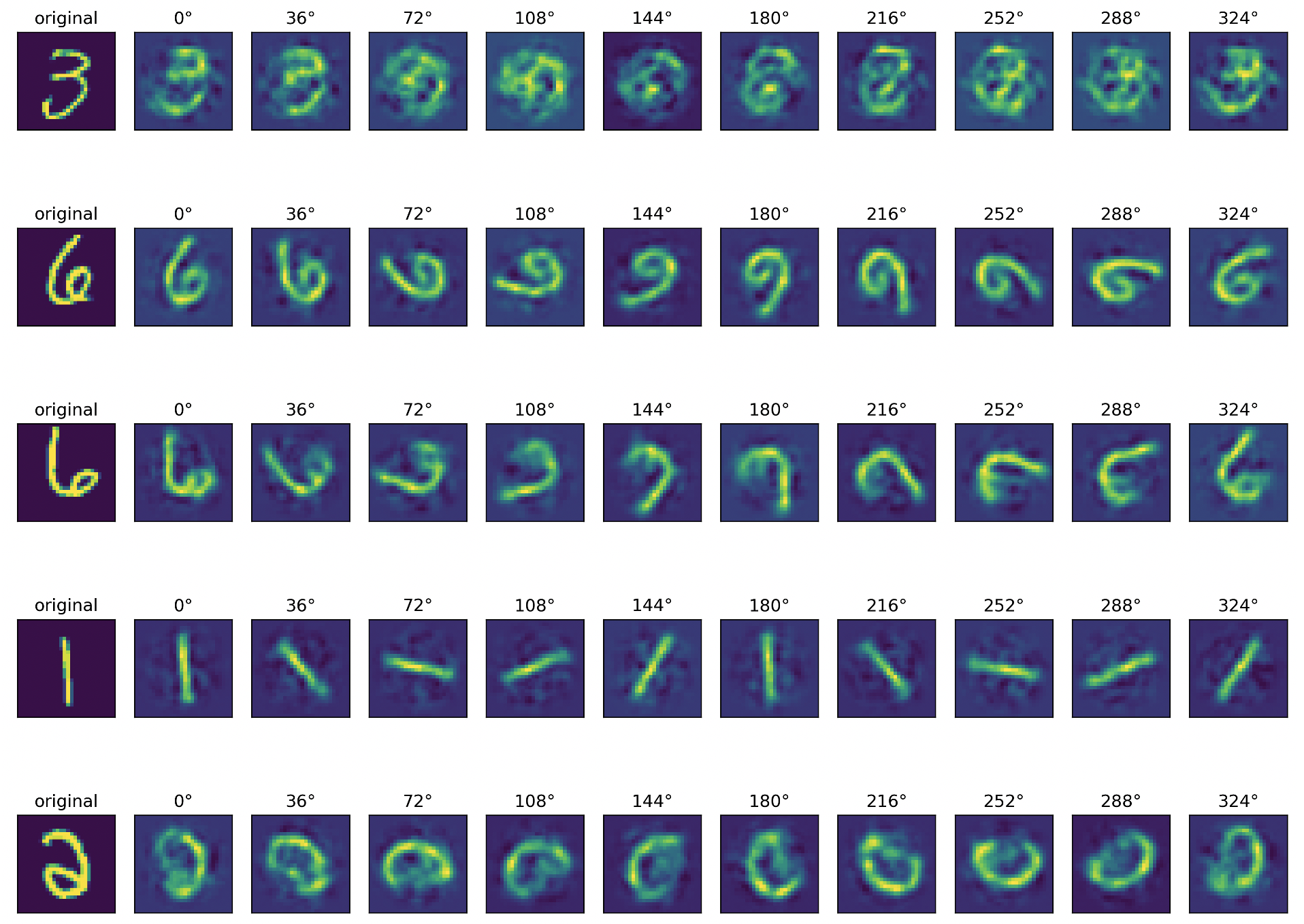}
\caption{Supervised disentangled operator on Rotated MNIST ($10$ rotations).}
\label{fig:appendixdisentangledMNIST}
\end{subfigure}
\caption{Non-linear disentangled operator with latent rotations}
\label{fig:appendixdisentangled}
\end{figure}

%\begin{figure}[ht]
%      \includegraphics[width=\textwidth]{standardAE_shapes.png}
%\caption{\textbf{Standard autoencoder with 30 latents}: We see a standard autoencoder with the same training setup as the disentangled operator in Fig. \ref{fig:fig1}.F is able to reconstruct several shapes the disentangled operator was not. In addition, we see a noticeable difference in the total MSE on the test set: 0.010 (standard autoencoder) versus 0.013 (disentangled operator).
%}
%\label{fig:appendixDisentangledControl2}
%\end{figure}

%\begin{figure}[ht]
%\includegraphics[width=\textwidth]{disentangledShapes_800_latents.png}
%\caption{\textbf{Non-linear supervised disentangled operator with 800 latents}: we see 
%the failure of the disentangled operator to reconstruct certain shapes even when we increase the number of latent dimension beyond the pixel size to 800 latents. This model was selected using the same hyper-parmeter cross validation process as other the disentangled models trained on 2000 shapes with $10$ rotations.}
%\label{fig:appendixDisentangledControl}
%\end{figure}

\begin{figure}[ht]
    \includegraphics[width=\textwidth]{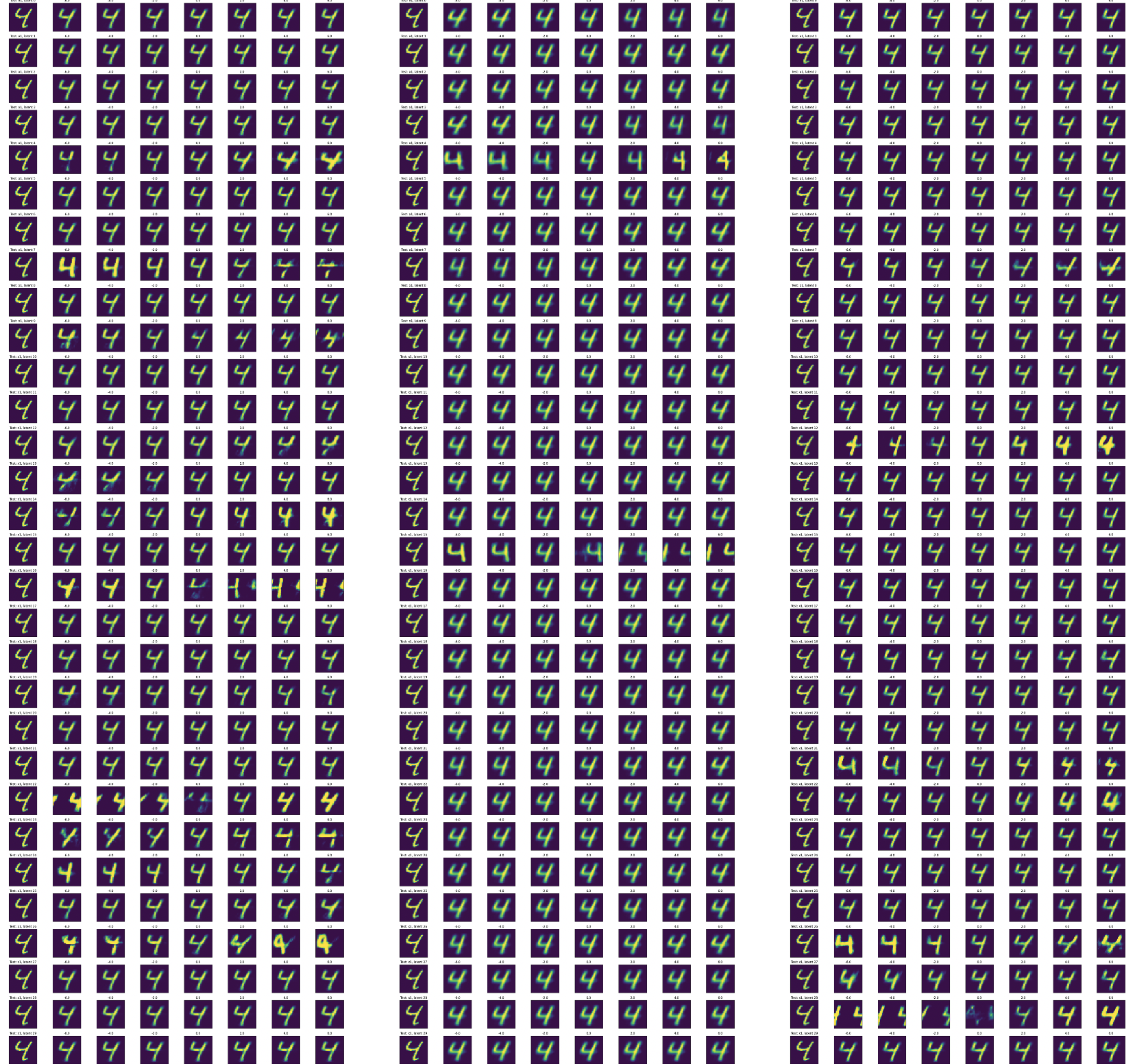}
\caption{\textbf{Single MNIST Digit Label Translated along X-Axis}: Latent traversals for VAE (left), $\beta$-VAE (middle), CCI-VAE (right) trained on a single MNIST digit translated along the $x-$axis ($10$ translations). Latent traversal spans the range $[-6, 6]$ for each latent dimension.}
\label{fig:appendixVAEsingleShiftX}
\end{figure}

\begin{figure}[ht]
    \includegraphics[width=\textwidth]{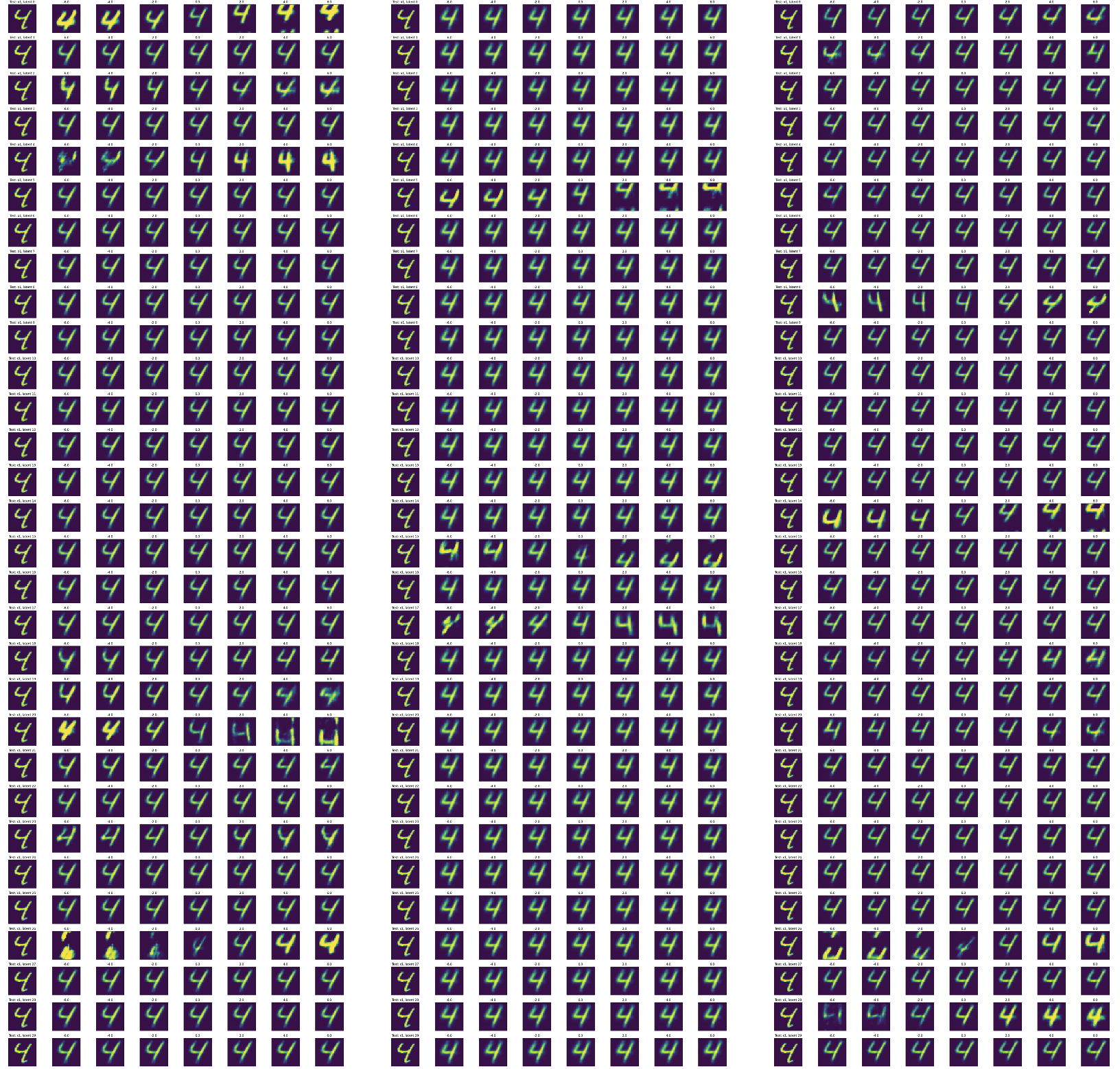}
\caption{\textbf{Single MNIST Digit Label Translated along Y-Axis}: Latent traversals for VAE (left), $\beta$-VAE (middle), CCI-VAE (right) trained on a single MNIST digit translated along the $y-$axis ($10$ translations). Latent traversal spans the range $[-6, 6]$ for each latent dimension.}
\label{fig:appendixVAEsingleShiftY}
\end{figure}

\begin{figure}[ht]
    \includegraphics[width=\textwidth]{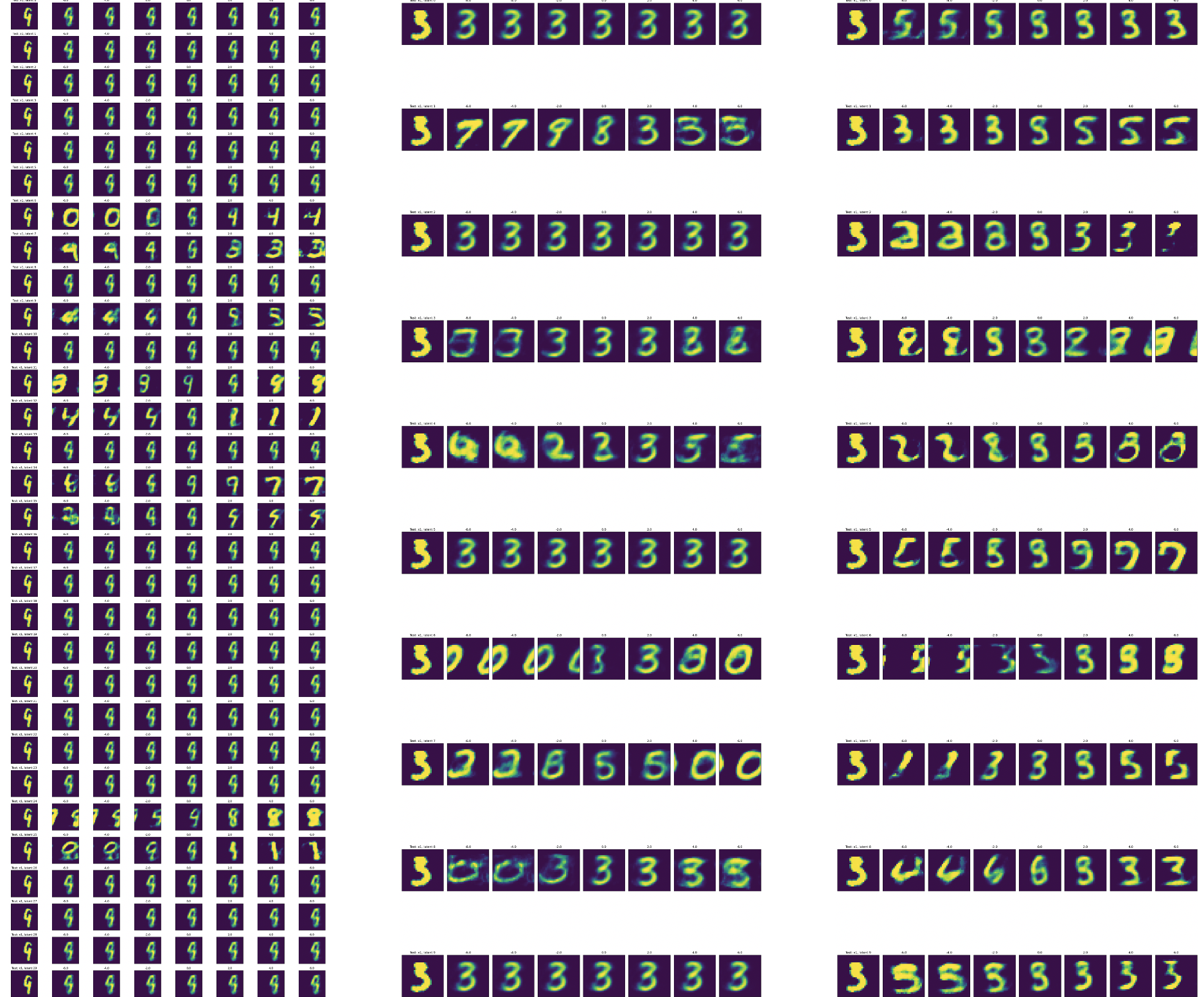}
\caption{\textbf{MNIST Translated along X-Axis}: Latent traversals for VAE (left), $\beta$-VAE (middle), CCI-VAE (right) trained on all MNIST digits translated along the x-axis ($10$ translations). Latent traversal spans the range $[-6, 6]$ for each latent dimension.}
\label{fig:appendixVAEmnistShiftX}
\end{figure}

\begin{figure}[ht]
    \includegraphics[width=\textwidth]{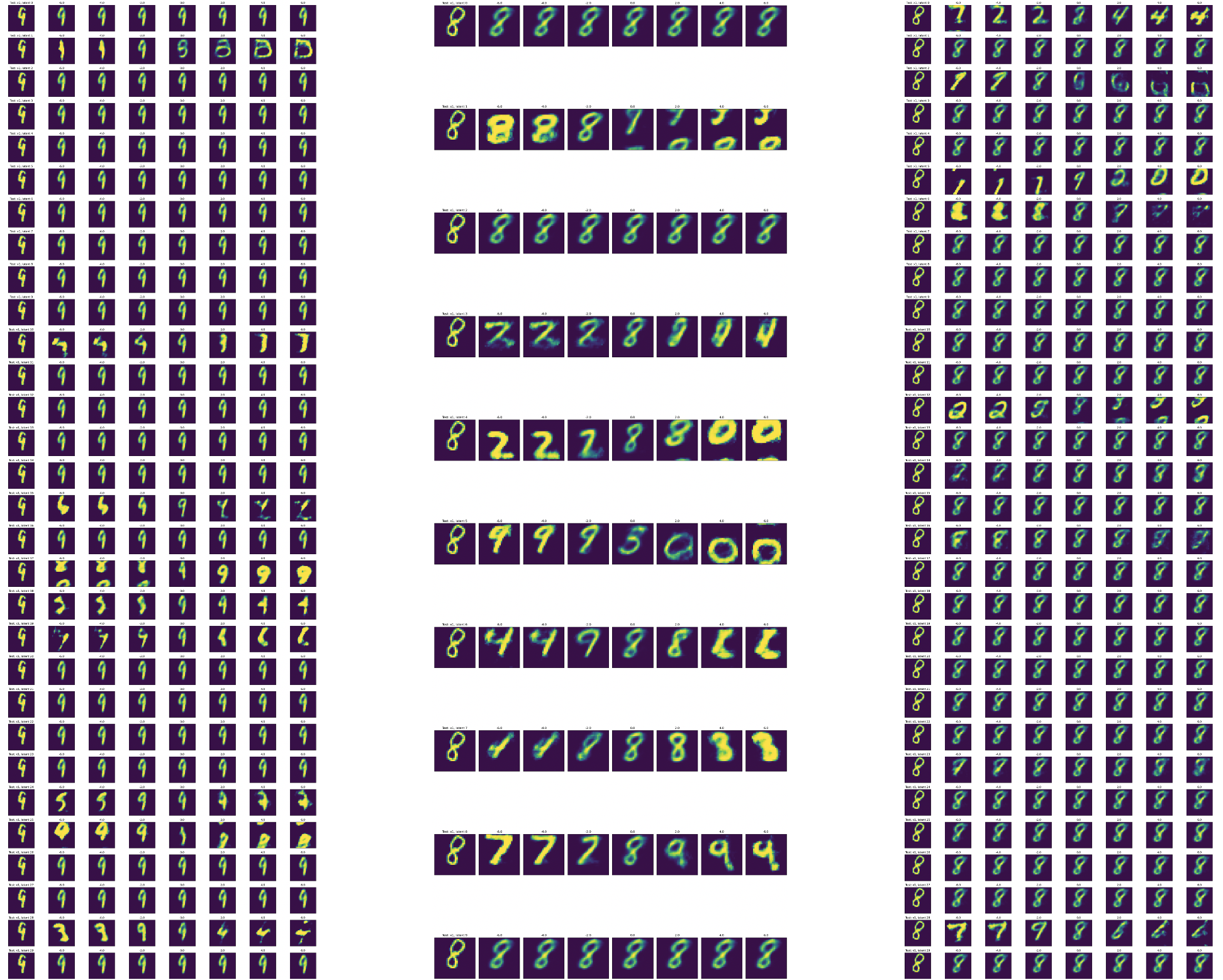}
\caption{\textbf{MNIST Translated along Y-Axis}: Latent traversals for VAE (left), $\beta$-VAE (middle), CCI-VAE (right) trained on all MNIST digits translated along the y-axis ($10$ translations). Latent traversal spans the range $[-6, 6]$ for each latent dimension.}
\label{fig:appendixVAEmnistShiftY}
\end{figure}

\begin{figure}[ht]
\begin{subfigure}{.45\textwidth}
  \centering
      \includegraphics[width=\textwidth]{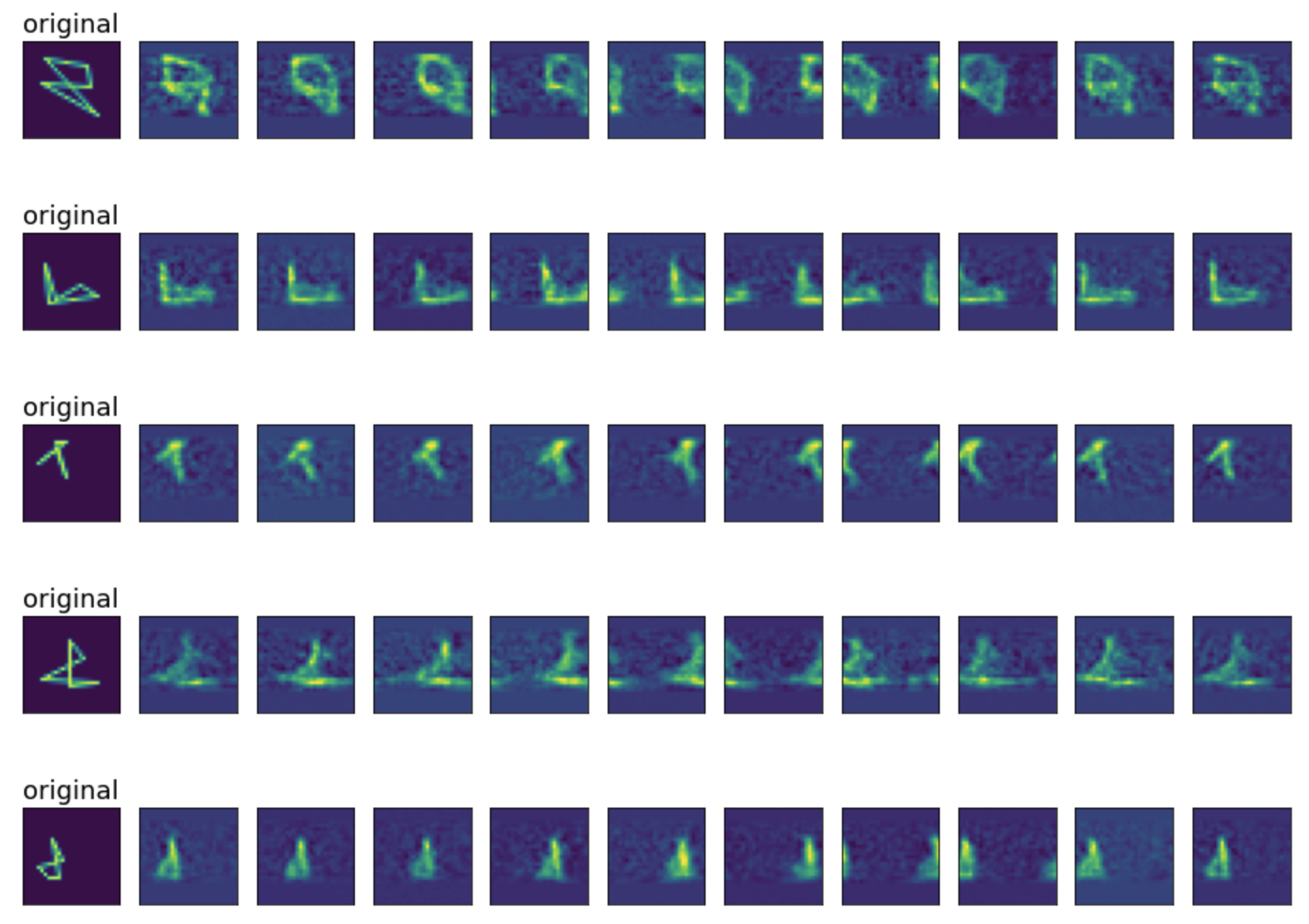}
\caption{Supervised disentangled operator on translated shapes along the x-axis ($10$ translations)}
\label{fig:appendixdisentangledShapeShiftX}
\end{subfigure}
\hfill
\begin{subfigure}{.45\textwidth}
  \centering
    \includegraphics[width=\textwidth]{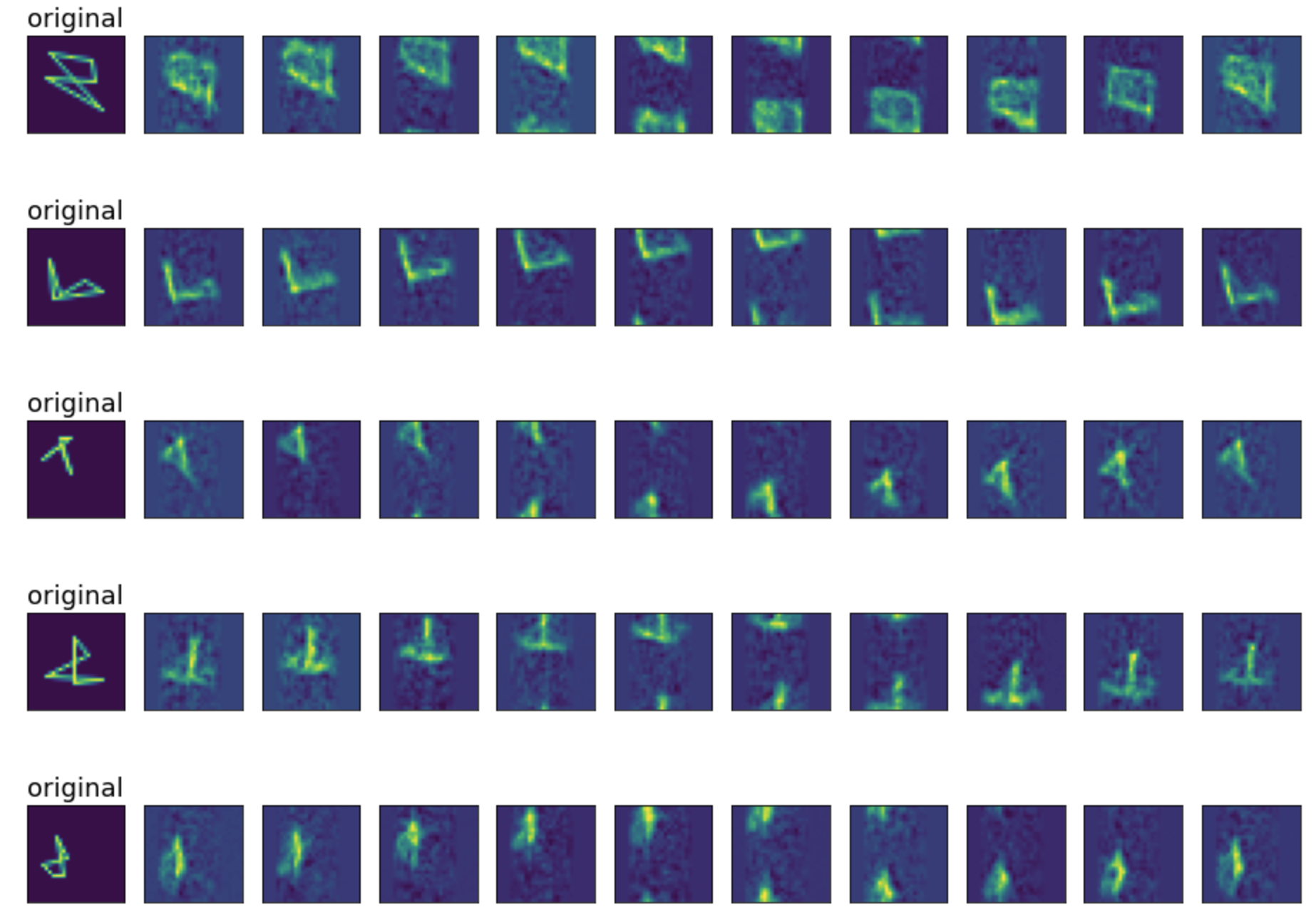}
\caption{Supervised disentangled operator on translated shapes along y-axis ($10$ translations).}
\label{fig:appendixdisentangledShapeShiftY}
\end{subfigure}
\caption{Non-linear disentangled operator with latent translations}
\label{fig:appendixDisentangledShifts}
\end{figure}

\clearpage
\begin{figure}[ht]
\begin{subfigure}{.45\textwidth}
  \centering
    \includegraphics[width=\textwidth]{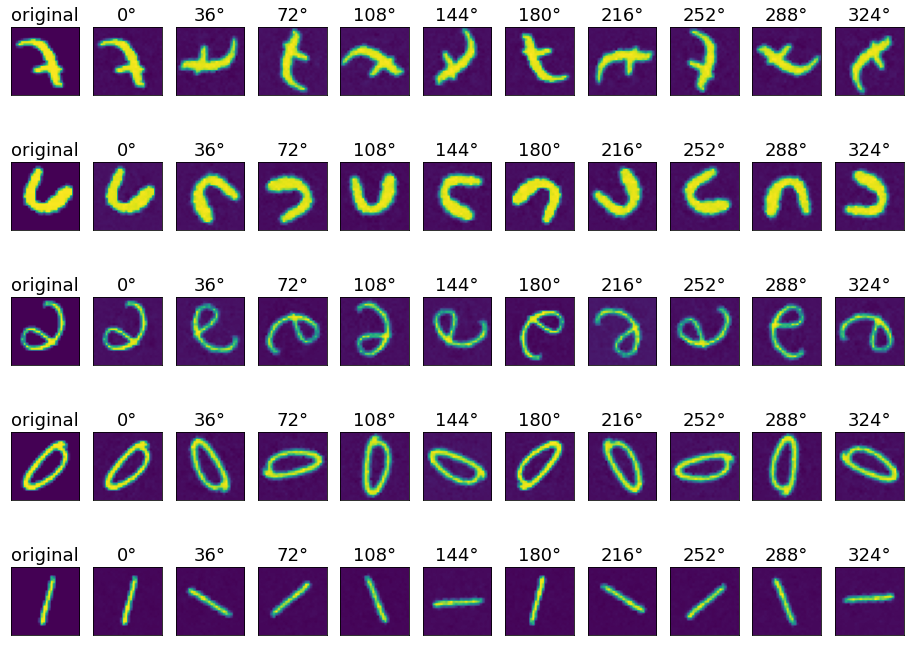}
\caption{Weakly supervised shift operator on Rotated MNIST ($10$ rotations).}
\label{fig:MNISTweak}
\end{subfigure}
\hfill
\begin{subfigure}{.45\textwidth}
  \centering
      \includegraphics[width=\textwidth]{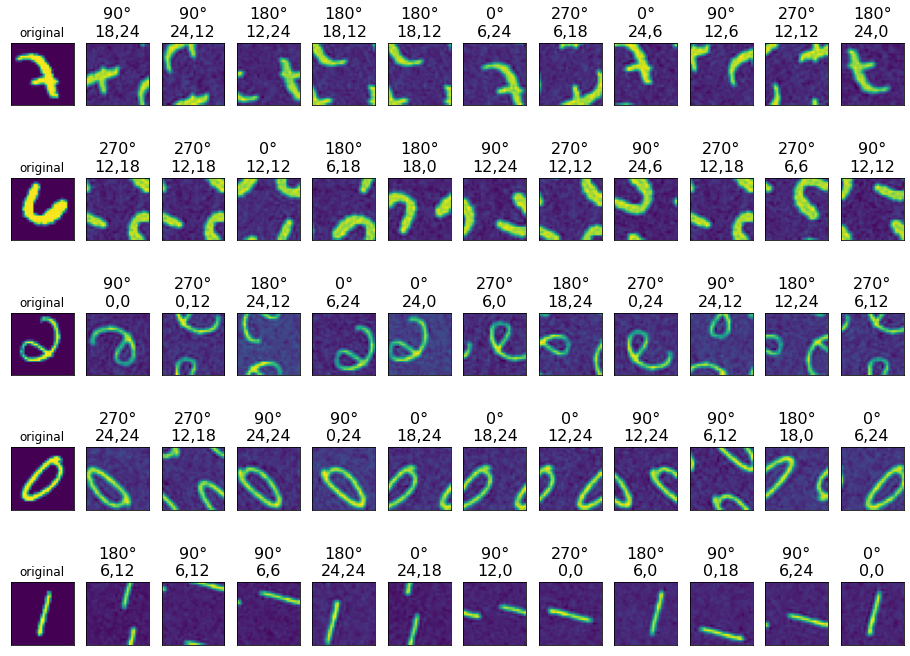}
\caption{Supervised shift operator on Rotated-Translated MNIST ($4$ rotations, $5$ $x$-translations and 5 $y$-translations).}
\label{fig:MNISTSM}
\end{subfigure}
\caption{MNIST additional experiments.}
\label{fig:appendixShiftMNIST}
\end{figure}

\begin{figure}[ht]
\begin{subfigure}{.45\textwidth}
  \centering
    \includegraphics[width=\textwidth]{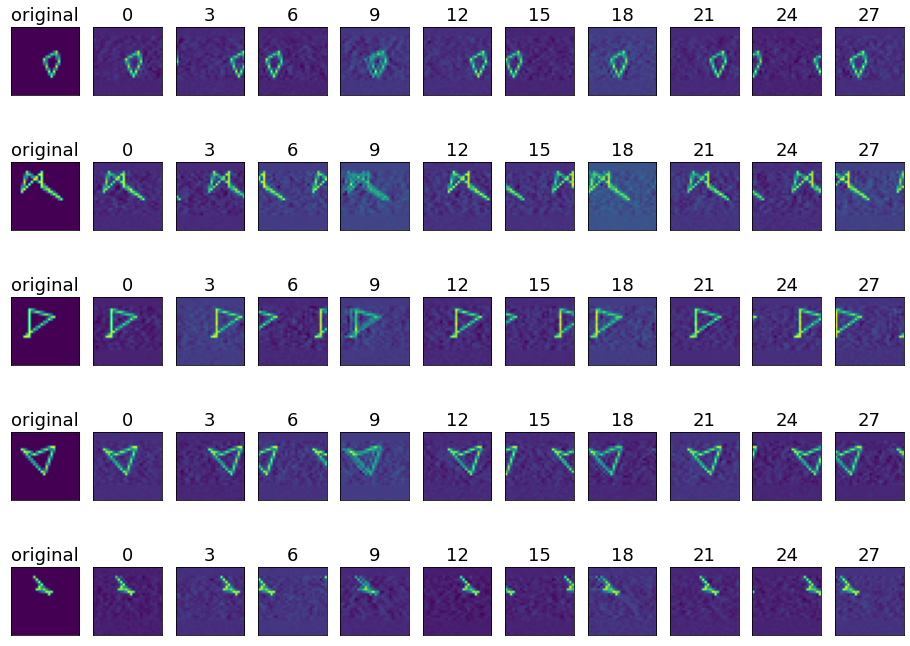}
\caption{Weakly supervised shift operator on $x$-translations ($10$ translations).}
\label{fig:ShapesTxweak}
\end{subfigure}
\hfill
\begin{subfigure}{.45\textwidth}
  \centering
      \includegraphics[width=\textwidth]{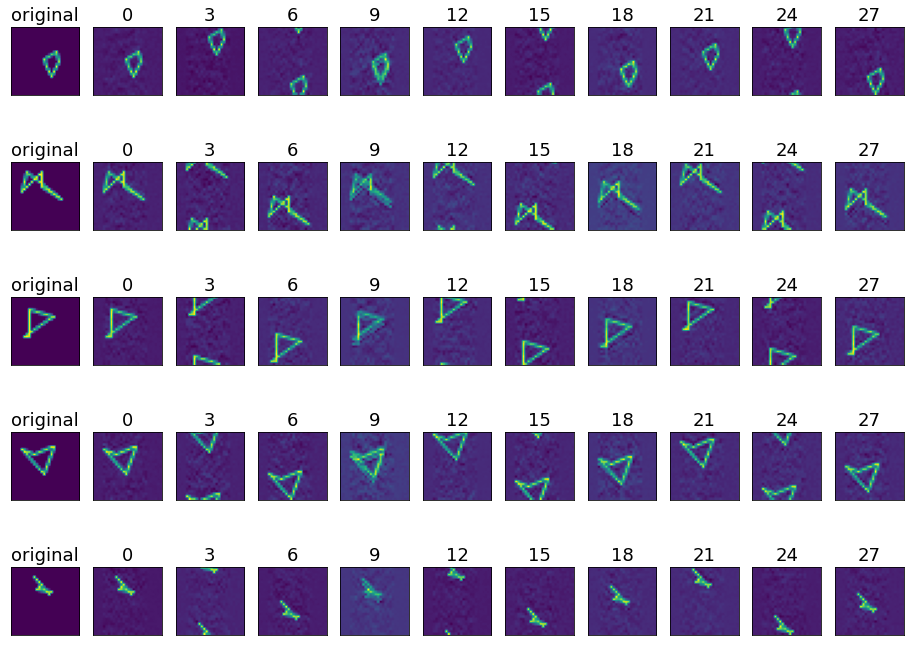}
\caption{Weakly supervised shift operator on $y$-translations ($10$ translations).}
\label{fig:ShapesTyweak}
\end{subfigure}
\caption{Simple shapes additional experiments.}
\label{fig:appendixShiftShapes}
\end{figure}

\begin{figure}[h]
  \centering
    \includegraphics[width=0.5\textwidth]{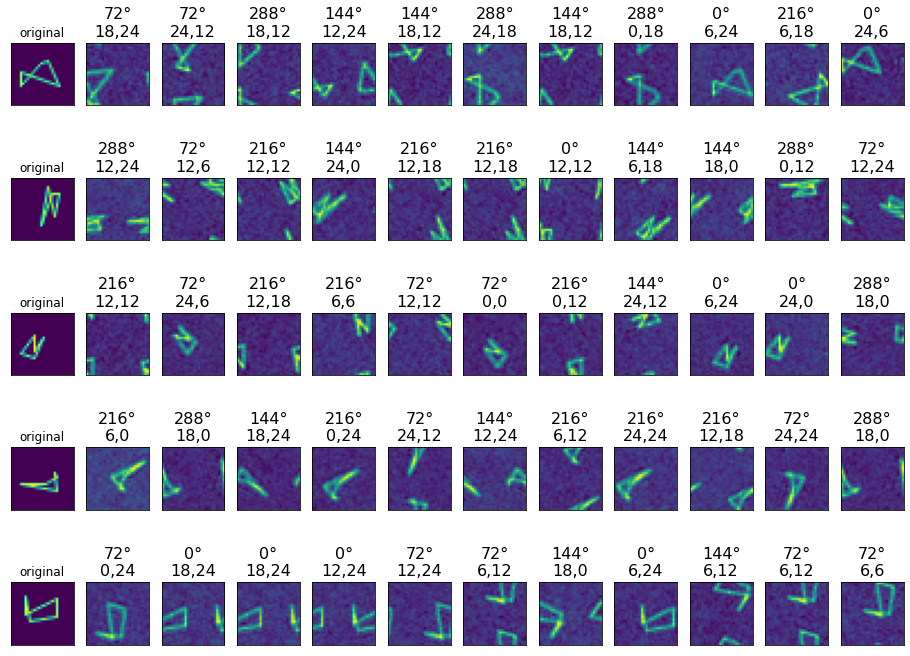}
\caption{Supervised shift operator on Rotated-Translated simple shapes when the semi-direct product structure is not respected as rotations angles are $j \frac{\pi}{5},~j=1,\ldots 5$ ($5$ rotations, $5$ $x$-translations and 5 $y$-translations).}
\label{fig:appendixShapesSM2}
\end{figure}

\begin{figure}[h]
  \centering
    \includegraphics[width=0.5\textwidth]{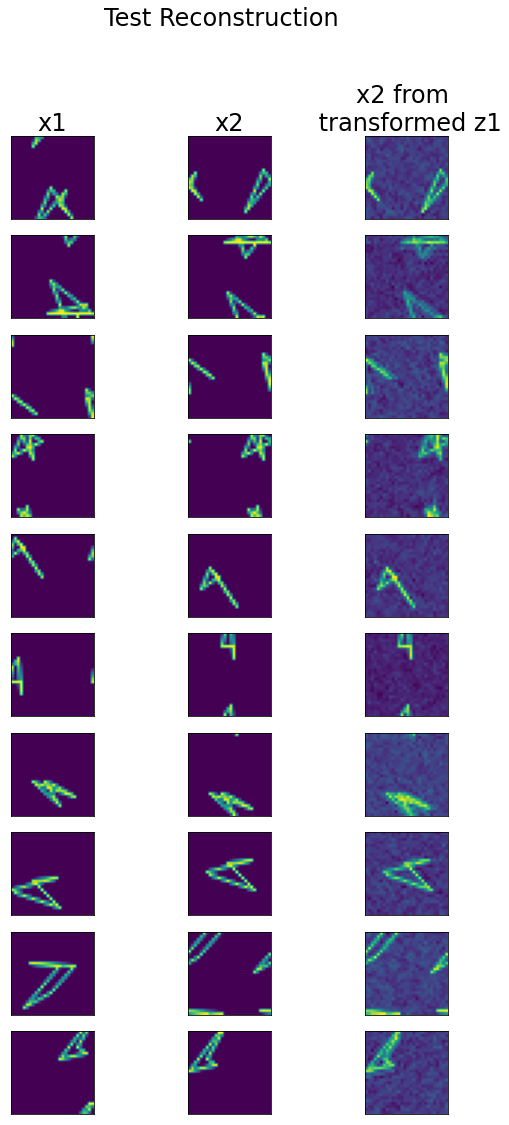}
\caption{Pairs of test samples and their reconstructions for the stacked shift model with $5$ translations in both $x$ and $y$.}
\label{fig:reconstructionstxty}
\end{figure}

\begin{figure}[h]
  \centering
     \includegraphics[width=0.5\textwidth]{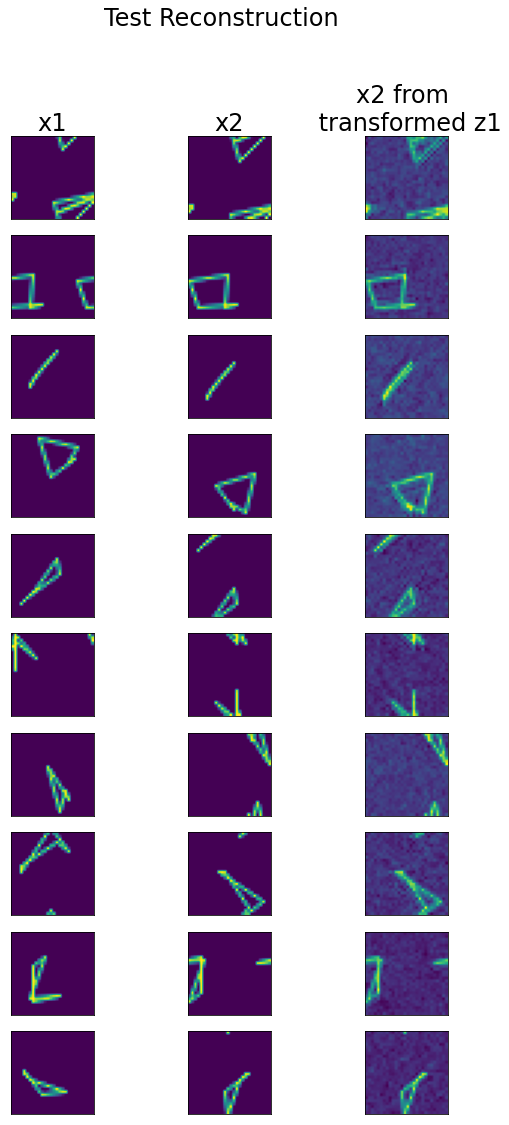}
\caption{Pairs of test samples and their reconstructions for the stacked shift model with $4$ rotations and $5$ translations in both $x$ and $y$.}
\label{fig:reconstructionsRtxty}
\end{figure}

\end{document}